%% file: main.tex
\pdfoutput=1

\documentclass[11pt]{article}

\usepackage[preprint]{acl}

\usepackage{times}
\usepackage{latexsym}

\usepackage[T1]{fontenc}

\usepackage[utf8]{inputenc}

\usepackage{microtype}

\usepackage{inconsolata}

\usepackage{lipsum}
\usepackage{array}
\usepackage{booktabs}
\usepackage{makecell}
\usepackage{amsmath}
\usepackage{amssymb}
\usepackage{graphicx}
\usepackage{multirow}
\usepackage{color, colortbl}
\usepackage{bold-extra} 
\usepackage[most]{tcolorbox}
\usepackage{epsfig}
\usepackage{csquotes}
\usepackage{siunitx} 
\usepackage{subcaption}
\usepackage{capt-of}
\usepackage{hyperref}
\usepackage{placeins}

\usepackage{fontawesome5}
\usepackage{pifont, xcolor}
\usepackage[frozencache,cachedir=.]{minted}
\usepackage{listings}

\input{_commands}

\title{\texttt{GIMMICK}\\Globally Inclusive Multimodal Multitask  Cultural Knowledge Benchmarking}

\author{
  \textbf{Florian Schneider\textsuperscript{1}},
  \textbf{Carolin Holtermann\textsuperscript{2}},
  \textbf{Chris Biemann\textsuperscript{1}},
  \textbf{Anne Lauscher\textsuperscript{2}}
\\
  \textsuperscript{1}Language Technology Group, University of Hamburg
\\
  \textsuperscript{2}Data Science Group, University of Hamburg
\\
  \small
  \texttt{
    \href{mailto:florian.schneider-1@uni-hamburg.de}{florian.schneider-1@uni-hamburg.de}
  }
}

\begin{document}
\maketitle
\input{src/00_abstract}
\input{src/01_intro}
\input{src/02_related_work}
%
\input{src/030_gimmick_overview}
\input{src/040_analyses}
%
\input{src/880_conclusion}
\input{src/881_limitations}
\input{src/882_ethical}
\input{src/883_ack}

\clearpage
\bibliography{custom}
\input{src/990_appendix}
\end{document}

%% file: _commands.tex
\usepackage{xspace} 

\newcommand{\RegW}{\textcolor[HTML]{636EFA}{\rule{7pt}{7pt}}\hspace{0.15em}\large\texttt{\textbf{W}}\xspace}
\newcommand{\RegE}{\textcolor[HTML]{EF553B}{\rule{7pt}{7pt}}\hspace{0.15em}\large\texttt{\textbf{E}}\xspace}
\newcommand{\RegAP}{\textcolor[HTML]{00CC96}{\rule{7pt}{7pt}}\hspace{0.15em}\large\texttt{\textbf{AP}}\xspace}
\newcommand{\RegA}{\textcolor[HTML]{AB63FA}{\rule{7pt}{7pt}}\hspace{0.15em}\large\texttt{\textbf{A}}\xspace}
\newcommand{\RegLAC}{\textcolor[HTML]{FFA15A}{\rule{7pt}{7pt}}\hspace{0.15em}\large\texttt{\textbf{LAC}}\xspace}
\newcommand{\RegSA}{\textcolor[HTML]{19D3F3}{\rule{7pt}{7pt}}\hspace{0.15em}\large\texttt{\textbf{SA}}\xspace}

\newcommand{\dsname}{\texttt{GIMMICK}\xspace}
\newcommand{\sivqa}{\texttt{CIVQA}\xspace}
\newcommand{\vvqa}{\texttt{CVVQA}\xspace}
\newcommand{\coqa}{\texttt{COQA}\xspace}
\newcommand{\coqar}{\texttt{COQA}\textsubscript{\texttt{\textbf{R}}}\xspace}
\newcommand{\coqac}{\texttt{COQA}\textsubscript{\texttt{\textbf{C}}}\xspace}
\newcommand{\ckqa}{\texttt{CKQA}\xspace}
\newcommand{\ckqad}{\texttt{CKQA}\textsubscript{\texttt{\textbf{D}}}\xspace}
\newcommand{\ckqan}{\texttt{CKQA}\textsubscript{\texttt{\textbf{N}}}\xspace}

\newcommand{\m}[1]{{\small\textsc{#1}}\xspace}

\newcommand{\rparagraph}[1]{\vspace{1.4mm}\noindent\textbf{#1.}}
\newcommand{\rrparagraph}[1]{\vspace{0.4mm}\noindent\textit{#1.}}

\newtcolorbox{promptbox}[1]{colback=gray!5!white,colframe=black!75!black,fonttitle=\bfseries\scriptsize,fontupper=\ttfamily\footnotesize,title=#1}

%% file: src/00_abstract.tex
\begin{abstract}
Large Vision-Language Models (LVLMs) have recently gained attention due to their distinctive performance and broad applicability.
While it has been previously shown that their efficacy in usage scenarios involving non-Western contexts falls short, existing studies are limited in scope, covering just a narrow range of cultures, focusing exclusively on a small number of cultural aspects, or evaluating a limited selection of models on a single task only.
Towards globally inclusive LVLM research, we introduce \dsname, an extensive multimodal benchmark designed to assess a broad spectrum of cultural knowledge across 144 countries representing six global macro-regions.
\dsname comprises six tasks built upon three new datasets that span 728 unique cultural events or facets on which we evaluated 20 LVLMs and 11 LLMs, including five proprietary and 26 open-weight models of all sizes.
We systematically examine (1) regional cultural biases, (2) the influence of model size, (3) input modalities, and (4) external cues.
Our analyses reveal strong biases toward Western cultures across models and tasks and highlight strong correlations between model size and performance, as well as the effectiveness of multimodal input and external geographic cues.
We further find that models have more knowledge of tangible than intangible aspects (e.g., \emph{food} vs. \emph{rituals}) and that they excel in recognizing broad cultural origins but struggle with a more nuanced understanding.\footnote{\href{https://github.com/floschne/gimmick}{http://github.com/floschne/gimmick}}
\end{abstract}

%% file: src/01_intro.tex
\section{Introduction}
\label{sec:intro}
\begin{figure*}[t]
    \centering
    \includegraphics[width=1.\linewidth, trim=0 191.25 0 190.25, clip]{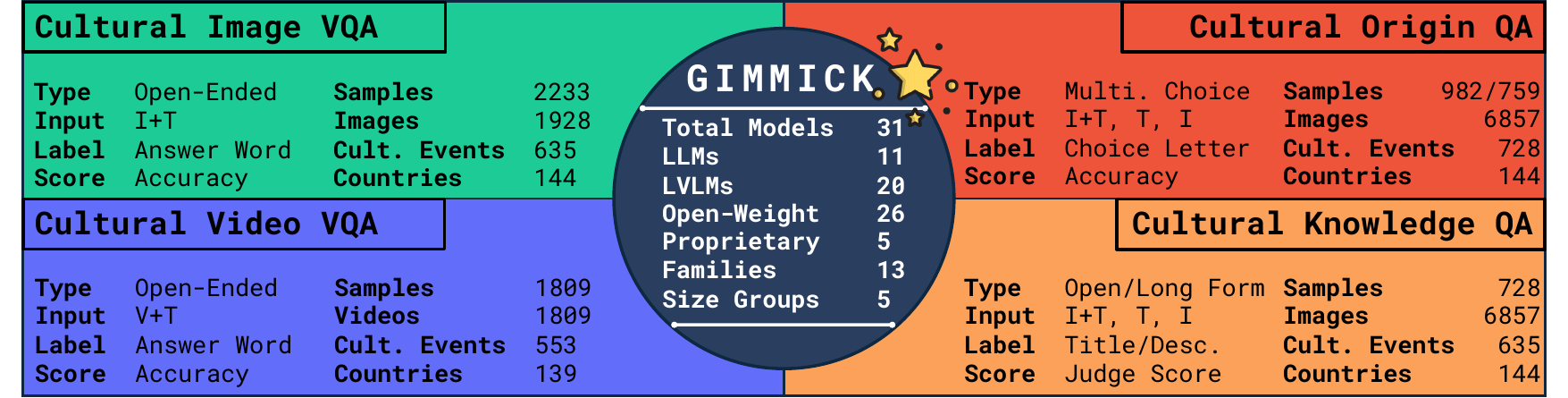}
    \caption{An overview of the \dsname benchmark and its tasks.}
    \label{fig:figure1}
\end{figure*}

Recently, proprietary as well as open-weight Large Vision-Language Models (LVLMs)~\cite[\textit{inter alia}]{openai2023gpt4v,liu2023llava,wang2024qwen2vl,chen2024internvl} have attracted marked attention due to their broad applicability across various domains.
Several large-scale holistic benchmarks~\cite{duan2024vlmevalkit,yue2023mmmu,fu2023mme} demonstrate LVLMs' remarkable performances in a wide range of multimodal tasks. 
However, most benchmarks concentrate on Western-centric English tasks, and multilingual benchmarks~\cite{ahuja2023megaverse,schneider-sitaram-2024-m5} reveal a significant deterioration in performance on non-English tasks.
While multilingualism is essential for globally equitable AI, \emph{multi-culturalism}~\cite{gabriel2020aivalues,adilazuarda-etal-2024-towards} is equally crucial for models to reflect and respect the diverse cultural backgrounds of users worldwide.
In this context, it has been shown that current LLMs~\cite{myung2024blend,chiu2024culturalbench} and LVLMs suffer in tasks involving knowledge from non-Western cultures. However, the scope of existing multimodal cultural studies is still severely limited:
Existing research often focuses only on specific concepts like food or dance~\cite{winata2024worldcuisines,burda2024culturally}, covers a limited number of cultures~\cite{urailertprasert-etal-2024-sea,baek2024kviscuit}, evaluates only a small selection of LVLM models~\cite{cao2024exploring,nayak-etal-2024-benchmarking}, or tests only a single combination of input modalities.

\noindent To address these gaps, we introduce \dsname, a comprehensive evaluation framework assessing 31 state-of-the-art models, ranging from proprietary LVLMs to open-weight LLMs and LVLMs of all sizes---from 500M to 78B parameters---across multiple model families.
It comprises six tasks built on three novel datasets that contain 728 unique cultural events or facets (CEFs) from 144 countries in six global macro-regions and target both high-level and nuanced cultural knowledge through multimodal and unimodal tasks.
Our VQA tasks span a total of $57$ cultural aspects (see~\S\ref{appedix:sec:sivqa:aspects})
Ultimately, \dsname enables us to answer four research questions:

\noindent\textbf{(RQ1) \emph{Are there regional biases in LLMs' and LVLMs' cultural knowledge, and if so, which?}} For the most complex tasks, we observe consistent cultural regional biases (up to 14.72pp difference between instances targeting Western Europe \& North America vs. Subsaharian Africa; \S\ref{sec:analyses:a1_bias}) -- even for the largest models. For less complex tasks, these differences flatten out. 

\noindent\textbf{(RQ2) \emph{To what degree does model size influence performance?}} 
We show that increasing the number of parameters significantly boosts performance on complex tasks, with larger models exhibiting less regional biases~(\S\ref{sec:analyses:a2_model}). Still, even the largest models still struggle with nuanced cultural understanding.

\noindent\textbf{(RQ3) \emph{How do input modalities affect cultural understanding?}} We observe that providing input in multiple modalities typically leads to the best results, as models leverage the cultural cues present in the visual inputs we provide (\S\ref{sec:analyses:a3_modality}). Interestingly, on text-only tasks, LVLMs perform consistently worse than their LLM backbones, indicating a loss of cultural knowledge during integration training. 

\noindent\textbf{(RQ4) \emph{What is the influence of external cultural cues?}} We demonstrate that providing country information consistently guides the models towards better answers, especially for regions for which the models perform poorly (\S\ref{sec:analyses:a4_external}). 
%
%
%
%
Overall, with \dsname, we hope to encourage more research on culturally-aware and more globally-inclusive AI. 

%% file: src/02_related_work.tex
\section{Related Work}
\label{sec:related}

\begin{table}[t]
  \centering
  \small
    \renewcommand{\arraystretch}{.85}
    \resizebox{\linewidth}{!}{%
\begin{tabular}{@{}l l l l l l l l@{}}
	\toprule
	\textsc{Benchmark}                    & \textsc{\#M} & \textsc{\#DS} & \textsc{\#T} & \textsc{\#S} & \textsc{\#C} & \textsc{\#R} & \textsc{Mods}                   \\ \midrule
	\makecell[l]{SEA-VQA\\\citet{urailertprasert-etal-2024-sea}} & 2            & 1             & 1            & 1,999         & 8            & 1            & \textbf{T}+\textbf{I}           \\
    \makecell[l]{WorldCuisines\\\citet{winata2024worldcuisines}}       & 18           & 1             & 2            & 1.15M        & 189          & 6            & \textbf{T}+\textbf{I}           \\
	\makecell[l]{CROPE\\\citet{nikandrou2024crope}}            & 17           & 1             & 1            & 1,060         & 6            & 3            & \textbf{T}+\textbf{I}           \\
	\makecell[l]{CulturalVQA\\\citet{nayak-etal-2024-benchmarking}}  & 8            & 1             & 1            & 2,378         & 11           & 5            & \textbf{T}+\textbf{I}           \\
	\citet{ananthram2024see}              & 10           & {--}          & 3            & {--}         & 2            & 2            & \textbf{T}+\textbf{I}           \\
	\makecell[l]{GlobalRG\\\citet{bhatia-etal-2024-local}}        & 12           & 2             & 2            & 3,591         & 51           & 6            & \textbf{T}+\textbf{I}           \\
	\makecell[l]{MOSAIC-1.5K\\\citet{burda2024culturally}}           & 4            & 1             & 1            & 1,500         & {--}         & 6            & \textbf{T}+\textbf{I}           \\
	\makecell[l]{FoodieQA\\\citet{li-etal-2024-foodieqa}}         & 8            & 1             & 3            & 1,839         & 1            & 1            & \textbf{T}+\textbf{I}           \\
	\citet{cao2024exploring}              & 1            & {--}          & 3            & {--}         & 5            & 3            & \textbf{T}+\textbf{I}           \\
	\makecell[l]{K-VISCUIT\\\citet{baek2024kviscuit}}              & 13           & 1             & 1            & 657          & 1            & 1            & \textbf{T}+\textbf{I}           \\
	\makecell[l]{CVQA\\\citet{romero2024cvqa}}                & 8            & 1             & 1            & 10,374        & 30           & 6            & \textbf{T}+\textbf{I}           \\
	\makecell[l]{CulturalBench\\\citet{chiu2024culturalbench}}                & 30            & 2             & 1            & 6,135        & 45           & 6            & \textbf{T}+\textbf{I}           \\ \midrule
	\dsname \textbf{(ours)}               & 31           & 3             & 6            & 7,239         & 144          & 6            & \makecell[l]{\textbf{T}+\textbf{I}\\\textbf{V}+\textbf{T}\\\textbf{T}, \textbf{I}} \\
	\bottomrule
\end{tabular}
  }
  \caption{A comparative overview of recent benchmarks assessing cultural knowledge of LVLMs. The abbreviations in the columns stand for the (combined) number of: (unique) \textbf{M}odels, \textbf{D}ata\textbf{s}ets, \textbf{T}asks, \textbf{S}amples, \textbf{C}ountries, or \textbf{Regions} contained. The \textbf{Mod}alities column lists the input modalities---\textbf{T}ext, \textbf{I}mage, \textbf{V}ideo---contained.}.
  \label{tab:rw:compare}
\end{table}

\rparagraph{Multicultural LLM Benchmarks}
%
\citet{naous-etal-2024-beer} introduce CAMeL, a dataset that contrasts Arab and Western cultures to measure cultural biases in LLMs through extrinsic and intrinsic evaluations on core NLP tasks.
With CultureAtlas, \citet{fung2024massively} introduced an approach for massively multicultural knowledge acquisition and benchmarking of 5 LLMs from Wikipedia articles on cultural topics.
BLEnD~\cite{myung2024blend} is a large benchmark to evaluate LLMs' everyday knowledge across diverse cultures and from 16 countries in 13 different languages.
\cite{mukherjee-etal-2024-cultural} test four popular LLMs with culturally sensitive and non-sensitive prompts on both sensitive and neutral datasets.
%
Instead of assessing models' intrinsic cultural knowledge, \cite{bhatt-diaz-2024-extrinsic} focuses on the extrinsic evaluation of cultural competence, e.g., in user-interaction, in two text generation tasks, open-ended question answering, and story generation of 6 LLMs.

\rparagraph{Multicultural LVLM Benchmarks}
%
%
\citet{bhatia-etal-2024-local} introduced the GlobalRG benchmark, which comprises two tasks: retrieving culturally diverse images for universal concepts from 50 countries and grounding culture-specific concepts within images from 15 countries.
\citet{karamolegkou-etal-2024-vision} proposed a culture-centric evaluation benchmark investigating the reliability of LVLMs as visual assistants for blind people in a culturally diverse setting. 
Using the CulturalVQA \cite{nayak-etal-2024-benchmarking}, the authors assessed geo-diverse cultural understanding of nine ``1st-Gen'' LVLMs on a curated dataset of 2,378 VQA pairs representing cultures from 11 countries and five cultural aspects.
CulturalBench~\cite{chiu2024culturalbench} is a dataset of 1,227 human-written and human-verified questions for evaluating LLMs' cultural knowledge, covering 45 global ``regions''.
\citet{nikandrou2024crope} propose CROPE, a VQA benchmark designed to probe the knowledge of culture-specific concepts and evaluate the capacity for cultural adaptation through contextual information featuring over 1M data points across 30 languages and dialects.
See Table~\ref{tab:rw:compare} for an overview and a comparison or related work with \dsname.

\rparagraph{Multilingual Multicultural LVLM Benchmarks}
Several studies evaluate the cultural awareness and capabilities of LVLMs in a multilingual setting.
\citet{geigle2025centurio} extensively benchmarked state-of-the-art LVLMs across multiple multilingual and multicultural datasets, including MaRVL~\cite{liu2021marvl}, XM3600~\cite{thapliyal_xm3600_2022} and MaXM\cite{changpinyo-etal-2023-maxm}, M5B-VGR and M5B-VLOD~\cite{schneider-sitaram-2024-m5}, CVQA~\cite{romero2024cvqa}
\citet{winata2024worldcuisines} created WorldCuisines, a large-scale benchmark for multilingual and multicultural VQA on global cuisines.
However, in \dsname, we focus on the English language, considering English performance as an upper bound.
%


%% file: src/030_gimmick_overview.tex
\section{The \dsname Benchmark}
\label{sec:benchmark}
%
%
\paragraph{Cultural Benchmark Positioning}
\citet{adilazuarda-etal-2024-towards} surveyed 90+ recent papers on cultural awareness in LLMs and found that \emph{none} explicitly define ``culture''.
Instead, these studies evaluate models on datasets capturing only specific cultural aspects, which the authors organize into two dimensions: \textit{demographic} and \textit{semantic} proxies (with seven and five subsets, respectively).
In \dsname, we adopt the proposed taxonomy by using countries and regions as \textit{demographic} cultural proxies.
Our tasks span all five \textit{semantic} proxies: ``emotions and values'', ``food and drink'', ``social and political relations'', ``basic actions and technology'', and ``names''.
We implement primarily ``black-box'' generative and discriminative probing approaches.

\rparagraph{UNESCO Intangible Cultural Heritage}
\label{sec:benchmark:datasource}
All tasks in \dsname are based on high-quality open-access data from the UNESCO Intangible Cultural Heritage (ICH) project\footnote{\url{https://ich.unesco.org}}, which aims to safeguard cultural traditions and practices vital to the identity and heritage of communities worldwide while honoring cultural diversity.
Intangible cultural heritage encompasses oral traditions, performing arts, rituals, festive events, traditional craftsmanship, and cultural knowledge.
The open-access dataset is structured as a knowledge graph, where most nodes represent cultural events or facets (CEFs; e.g., \emph{Yuki-tsumugi}, a silk fabric production technique from Japan\footnote{More examples including images are shown in \S\ref{appendix:sec:benchmark:cef:examples}}), with additional nodes including countries, regions, case studies in which the CEFs occur.
%
%
%
For \dsname, we extract the CEFs, each together with their title, description, associated macro-regions and countries, and several images depicting different aspects of the CEF.
Moreover, each CEF is detailed in one or more YouTube videos.
In total, \dsname contains 728 CEFs from 144 countries represented by 6,887 images and 993 videos\footnote{We provide licensing details in \S\ref{appedix:sec:benchmark:license}}.
While most CEFs ($88.60\%$) are associated with one country, some are associated with two or more countries.
The UNESCO ICH project groups the countries into six global macro-regions\footnote{We provide a comprehensive list in Table~\ref{tab:benchmark:regions_full} in \S\ref{appendix:sec:benchmark:regions}}, which we adopt in this work. Throughout the paper---including all figures and tables---we use the region abbreviations listed in Table~\ref{tab:benchmark:regions}.
\begin{table}[t]
    \centering
    \footnotesize
    \renewcommand{\arraystretch}{.85}
    \resizebox{\linewidth}{!}{%
    \begin{tabular}{llrr}
    \toprule
    \textsc{Region} & \textsc{Abbrv.} & \textsc{\#C} & \textsc{\#CEF} \\
    \midrule
    Arab & \RegA & 18 &  76 \\
    Asia \& Pacific & \RegAP & 35 & 226 \\
    Eastern Europe & \RegE & 25 &  150 \\
    Latin-America \& Caribbean & \RegLAC & 28 & 98  \\
    Subsaharian Africa & \RegSA & 40 & 73 \\
    Western Europe \& North America & \RegW & 23 & 149 \\
    \midrule
    \multicolumn{2}{l}{\textit{Unique}} & 144 & 728 \\
    \bottomrule
    \end{tabular}
    }
    \caption{Regions within \dsname. \textsc{\#C} and \textsc{\#CEF} stand for the number of Countries and CEFs related to the respective region. Some CEFs may span multiple regions.}
    \label{tab:benchmark:regions}
\end{table}
%


\subsection{Datasets and Tasks}
\label{sec:benchmark:datasets}
We created three novel multimodal datasets that serve as the foundation for six tasks designed to evaluate the cultural knowledge of models.
See Figure~\ref{fig:figure1} for an overview of the different tasks.\footnote{Sample counts per task \& region are shown in \S\ref{appendix:sec:benchmark:samples}}
%

%
%
\input{src/031_sivqa}
\input{src/032_vvqa}
\input{src/033_coqa}
\input{src/034_ckqa}

%% file: src/031_sivqa.tex
\subsection{Cultural Image VQA}
\label{sec:sivqa}
In the Cultural Image VQA (\sivqa) task, models are presented with an image depicting a CEF and a question that relates to a particular CEF aspect (see \S\ref{appendix:sec:sivqa:examples} for examples).
Models are evaluated based on answer correctness.
To create the data for \sivqa, we couple synthetic data generation with a two-stage annotation process.

\rparagraph{Synthetic Data Generation}
\label{sec:sivqa:collection}
Building on the high-quality UNESCO ICH data, we applied synthetic data generation by prompting GPT-4o\footnote{\texttt{gpt-4o-2024-08-06}} to construct the basis for our dataset.
Each VQA pair is related to a CEF and consists of an image depicting one aspect of the CEF, a question related to the CEF and the image, and an answer.
Maximizing the quality of the generated silver data, we applied extensive prompt engineering combining techniques such as Few-Shot, Chain-of-Thought, ReAct~\cite{wei2022cot,zhang2023autocot,zheng2024react,sahoo2024promptsurvey} to craft the prompt.
Key aspects of the prompt are a role description, a general task description, detailed annotation guidelines, a step-by-step strategy, an expected output format, few-shot examples, and the information of the target CEF (see \S\ref{appendix:sec:sivqa:synth} for the full prompt).
We then generated silver VQA pairs for each of the 6,827 images contained in the ICH data source, which resulted in 17,369 pairs.
Afterward, we automatically removed pairs where 1) the question contained words that introduce subjectiveness or ambiguity (``\textit{could}'', ``\textit{should}'', ``\textit{maybe}'', etc.); 2) the answer contained abstract words that are hard to depict visually; and 3) where the answer is not a substring of the description of the related CEF.
This way, we obtained 9,900 silver VQA samples related to 5,517 images from all 728 CEFs.

\rparagraph{Annotation Process}
\label{sec:sivqa:collection:annotation}
Opting for high-quality VQA pairs as well as cultural diversity, we devised a two-stage annotation process with 18 trained experts from various cultural backgrounds covering all six regions (see Table~\ref{tab:sivqa:anno:demographics} in \S\ref{appendix:sec:sivqa:anno}).
Each silver pair was evaluated using two questionnaires---one with seven question-related requirements and another with four answer-related requirements.
Questions had to target the CEF and image content directly, require cultural knowledge, and depend on visual evidence \cite{chen2024mmstar}.
Answers needed to be clear, objective, concise, and depictable.
For details on the annotation process, see \S\ref{appendix:sec:sivqa:anno}.

In the first round, we annotated each sample once, resulting in 4,114 samples, of which 2,826 (68.69\%) met all criteria.
In the second round, five annotators re-evaluated these, retaining only samples with concordant approval.
This process finally yielded 2,233 samples for 1,928 images from 728 CEFs across 144 countries in six global regions.

%% file: src/032_vvqa.tex
\subsection{Cultural Video VQA}
\label{sec:vvqa}
In this task, models are evaluated on questions relating to videos instead of single images, again employing accuracy as the metric.
To this end, we extend \sivqa in two steps: synthetic data generation and quality annotation.

\rparagraph{Synthetic Data Generation}
\label{sec:vvqa:collection}
First, we adjusted the \sivqa questions by replacing the term  \emph{``image''} with \emph{``video''}. We then coupled the question with a short video clip, for which we started from the CEF's associated YouTube video. We ensured that the shortened clip contains relevant information for answering the question as follows:
From each video, we extracted one frame per second, and computed image embeddings for both the frames and the \sivqa image, using DINOv2\footnote{\texttt{facebook/dinov2-with-registers-large}}~\cite{oquab2024dinov2,darcet2024dinov2registers}.
We then identified the frame that best matches the original image by calculating Cosine similarity. We selected this frame as the center (at $t=0$) for a 10-second clip\footnote{
We do not include the audio stream in our clips.} (from $t=-5$ to $t=5$).
We only include clips with a best-matching frame similarity $>0.5$, which we found to yield high-quality instances based on a manual inspection of random samples.
Overall, this procedure resulted in 2,001 silver samples.

\rparagraph{Annotation Process}
\label{sec:vvqa:collection:annotation}
For additional quality control, a trained expert annotated 20\% of the silver data (400 samples).
Each sample was evaluated using a three-item questionnaire\footnote{cf. \S\ref{appendix:sec:vvqa} for details.} assessing whether (1) the video contained frames resembling the CEF image, (2) it clearly answered the question, or (3) neither condition was met.
Overall, 95\% of the annotated samples were accepted.
For closer inspection, we stratified the annotated samples into four similarity bins, revealing that roughly 10\% of those in the lower bins ($[0.5, 0.75[)$ were rejected, while nearly all, i.e., 99\% and 100\%, in the higher bins ($[0.75, 1.0]$) were retained.
The residual 5\% label noise was considered acceptable based on further manual analysis.
Notably, we found that of the 20 rejected samples, only 9 were unanswerable based on the video, while the remaining 11 exhibited only a suboptimal frame match w.r.t. the \sivqa image.
The final \dsname \vvqa dataset contains 1,809 samples (see \S\ref{appendix:sec:vvqa:examples} for examples) linked to 553 CEFs from 139 countries.

%% file: src/033_coqa.tex
\subsection{Cultural Origin QA}
\label{sec:coqa}
With Cultural Origin QA (\coqa), we test a model's ability to capture coarse-grained cultural knowledge.
Given a CEF's images, title, or both, the models must select its cultural origin (multiple-choice).
We refer to the task as \coqar when the origin is a region and as \coqac when it is a country.

\rparagraph{Dataset Construction}
\label{sec:coqa:collection}
The \coqa dataset contains all 728 CEFs from UNESCO ICH.
To ensure that each instance corresponds to a unique origin, we replicate each CEF $N$ times—where $N$ represents the number of associated regions (for \coqar) or countries (for \coqac).
For \coqar, three negatives are randomly sampled from the remaining pool.
%
%
Negatives for \coqac drawn from those within the same region as the target country.

\rparagraph{Input Modalities and Prompts}
\label{sec:coqa:config}
The \coqa tasks support multiple input configurations alongside the task prompt.
In the text-only setting, only the title of the CEF is provided, whereas in the ``image-only'' setting, \emph{all} images associated with the CEF are included.
Both the title and the images are used in the text-image setting.
Examples and complete prompts for all variations are shown in \S\ref{appendix:sec:coqa:examples}.

%% file: src/034_ckqa.tex
\subsection{Cultural Knowledge QA}
\label{sec:ckqa}
In \dsname Cultural Knowledge QA (\ckqa), we evaluate whether current AI models capture fine-grained cultural knowledge.
%
%
The dataset supports two open-answer tasks: naming (\ckqan) and describing (\ckqad).
For \ckqan, the ground truth corresponds to the title of the CEF, while for \ckqad, it is the detailed description.
For both tasks, we leverage all 728 CEFs from UNESCO ICH.
As with \coqa, \ckqa supports multiple input configurations: text-only, ``image-only'', and text+image.
We provide examples and prompts for all variations in \S\ref{appendix:sec:ckqa:prompts}.

%% file: src/040_analyses.tex
\section{Experimental Setup}
\label{sec:exsetup}
\rparagraph{Models and Inference}
\label{sec:benchmark:models}
We evaluate a total of 31 models, including five proprietary LVLMs, 15 open-weight LVLMs, and 11 open-weight LLMs---the backbones of the respective LVLMs---covering 9 LVLM and 4 LLM model families.
The sizes of the open-weight models vary, categorized as small, medium, large, and extra-large (see Table~\ref{tab:benchmark:model_groups}).
A comprehensive list of models is provided in Table~\ref{tab:benchmark:models} in \S\ref{appendix:sec:benchmark:models}.
\begin{table}[t]
    \centering
    \footnotesize
    \renewcommand{\arraystretch}{.8}
    \begin{tabular}{l l r r}
        \toprule
        \textsc{Group} & \textsc{Parameters (B)}  & \textsc{LLMs} & \textsc{LVLMs} \\
        \midrule
        S  & 0.5 -- 4 & 5 & 5 \\
        M  & 7 -- 11 & 3 & 6 \\
        L  & 26 -- 38 & 2 & 2 \\
        XL & 72 -- 78 & 1 & 2 \\
        Closed  & unkown & 0 & 5 \\
        \midrule
        \multicolumn{2}{l}{\textit{Total}} & 11 & 20 \\
        \bottomrule
    \end{tabular}
    \caption{The size groups we define for result aggregation according to models' number of parameters.}
    \label{tab:benchmark:model_groups}
\end{table}
%
%
For our experiments, we download open weights from the respective Huggingface~\cite{wolf2019hftransformers} repositories (see Table~\ref{tab:benchmark:models}) and generate responses employing greedy decoding.
For proprietary models, we use the official Python SDKs.
More details are reported in \S\ref{appendix:sec:setup}.

\rparagraph{Metrics}
%
%
For the \sivqa, \vvqa, and \coqa tasks, we report relaxed answer accuracy, for which we consider a generated answer correct if it starts with the ground truth answer.
For \ckqad and \ckqan, due to their generative nature, we use GPT-4o\footnote{\texttt{gpt-4o-2024-11-20}} in an ``LVLM-as-a-Judge''~\cite{zheng2023llm-as-a-judge, xiong2024llava-critic} setup to judge responses with a score $s \in [0, 100]$.
Where $s=0$, $s=50$, and $s=100$ indicate \textit{completely incorrect or irrelevant}, \textit{partially correct or relevant}, and \textit{perfectly correct and complete} answers, respectively.
%
%

\rparagraph{Video Processing}
The 10-second video clips from \vvqa do not contain an audio stream, and we only use the visual information.
Following established praxis~\cite[e.g.,][]{wang2024qwen2vl}, we extract one frame per second from the videos and provide them to the models as input alongside the textual prompt.
Specifics about the image and video processing of the individual models are documented in the code.

\section{Results and Analyses}
\label{sec:analyses}
In this section, we present a series of in-depth analyses based on the outcomes of our benchmark.
We show aggregated results: open-weight models are grouped and averaged by parameter size, and proprietary models are averaged together (see Table~\ref{tab:benchmark:model_groups}).
We provide the complete numerical results for all tasks and models in \S~\ref{appendix:sec:analyses}.
In the following, we use abbreviations for regions‚ as defined in Table~\ref{tab:benchmark:regions}.

\input{src/041_analyses_a1}
%
\input{src/042_analyses_a2}
%
\input{src/043_analyses_a3}
\input{src/044_analyses_a4}
%


%% file: src/041_analyses_a1.tex
\subsection{General Trends and Cultural Bias}
\label{sec:analyses:a1_bias}
\begin{figure}[t]
     \begin{subfigure}{1.\linewidth}
        \centering
        \includegraphics[width=1.\linewidth]{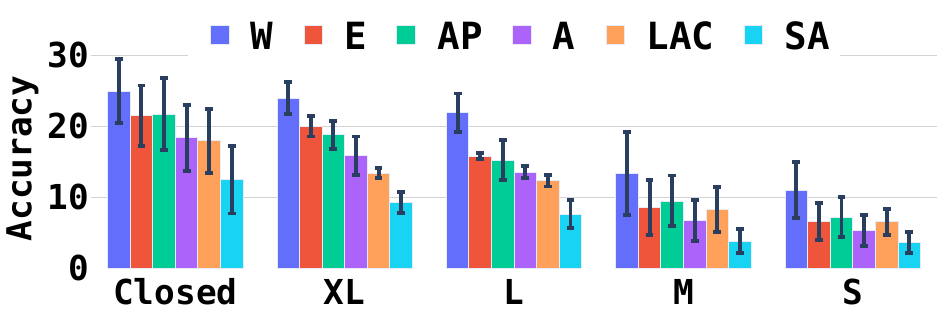}
        \caption{\sivqa Accuracy}
        \label{fig:analyses:a1_bias:scores:sivqa}
    \end{subfigure}

    \begin{subfigure}{1.\linewidth} 
        \centering
        \includegraphics[width=1.\linewidth, trim=0 0 0 31.5, clip]{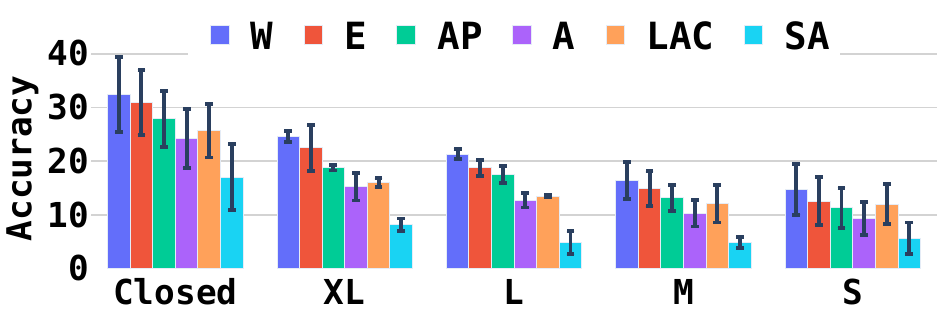}
        \caption{\vvqa Accuracy}
        \label{fig:analyses:a1_bias:scores:vvqa}
    \end{subfigure}

    \begin{subfigure}{1.\linewidth}
        \centering
        \includegraphics[width=1.\linewidth, trim=0 0 0 25, clip]{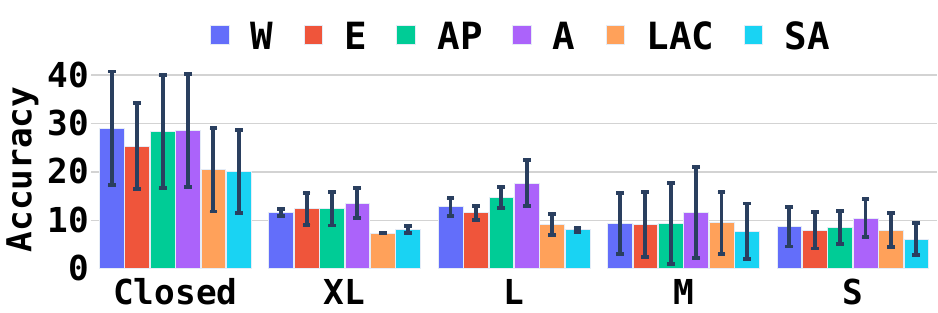}
        \caption{\ckqan Accuracy}
        \label{fig:analyses:a1_bias:ckqa-name}
    \end{subfigure}
    \caption{Aggregated results of the VQA tasks.}
    \label{fig:analyses:a1_bias:scores}
\end{figure}

\begin{figure}[t]
    \centering
    \includegraphics[width=0.8\linewidth]{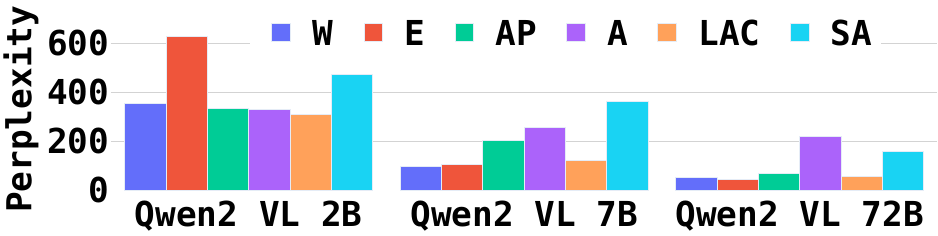}
    \caption{\sivqa ground-truth answer perplexity.}
    \label{fig:analyses:a1_bias:ppl}
\end{figure}

We discuss general trends and investigate cultural bias across regions (Figures~\ref{fig:analyses:a1_bias:scores} and~\ref{fig:analyses:a1_bias:ppl}). 


\rrparagraph{\sivqa \& \vvqa}
Figures~\ref{fig:analyses:a1_bias:scores}a–c show clear regional performance disparities.
Across all models---proprietary and open-weight, regardless of size---scores are highest for Western and Asian targets (\RegW, \RegE, and \RegAP) and lowest for \RegSA.
XL models, e.g., reach $24.04$ on \RegW and $9.32$ on \RegSA on average.
\RegA and \RegLAC fall in between, with model performance varying by size.
Since \sivqa is an open-answer task, often with rare culturally specific terms, we also evaluated the task with GPT-4o as LVLM-as-a-Judge to account for imperfect naming or spelling.
While this method yields higher scores, it confirms the same trend: models exhibit a strong bias toward Western contexts.
However, even the best model (\m{GPT-4o}) scores only $31.58$\% on \RegW and $25.44$\% on average, highlighting \dsname as a challenging benchmark and the lack of fine-grained cultural knowledge in current models.
We supplement our analysis with a more fine-grained investigation of how well models ``know'' the cultural concepts discussed.
Here, we focus on the \m{QwenVL} models on \sivqa and the compute perplexity of ground truth answers (conditioned on the input context) as a proxy of model cultural knowledge (details in \S\ref{appendix:sec:analyses:sivqa:results:ppl}). 
Figure~\ref{fig:analyses:a1_bias:ppl} shows that for the 7B and 72B models, perplexity is consistently lower for \RegW, \RegE, and \RegAP compared to \RegA and \RegSA, aligning with our performance findings.
For the 2B model, however, \RegE and \RegSA yield the highest perplexities, which we attribute to the overall brittleness of the model.
Moreover, we revisit the performance on questions about the prevalent cultural aspects in \sivqa (details in \S\ref{sec:appendix:sec:analyses:sivqa:results:aspects}) and find that models perform notably better on tangible cultural aspects than on intangible ones.
For instance, closed models achieve an accuracy of 30\% for food-related questions and only 8\% and 10\% for questions concerning rituals or festivals.
This highlights biases along the cultural dimension, which are particularly pronounced in non-Western contexts.
%

%
%

\rrparagraph{\ckqan \& \ckqad}
For \ckqan, regional differences are minor, though proprietary models significantly outperform open-weight ones (see Figure~\ref{fig:analyses:a1_bias:ckqa-name}).
The large error bars for closed models indicate inconsistent performance---particularly from \m{GPT-4o Mini} and \m{Gemini Flash} models, which perform similarly to large open-weight models.
XL and L models perform worst on \RegSA and \RegLAC and best on \RegA and \RegAP with minor differences to \RegW and \RegE.
For \ckqad (Figure~\ref{fig:analyses:a3_modality:ckqa-desc}), performance is $10–20\%$ higher than on \ckqan, likely because describing a CEF is easier than exactly naming it.
However, regional biases are larger, with consistently higher scores on \RegW than on \RegSA, primarily for closed models like \m{GPT-4o}, which reaches $53.66$ for \RegW and $43.70$ on \RegSA.

\rrparagraph{\coqac \& \coqar}
Figure~\ref{fig:analyses:a3_modality:coqa-countries} shows minimal regional differences for \coqac.
Average accuracies range from close to or above $90\%$ for closed, XL, and L models to $77.42$\% for S models.
However, performance on \coqar is lower than on \coqac---$85.02\%$ vs. $81.17\%$ on average over all models and regions--- with models achieving the highest scores in \RegAP.
Notably, the regional ranking is mostly inverted compared to other tasks---\RegSA, \RegA, \RegLAC, \RegE, and \RegAP score higher than \RegW---suggesting more distinct visual and linguistic features in non-Western regions.
%

%
%
%
%
%
%
%

%% file: src/042_analyses_a2.tex
\subsection{Influence of Model Size}
\label{sec:analyses:a2_model}
%
We assess how model size impacts performance and whether it affects regions equally.

\begin{figure}[t]
    \centering
    \begin{subfigure}{0.30\linewidth}
        \centering
        \includegraphics[width=\linewidth]{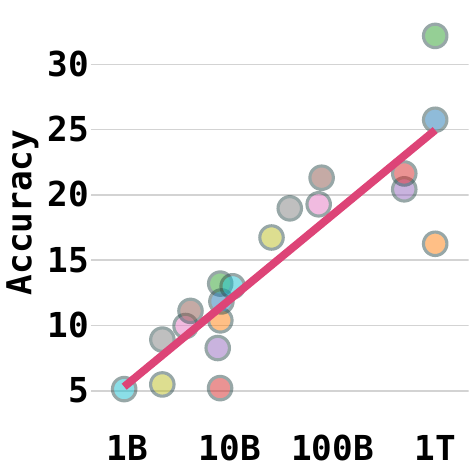}
        \caption{\sivqa, \tiny{$r$=$0.62$\textsuperscript{*}}}
    \end{subfigure}
    \begin{subfigure}{0.30\linewidth}
        \centering
        \includegraphics[width=\linewidth]{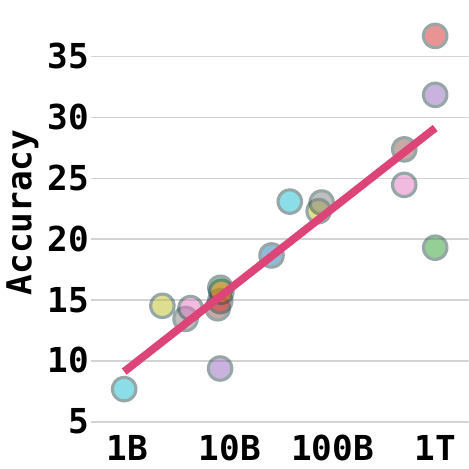}
        \caption{\vvqa, \tiny{$r$=$0.36$\textsuperscript{*}}}
    \end{subfigure}
    \begin{subfigure}{0.30\linewidth}
        \centering
        \includegraphics[width=\linewidth]{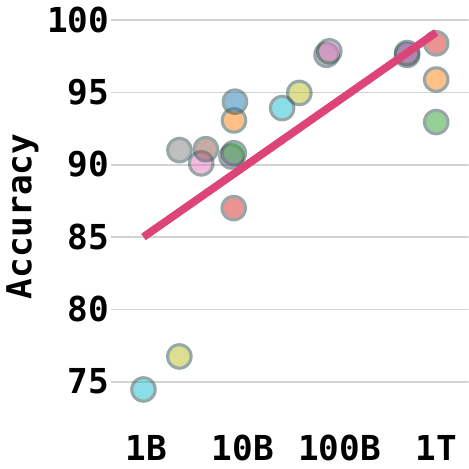}
        \caption{\coqac, \tiny{$r$=$0.39$\textsuperscript{*}}}
    \end{subfigure}

    \begin{subfigure}{0.30\linewidth}
        \centering
        \includegraphics[width=\linewidth]{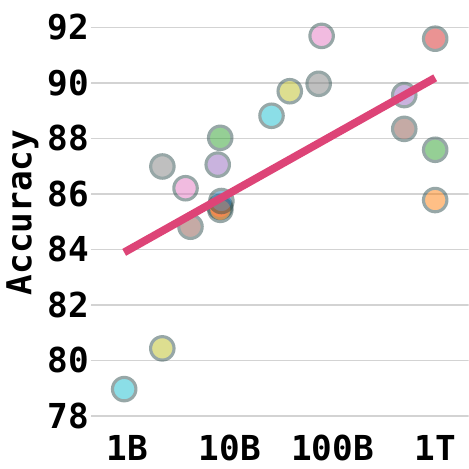}
        \caption{\coqar, \tiny{$r$=$0.16$\textsuperscript{*}}}
    \end{subfigure}
    \begin{subfigure}{0.30\linewidth}
        \centering
        \includegraphics[width=\linewidth]{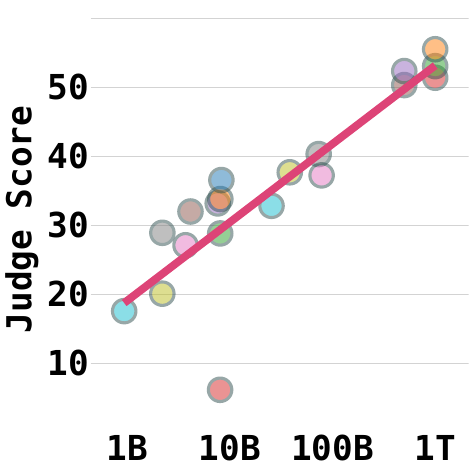}
        \caption{\ckqad, \tiny{$r$=$0.46$}\textsuperscript{*}}
    \end{subfigure}
    \begin{subfigure}{0.30\linewidth}
        \centering
        \includegraphics[width=\linewidth]{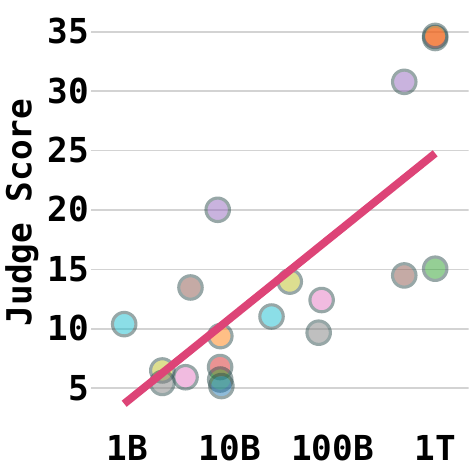}
        \caption{\ckqan, \tiny{$r$=$0.39$}\textsuperscript{*}}
    \end{subfigure}
    \caption{Model size vs. performance on \dsname tasks. The x-axis is in log scale. The trend line was computed using OLS regression. We report the Pearson correlation coefficient $r$ ( \textsuperscript{*} indicates statistical significance).}
    \label{fig:sec:analyses:a2_model:size}
\end{figure}
\begin{figure}[t]
    \centering
    \includegraphics[width=1.\linewidth]{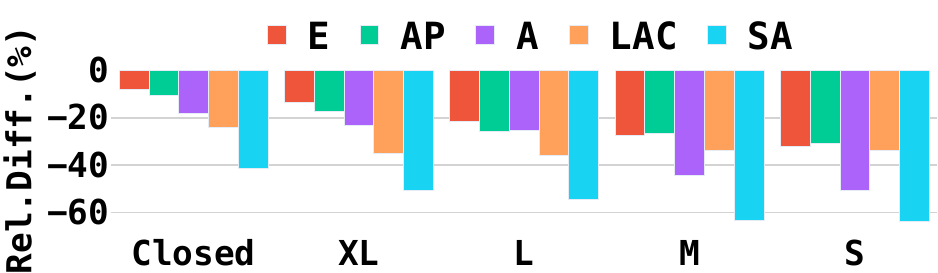}
    \caption{Relative Difference to \RegW for \sivqa.}
    \label{fig:sec:analyses:a2_model:sivqa_relative_diffs_to_w}
\end{figure}

Figure~\ref{fig:sec:analyses:a2_model:size} shows that model size\footnote{For closed source models, we manually set the number of parameters to 1T, except for Gemini Flash and GPT-4o mini, for which we set the number to 500B.} significantly influences performance, with moderate to strong Pearson correlations and steep regression lines across tasks except \coqar, where the effect is minimal.
Figure~\ref{fig:sec:analyses:a2_model:sivqa_relative_diffs_to_w} shows that relative performance declines from the best-performing region (\RegW) to others, particularly \RegSA, varying by model size: the drops are $-63.39$ (S), $-63.85$ (M), $-50.60$ (L), $-54.57$ (XL), and $-41.52$ (Closed). We conclude that bigger sizes tend to result in smaller gaps without size presenting a strict ordering criterion.
%


%% file: src/043_analyses_a3.tex
\subsection{Influence of Modalities}
\label{sec:analyses:a3_modality}
We explore how input modality---text-only, image-only, or text+image---affects performance on \coqac, \coqar, and \ckqad.
Further, we compare LVLMs to their LLM backbones to assess potential losses in cultural knowledge during multimodal training.

\begin{figure*}[t]
    \centering
     \begin{subfigure}{1.\linewidth}
        \centering
        \includegraphics[width=1.\linewidth]{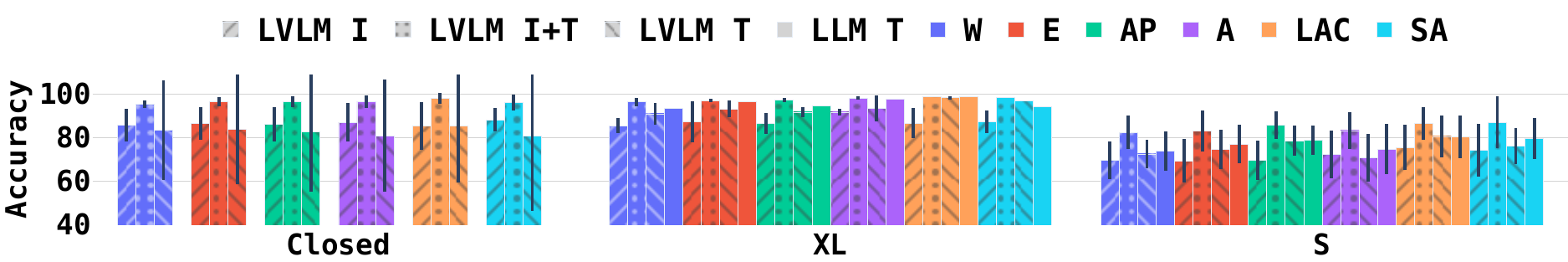}
        \caption{\coqac}
        \label{fig:analyses:a3_modality:coqa-countries}
    \end{subfigure}
    
     \begin{subfigure}{1.\linewidth}
        \centering
        \includegraphics[width=1.\linewidth, trim=0 0 0 25, clip]{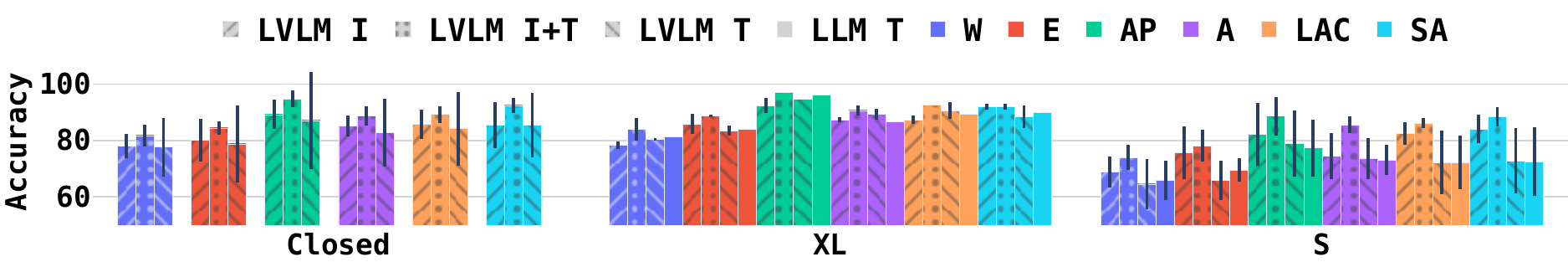}
        \caption{\coqar}
        \label{fig:analyses:a3_modality:coqa-regions}
    \end{subfigure}

    \begin{subfigure}{1.\linewidth}
        \centering
        \includegraphics[width=1.\linewidth, trim=0 0 0 25, clip]{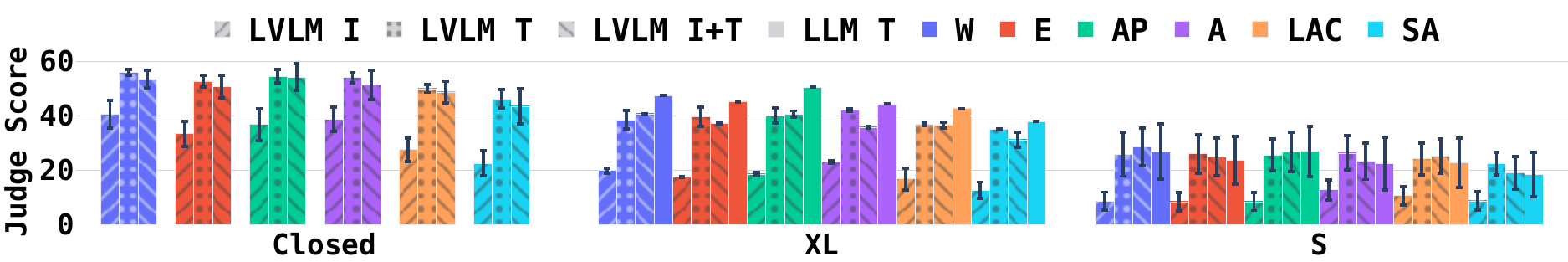}
        \caption{\ckqad}
        \label{fig:analyses:a3_modality:ckqa-desc}
    \end{subfigure}
    \caption{Aggregated results including multimodal input variations: \textbf{T}ext-only, \textbf{I}mage-only, \textbf{T}ext+\textbf{I}mage.}
    \label{fig:analyses:a3_modality:scores}
\end{figure*}
\rparagraph{Input Modalities}
Figure~\ref{fig:analyses:a3_modality:scores} shows that text+image (\texttt{I+T}) inputs consistently yield the highest performance across all tasks, confirming that textual and visual data provide complementary cultural cues.
The gap between \texttt{I+T} and text-only (\texttt{T}) is slightly more prominent for \coqac than \coqar, suggesting that visual information aids in inferring fine-grained, country-level details.
In contrast, image-only (\texttt{I}) inputs perform poorly, indicating that textual information, such as CEF titles, carries more cultural context.
The high variance in \texttt{T} results for the \coqa tasks stems from the performance disparity between \m{Gemini Pro} and \m{Claude 3.5 Sonnet} (e.g., $59.38$ vs. $83.75$ for \RegW).

\rparagraph{LVLM vs. LLM-Backbone}
Comparing LVLMs with their LLM backbones reveals that multimodal training can impair the acquisition of detailed cultural knowledge (notably in \ckqad) while having minimal impact on coarse-grained cultural understanding (\coqa).
For large models, significant performance gaps---$50.62$ for \m{Qwen2.5 72B} vs. $40.02$ for \m{Qwen2VL 72B} on \RegAP---on the \ckqad task between the LVLMS and their LLM backbones can be observed, whereas, for smaller models, the effect is subtle.
Overall, our findings highlight that while images complement text for culturally grounded tasks, it is ultimately the synergy between both modalities that leads to robust and broad cultural understanding.

%% file: src/044_analyses_a4.tex
\subsection{Influence of External Cues}
\label{sec:analyses:a4_external}
We examine how external hints, i.e., informing a model about the country or region of a CEF, affect VQA performance.
\begin{figure*}[t]
    \centering
    \begin{subfigure}{1.\linewidth}
        \centering
        \includegraphics[width=1.\linewidth]{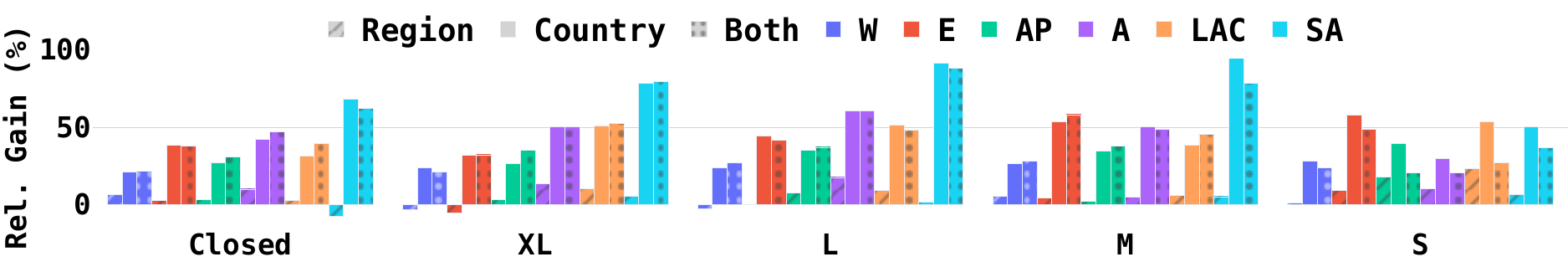}
        \caption{\texttt{SIVQA}}
        \label{fig:analyses:a4:relative:sivqa}
    \end{subfigure}
    
    \begin{subfigure}{1.\linewidth}
        \centering
        \includegraphics[width=1.\linewidth, trim=0 0 0 0, clip]{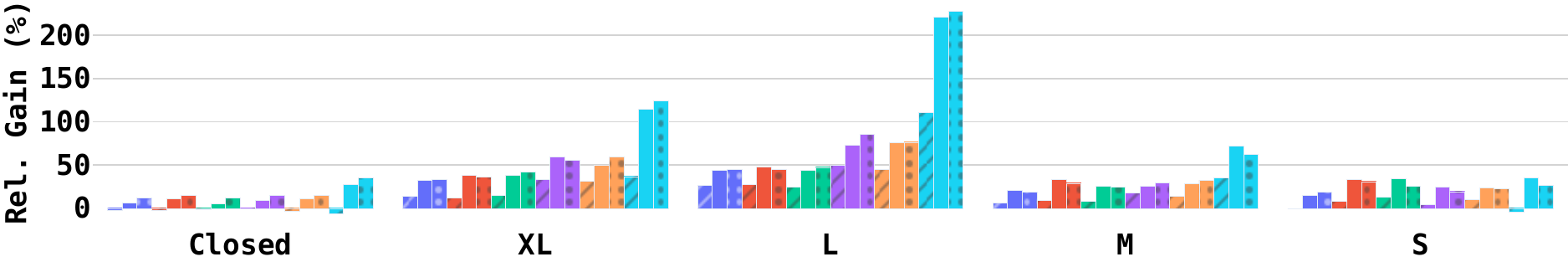}
        \caption{\texttt{VVQA}}
        \label{fig:analyses:a4:relative:vvqa}
    \end{subfigure}
    \caption{Relative gains on VQA tasks from providing external geographical hints.}
    \label{fig:analyses:a4:relative}
\end{figure*}
For \sivqa (Figure~\ref{fig:analyses:a4:relative:sivqa}), country hints consistently boost performance across model sizes and regions, while regional cues yield only modest—or even slightly adverse—effects in larger models.
Gains from country hints are around 50\% for most regions, but in \RegSA, improvements nearly double (e.g., $97.48\%$ for \m{InternVL 2.5 78B} and $97.13\%$ for \m{InternVL 2.5 38B}).
A similar pattern emerges for \vvqa (Figure~\ref{fig:analyses:a4:relative:vvqa}).
Hints generally enhance performance across regions and models, with \RegSA showing the most significant gains.
Proprietary and small models exhibit subtle improvements, whereas L and XL models see much higher relative gains---up to $240.7\%$ for \m{Intern VL 38B}.
Notably, regional cues have a more positive impact on \vvqa than on \sivqa.
%

%% file: src/880_conclusion.tex
\section{Conclusion}
\label{sec:conclusion}
We introduce \dsname, a comprehensive benchmark to assess various aspects of cultural knowledge of current LVLMs and LLMs and introduce six tasks built upon three novel datasets, which span 728 unique cultural events or facets (CEFs) from 144 countries grouped into six global macro-regions.
Through extensive analyses, we study general cultural biases and the influence of model size, input modalities, and external cues.
Our results consistently reveal a prominent bias toward Western cultures across all models.
Interestingly, when only coarse cultural knowledge is required---such as regional origins---models performed remarkably better.
Across all tasks, significant correlations between a model's performance and its size are evident, with a substantial gap between proprietary and open-weight models.
Our analyses show that while models grasp broad cultural categories, they struggle with nuanced understanding.
This suggests that \dsname poses a challenging benchmark and highlights the need for further advances in modeling broad cultural awareness.

%% file: src/881_limitations.tex
\newpage
\section*{Limitations}
\label{sec:limitations}
\paragraph{English-Only Benchmark}
Although we consider the performance on tasks requiring cultural understanding in English as an upper bound for the majority of models, it is yet to be tested if that hypothesis generally holds across tasks, model size, and model family.
Especially for models like \m{QwenVL} and \m{InternVL}, which were pretrained on large portions of Chinese textual data, Chinese could be pivotal instead of English.
Moreover, some cultural nuances might not be translatable to other languages.

\rparagraph{Open-Ended VQA}
\sivqa and \vvqa comprise open-ended answers to their questions, imposing challenges for adequate evaluation, especially when employing binary metrics like accuracy.
This is especially true for rare, culturally specific answer terms, such as in our tasks, which are prone to spelling inaccuracies or might have different names in different cultures or languages.
Although we alleviate this issue by computing scores using GPT-4o in an LVLM-as-a-Judge setting and thereby confirm our findings, this requires additional computational and financial resources.
A typical solution for this is transforming the questions into multiple choice questions, which, however, requires culturally expert annotators, which are complicated to find or train and expensive if hired via professional annotation companies.

\rparagraph{Small Number of Samples}
With a total of 7239 unique samples across all tasks in \dsname---2233 (\sivqa), 1809 (\vvqa), 982 (\coqac), 759 (\coqar), 728 (\ckqad), and 728 (\ckqan)---, the benchmark itself as the third most samples compared to other recent benchmarks.
However, the per-task number falls relatively low, leading to even fewer counts per country or culture, making judgments about single countries not informative.

%% file: src/882_ethical.tex
\section*{Ethical Considerations}
\label{sec:ethical}
\rparagraph{Country and Region Definitions}
\dsname adopts the country and region classifications from the UNESCO ICH dataset.
While these classifications are widely used, we recognize the potential for differing interpretations.

\rparagraph{Potentially Offensive Questions}
We employed semi-automatic data generation strategies to create the \sivqa dataset.
Here, the silver data was generated using GPT-4o, which we showed displays significant cultural biases towards Western contexts.
Although we provided the model with high-quality ground-truth information from the UNESCO ICH project and trained expert annotators with diverse cultural backgrounds to filter low-quality VQA samples, certain questions or their answers might still be offensive to people with certain cultural origins.
Since this is subjective, we need to accept it as is for now.
Nevertheless, we encourage contacting us if any offensive or otherwise harmful sample raises someone's attention.

%% file: src/883_ack.tex
\section*{Acknowledgements}
\label{sec:ack}
We thank our annotators for the \sivqa and \vvqa tasks with special thanks to Timm Dill, Narges Baba Ahmadi, Niloufar Baba Ahmadi, and Abdullah Abdelhafez for their extra efforts.
The work of Carolin Holtermann and Anne Lauscher is funded by the Excellence Strategy of the German Federal Government and the Federal States.

%% file: src/990_appendix.tex
\newpage
\appendix
\input{src/991_0_appendix_benchmark}
\input{src/992_0_appendix_results}

%% file: src/991_0_appendix_benchmark.tex
\newpage
\onecolumn
\section{\dsname Benchmark Details}
\label{appendix:sec:benchmark}

\subsection{Data License}
\label{appedix:sec:benchmark:license}
\dsname is built upon the open-access data from the UNESCO Intangible Cultural Heritage (ICH) project, which is organized as a knowledge graph.
The graph can be downloaded in English, French, and Spanish on the ICH project website: \href{https://ich.unesco.org/en/open-access-to-dive-data-01218}{https://ich.unesco.org/en/open-access-to-dive-data-01218}, with details about its structure and subsets also provided.
In \dsname, we work with the English graph only.
The open-access license of the knowledge graph is defined on the UNESCO website\footnote{\href{https://www.unesco.org/en/open-access}{https://www.unesco.org/en/open-access}} as follows:
\begin{displayquote}
\textit{By 'open access' to the literature, we mean its free availability on the public internet, permitting any users to read, download, copy, distribute, print, search, or link to the full texts of these articles, crawl them for indexing, pass them as data to software, or use them for any other lawful purpose, without financial, legal, or technical barriers other than those inseparable from gaining access to the internet itself.}
\end{displayquote}

\noindent
The images and videos within the data are shared via URLs and hosted by UNESCO or on YouTube, respectively.
Further, each image and video node in the knowledge graph has individual copyright information attached.
However, the licenses themselves are not discussed, and merely the name of the photographer or institution or UNESCO itself is stated.
Unfortunately, we did not receive an answer to multiple emails in which we asked for clarification.
Hence, we assume that the image and video content also fall under the definition of ''open access''.
If you are a copyright holder of any of the images or videos and do not want your intellectual property to be used or shared by us, please reach out via email: \href{mailto:florian.schneider-1@uni-hamburg.de}{florian.schneider-1@uni-hamburg.de}.

\subsection{Cultural Event or Facets (CEFs)}
\label{appendix:sec:benchmark:cef}
\subsubsection{Examples}
\label{appendix:sec:benchmark:cef:examples}
In the following, we provide one example of CEFs per region from the UNESCO ICH project.
We also use the same information for the \ckqan and \ckqad tasks. 
\subsubsection*{Western Europe (\RegW)}
\input{examples/cef/western-european-and-north-american-states_0}
\subsubsection*{Eastern Europe (\RegE)}
\input{examples/cef/eastern-european-states_0}
\subsubsection*{Arab (\RegA)}
\input{examples/cef/arab-states_0}
\subsubsection*{Asia and Pacific (\RegAP)}
\input{examples/cef/asian-and-pacific-states_0}
\subsubsection*{Latin America \& Caribbean (\RegLAC)}
\input{examples/cef/latin-american-and-caribbean-states_0}
\subsubsection*{Subsaharian Africa (\RegSA)}
\input{examples/cef/subsaharian-african-states_0}
\subsubsection{CEFs as Python a \texttt{dataclass}}
Listing~\ref{listing:benchmark:cef} presents a CEF implemented as a Python dataclass.

\begin{listing}[ht!]
\begin{minted}[fontsize=\footnotesize]{python}
from dataclasses import dataclass

@dataclass
class CEF:
    title: str
    description: str
    countries: list[str]
    regions: list[str]
    images: list[str]  # URLs
    videos: list[str]  # URLs
\end{minted}
\caption{Python pseudo-code for a dataclass representing a CEF.}
\label{listing:benchmark:cef}
\end{listing}

\subsection{Regions}
\label{appendix:sec:benchmark:regions}
\begin{table}[ht!]
    \centering
  \renewcommand{\arraystretch}{.97}
    \resizebox{\textwidth}{!}{%
    \begin{tabular}{llr p{10cm}}
    \toprule
    Region & Abbrv. & Countries & Countries \\
    \midrule
    Arab & \RegA & 18 & Algeria, Bahrain, Egypt, Iraq, Jordan, Kuwait, Lebanon, Mauritania, Morocco, Oman, Palestine, Qatar, Saudi Arabia, Sudan, Syria, Tunisia, United Arab Emirates, Yemen \\
    Asia \& Pacific & \RegAP & 35 & Lao People's Democratic Republic, Afghanistan, Australia, Bangladesh, Bhutan, Cambodia, China, Cook Islands, Democratic People’s Republic of Korea, Fiji, India, Indonesia, Iran, Japan, Kazakhstan, Korea, Kyrgyzstan, Malaysia, Micronesia, Mongolia, Myanmar, Nepal, New Zealand, Pakistan, Papua New Guinea, Philippines, Samoa, Singapore, Sri Lanka, Thailand, Timor-Leste, Tonga, Turkmenistan, Vanuatu, Vietnam \\
    Eastern Europe & \RegE & 25 & Albania, Armenia, Azerbaijan, Belarus, Bosnia and Herzegovina, Bulgaria, Croatia, Czechia, Estonia, Georgia, Hungary, Latvia, Lithuania, Moldova, Montenegro, North Macedonia, Poland, Romania, Russia, Serbia, Slovakia, Slovenia, Tajikistan, Ukraine, Uzbekistan \\
    Latin-America \& Caribbean & \RegLAC & 28 & Antigua and Barbuda, Argentina, Bahamas, Belize, Bolivia, Brazil, Chile, Colombia, Costa Rica, Cuba, Curaçao, Dominican Republic, Ecuador, El Salvador, Grenada, Guatemala, Haiti, Honduras, Jamaica, Mexico, Nicaragua, Panama, Paraguay, Peru, Saint Kitts and Nevis, Saint Vincent and the Grenadines, Uruguay, Venezuela \\
    Subsaharian Africa & \RegSA & 40 & Côte d'Ivoire, Angola, Benin, Botswana, Burkina Faso, Burundi, Cabo Verde, Cameroon, Central African Republic, Chad, Congo, Democratic Republic of the Congo, Djibouti, Eritrea, Eswatini, Ethiopia, Gabon, Gambia, Ghana, Guinea, Kenya, Lesotho, Madagascar, Malawi, Mali, Mauritius, Mozambique, Namibia, Niger, Nigeria, Rwanda, Senegal, Seychelles, Somalia, South Africa, South Sudan, Togo, Uganda, Zambia, Zimbabwe \\
    Western Europe \& North America & \RegW & 23 & Andorra, Austria, Belgium, Canada, Cyprus, Denmark, Finland, France, Germany, Greece, Iceland, Ireland, Italy, Luxembourg, Malta, Netherlands, Norway, Portugal, Spain, Sweden, Switzerland, Türkiye, United Kingdom of Great Britain and Northern Ireland \\
    \bottomrule
    \end{tabular}
    }%
    \caption{Caption}
    \label{tab:benchmark:regions_full}
\end{table}

\subsubsection{Number of Samples per Task per Region}
\label{appendix:sec:benchmark:samples}
\begin{table}[ht!]
    \centering
    \renewcommand{\arraystretch}{.94}
    \resizebox{\linewidth}{!}{%
    \begin{tabular}{lrrrrrr}
    \toprule
    \textsc{Region} & \sivqa & \vvqa & \coqar & \coqac & \ckqad & \ckqan \\
    \midrule
    \RegA & 375 & 296 & 71 & 127 & 71 & 71 \\
    \RegA \RegAP & 4 & 4 & 2 & 2 & 1 & 1 \\
    \RegA \RegAP \RegE \RegW & 5 & 5 & 0 & 36 & 2 & 2 \\
    \RegA \RegE \RegW & 1 & 0 & 3 & 7 & 1 & 1 \\
    \RegA \RegSA & 8 & 0 & 2 & 3 & 1 & 1 \\
    \RegAP & 444 & 407 & 211 & 222 & 211 & 211 \\
    \RegAP \RegE & 7 & 7 & 6 & 6 & 3 & 3 \\
    \RegAP \RegE \RegLAC \RegSA \RegW & 1 & 1 & 0 & 8 & 1 & 1 \\
    \RegAP \RegE \RegW & 10 & 7 & 21 & 35 & 7 & 7 \\
    \RegAP \RegW & 4 & 3 & 2 & 3 & 1 & 1 \\
    \RegE & 302 & 242 & 125 & 136 & 125 & 125 \\
    \RegE \RegW & 21 & 20 & 22 & 56 & 11 & 11 \\
    \RegLAC & 420 & 341 & 96 & 106 & 96 & 96 \\
    \RegLAC \RegW & 2 & 2 & 2 & 2 & 1 & 1 \\
    \RegSA & 388 & 299 & 71 & 80 & 71 & 71 \\
    \RegW & 241 & 175 & 125 & 153 & 125 & 125 \\
    \bottomrule
    \end{tabular}
    }%
    \caption{Number of samples per region(s) in \dsname tasks.}
    \label{tab:benchmark:datasets:samples}
\end{table}

\subsection{Models}
\label{appendix:sec:benchmark:models}
We present the comprehensive list of all 31 models evaluated in \dsname in Table~\ref{tab:benchmark:models}.
\begin{table}[t]
	\centering
	\renewcommand{\arraystretch}{.97}
	\resizebox{\textwidth}{!}{%
		\begin{tabular}{lllllll l}
			\toprule
			\textsc{Model ID}                                 & \textsc{Paper Name}  & \textsc{Open-Weight} & \textsc{Size Group} & \textsc{Image Input} & \textsc{Video Input} & \textsc{Text Input} & \textsc{LLM Backbone}                     \\
			\midrule
			\rowcolor{gray!20}
			\texttt{claude-3-5-sonnet-20241022}               & \makecell{Claude 3.5 Sonnet\\\cite{anthropic2024claude}} & No                   & A                   & Yes                  & Yes                  & Yes                 & \multicolumn{1}{c}{--}                    \\
			\texttt{gemini-1.5-pro-002}                       & \makecell{Gemini Pro\\\cite{team2024gemini1.5}} & No                   & A                   & Yes                  & Yes                  & Yes                 & \multicolumn{1}{c}{--}                    \\
			\rowcolor{gray!20}
			\texttt{gemini-1.5-flash-002}                     & \makecell{Gemini Flash\\\cite{team2024gemini1.5}} & No                   & A                   & Yes                  & Yes                  & Yes                 & \multicolumn{1}{c}{--}                    \\
			\texttt{gpt-4o-2024-11-20}                        & \makecell{GPT-4o\\\cite{hurst2024gpt4o}} & No                   & A                   & Yes                  & Yes                  & Yes                 & \multicolumn{1}{c}{--}                    \\
			\rowcolor{gray!20}
			\texttt{gpt-4o-mini-2024-07-18}                   & \makecell{GPT-4o Mini\\\cite{hurst2024gpt4o}} & No                   & A                   & Yes                  & Yes                  & Yes                 & \multicolumn{1}{c}{--}                    \\
			\midrule
			\texttt{opengvlab/internvl2\_5-78b}               & \makecell{InternVL2.5 78B\\\cite{chen2024internvl2_5}} & Yes                  & XL                  & Yes                  & Yes                  & Yes                 & \texttt{qwen/qwen2.5-72b-instruct}        \\
			\rowcolor{gray!20}
			\texttt{qwen/qwen2-vl-72b-instruct}               & \makecell{Qwen2 VL 72B\\\cite{wang2024qwen2vl}} & Yes                  & XL                  & Yes                  & Yes                  & Yes                 & \texttt{qwen/qwen2.5-72b-instruct}        \\
			\texttt{opengvlab/internvl2\_5-26b}               & \makecell{InternVL2.5 26B\\\cite{chen2024internvl2_5}} & Yes                  & L                   & Yes                  & Yes                  & Yes                 & \texttt{internlm/internlm2\_5-20b-chat}   \\
			\rowcolor{gray!20}
			\texttt{opengvlab/internvl2\_5-38b}               & \makecell{InternVL2.5 38B\\\cite{chen2024internvl2_5}} & Yes                  & L                   & Yes                  & Yes                  & Yes                 & \texttt{qwen/qwen2.5-32b-instruct}        \\
			\texttt{meta-llama/llama-3.2-11b-vision-instruct} & \makecell{Llama 3.2 11B Vision\\\cite{meta2024llama3_2_v}} & Yes                  & M                   & Yes                  & Yes                  & Yes                 & \multicolumn{1}{c}{--}                    \\
			\rowcolor{gray!20}
			\texttt{qwen/qwen2-vl-7b-instruct}                & \makecell{Qwen2 VL 7B\\\cite{wang2024qwen2vl}} & Yes                  & M                   & Yes                  & Yes                  & Yes                 & \texttt{qwen/qwen2.5-7b-instruct}         \\
			\texttt{openbmb/minicpm-v-2\_6}                   & \makecell{MiniCPM V 2.6\\\cite{yao2024minicpm}} & Yes                  & M                   & Yes                  & Yes                  & Yes                 & \multicolumn{1}{c}{--}                    \\
			\rowcolor{gray!20}
			\texttt{wuenlp/centurio\_aya}                     & \makecell{Centurio Aya\\\cite{geigle2025centurio}} & Yes                  & M                   & Yes                  & Yes                  & Yes                 & \texttt{cohereforai/aya-expanse-8b}       \\
			\texttt{opengvlab/internvl2\_5-8b}                & \makecell{InternVL2.5 8B\\\cite{chen2024internvl2_5}} & Yes                  & M                   & Yes                  & Yes                  & Yes                 & \texttt{internlm/internlm2\_5-7b-chat}    \\
			\rowcolor{gray!20}
			\texttt{wuenlp/centurio\_qwen}                    & \makecell{Centurio Qwen\\\cite{geigle2025centurio}} & Yes                  & M                   & Yes                  & Yes                  & Yes                 & \texttt{qwen/qwen2.5-7b-instruct}         \\
			\texttt{qwen/qwen2-vl-2b-instruct}                & \makecell{Qwen2 VL 2B\\\cite{wang2024qwen2vl}} & Yes                  & S                   & Yes                  & Yes                  & Yes                 & \texttt{qwen/qwen2.5-1.5b-instruct}       \\
			\rowcolor{gray!20}
			\texttt{microsoft/phi-3.5-vision-instruct}        & \makecell{Phi 3.5 Vision\\\cite{abdin2024phi3}} & Yes                  & S                   & Yes                  & Yes                  & Yes                 & \texttt{microsoft/phi-3.5-mini-instruct}  \\
			\texttt{opengvlab/internvl2\_5-4b}                & \makecell{InternVL2.5 4B\\\cite{chen2024internvl2_5}} & Yes                  & S                   & Yes                  & Yes                  & Yes                 & \texttt{qwen/qwen2.5-3b-instruct}         \\
			\rowcolor{gray!20}
			\texttt{opengvlab/internvl2\_5-1b}                & \makecell{InternVL2.5 1B\\\cite{chen2024internvl2_5}} & Yes                  & S                   & Yes                  & Yes                  & Yes                 & \texttt{qwen/qwen2.5-0.5b-instruct}       \\
			\texttt{opengvlab/internvl2\_5-2b}                & \makecell{InternVL2.5 2B\\\cite{chen2024internvl2_5}} & Yes                  & S                   & Yes                  & Yes                  & Yes                 & \texttt{internlm/internlm2\_5-1\_8b-chat} \\
			\midrule
			\rowcolor{gray!20}
			\texttt{qwen/qwen2.5-72b-instruct}                & \makecell{Qwen2.5 72B\\\cite{yang2024qwen2.5}} & Yes                  & XL                  & No                   & No                   & Yes                 & \multicolumn{1}{c}{--}                    \\
			\texttt{qwen/qwen2.5-32b-instruct}                & \makecell{Qwen2.5 32B\\\cite{yang2024qwen2.5}} & Yes                  & L                   & No                   & No                   & Yes                 & \multicolumn{1}{c}{--}                    \\
			\rowcolor{gray!20}
			\texttt{internlm/internlm2\_5-20b-chat}           & \makecell{InternLM2.5 20B\\\cite{cai2024internlm2}} & Yes                  & L                   & No                   & No                   & Yes                 & \multicolumn{1}{c}{--}                    \\
			\texttt{cohereforai/aya-expanse-8b}               & \makecell{Aya Expanse 8B\\\cite{dang2024ayaexpanse}} & Yes                  & M                   & No                   & No                   & Yes                 & \multicolumn{1}{c}{--}                    \\
			\rowcolor{gray!20}
			\texttt{internlm/internlm2\_5-7b-chat}            & \makecell{InternLM2.5 7B\\\cite{cai2024internlm2}} & Yes                  & M                   & No                   & No                   & Yes                 & \multicolumn{1}{c}{--}                    \\
			\texttt{qwen/qwen2.5-7b-instruct}                 & \makecell{Qwen2.5 7B\\\cite{yang2024qwen2.5}} & Yes                  & M                   & No                   & No                   & Yes                 & \multicolumn{1}{c}{--}                    \\
			\rowcolor{gray!20}
			\texttt{qwen/qwen2.5-0.5b-instruct}               & \makecell{Qwen2.5 0.5B\\\cite{yang2024qwen2.5}} & Yes                  & S                   & No                   & No                   & Yes                 & \multicolumn{1}{c}{--}                    \\
			\texttt{qwen/qwen2.5-3b-instruct}                 & \makecell{Qwen2.5 3B\\\cite{yang2024qwen2.5}} & Yes                  & S                   & No                   & No                   & Yes                 & \multicolumn{1}{c}{--}                    \\
			\rowcolor{gray!20}
			\texttt{qwen/qwen2.5-1.5b-instruct}               & \makecell{Qwen2.5 1.5B\\\cite{yang2024qwen2.5}} & Yes                  & S                   & No                   & No                   & Yes                 & \multicolumn{1}{c}{--}                    \\
			\texttt{internlm/internlm2\_5-1\_8b-chat}         & \makecell{InternLM2.5 1.8B\\\cite{cai2024internlm2}} & Yes                  & S                   & No                   & No                   & Yes                 & \multicolumn{1}{c}{--}                    \\
			\rowcolor{gray!20}
			\texttt{microsoft/phi-3.5-mini-instruct}          & \makecell{Phi 3.5 Mini\\\cite{abdin2024phi3}} & Yes                  & S                   & No                   & No                   & Yes                 & \multicolumn{1}{c}{--}                    \\
			\bottomrule
		\end{tabular}
	}%
	\caption{Details about the models evaluated within the \dsname benchmark. The size ``A'' indicates that the model is a proprietary API model with unknown size.}
	\label{tab:benchmark:models}
\end{table}
\input{src/991_1_0_appendix_sivqa}
\input{src/991_2_0_appendix_vvqa}
\input{src/991_3_0_appendix_coqa}
\input{src/991_4_0_appendix_ckqa}

%% file: examples/cef/western-european-and-north-american-states_0.tex
\begin{figure}[H]
\begin{tcolorbox}[colback=gray!5!white,colframe=black!75!black,fonttitle=\bfseries\scriptsize,fontupper=\ttfamily\footnotesize]
  {\large{Title:}} {\normalsize{The skills related to perfume in Pays de Grasse: the cultivation of perfume plants, the knowledge and processing of natural raw materials, and the art of perfume composition}}\\
  {\normalsize{Countries:}} France\\
  {\normalsize{Regions:}} Western European and North American States\\
  {\normalsize{Description:}}\\
  The skills related to perfume in Pays de Grasse cover three different aspects: the cultivation of perfume plants; the knowledge and processing of natural raw materials; and the art of perfume composition. The practice involves a wide range of communities and groups, brought together under the Association du Patrimoine Vivant du Pays de Grasse (Living Heritage Association of the Region of Grasse). Since at least the sixteenth century, the practices of growing and processing perfume plants and creating fragrant blends have been developed in Pays de Grasse, in a craft industry long dominated by leather tanning. Perfume plant cultivation involves a wide range of skills and knowledge, for instance pertaining to nature, soil, weather, biology, plant physiology and horticultural practices, as well as specific techniques such as extraction and hydraulic distillation methods. The inhabitants of Grasse have made these techniques their own and helped improve them. In addition to technical skills, however, the art also calls for imagination, memory and creativity. Perfume forges social bonds and provides an important source of seasonal labour. Related knowledge is mostly transmitted informally through a long learning process that still takes place primarily in perfumeries. In recent decades, however, there has been a growing interest in standardizing learning through formalized teaching.\\[2mm]
  {\normalsize{UNESCO ICH URL:}} \href{https://ich.unesco.org/en/RL/the-skills-related-to-perfume-in-pays-de-grasse-the-cultivation-of-perfume-plants-the-knowledge-and-processing-of-natural-raw-materials-and-the-art-of-perfume-composition-01207}{https://ich.unesco.org/en/RL/the-skills-related-to-perfume-i...}
  \begin{center}
    \begin{minipage}{0.18\linewidth}
      \centering
      \includegraphics[width=\linewidth]{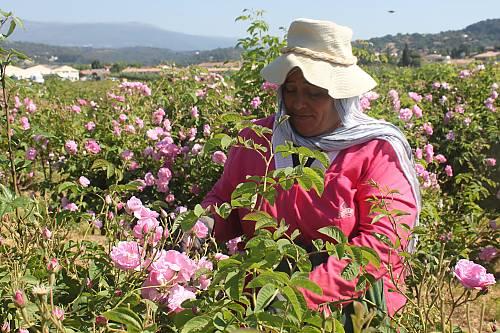}
      {\captionsetup{labelformat=empty}\captionof{figure}{\tiny\textit{Copyrigth: JM. Ghibaudo APVPG 2011}}}
    \end{minipage}\hfill
    \begin{minipage}{0.18\linewidth}
      \centering
      \includegraphics[width=\linewidth]{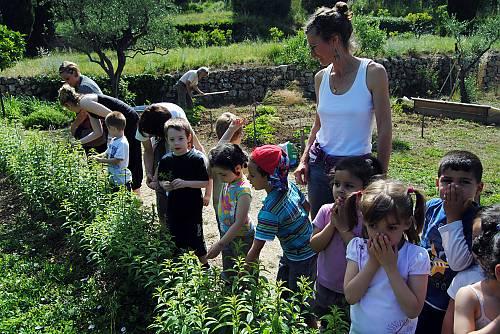}
      {\captionsetup{labelformat=empty}\captionof{figure}{\tiny\textit{Copyrigth: Musées de Grasse 2011}}}
    \end{minipage}\hfill
    \begin{minipage}{0.18\linewidth}
      \centering
      \includegraphics[width=\linewidth]{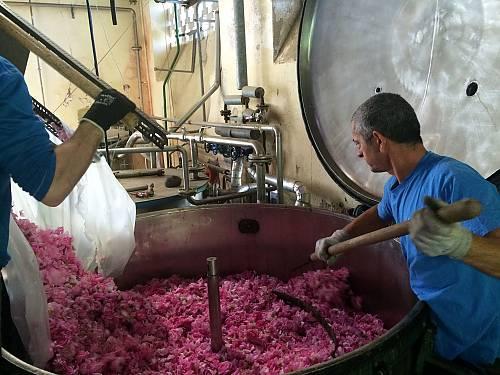}
      {\captionsetup{labelformat=empty}\captionof{figure}{\tiny\textit{Copyrigth: N. Bédar APVPG 2015}}}
    \end{minipage}\hfill
    \begin{minipage}{0.18\linewidth}
      \centering
      \includegraphics[width=\linewidth]{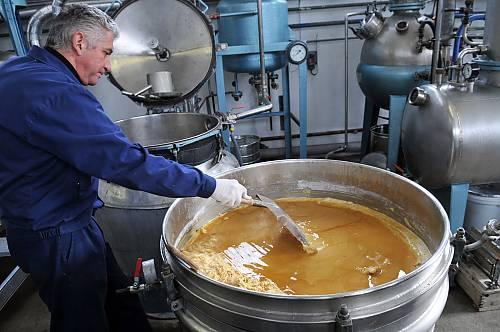}
      {\captionsetup{labelformat=empty}\captionof{figure}{\tiny\textit{Copyrigth: C. Barbiero/Musées de Grasse 2010}}}
    \end{minipage}\hfill
    \begin{minipage}{0.18\linewidth}
      \centering
      \includegraphics[width=\linewidth]{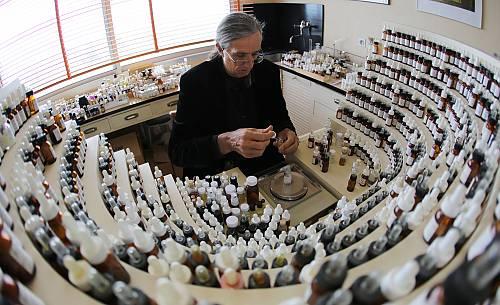}
      {\captionsetup{labelformat=empty}\captionof{figure}{\tiny\textit{Copyrigth: Daniel, Serre, M. Roudnitska APVPG 2014}}}
    \end{minipage}\hfill
  \\[4mm]
    \begin{minipage}{0.18\linewidth}
      \centering
      \includegraphics[width=\linewidth]{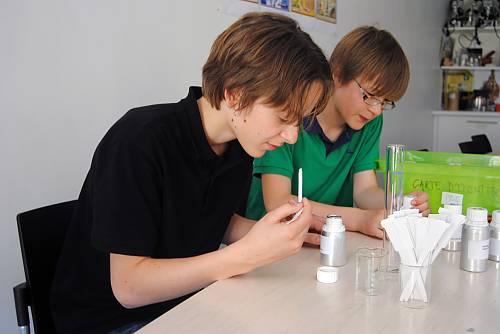}
      {\captionsetup{labelformat=empty}\captionof{figure}{\tiny\textit{Copyrigth: Musées de Grasse 2012}}}
    \end{minipage}\hfill
    \begin{minipage}{0.18\linewidth}
      \centering
      \includegraphics[width=\linewidth]{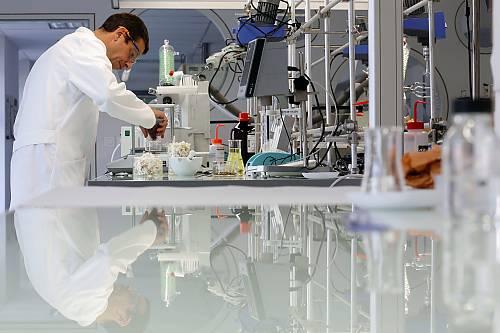}
      {\captionsetup{labelformat=empty}\captionof{figure}{\tiny\textit{Copyrigth: G. Voinot/Université Sophia Antiopolis 2011}}}
    \end{minipage}\hfill
    \begin{minipage}{0.18\linewidth}
      \centering
      \includegraphics[width=\linewidth]{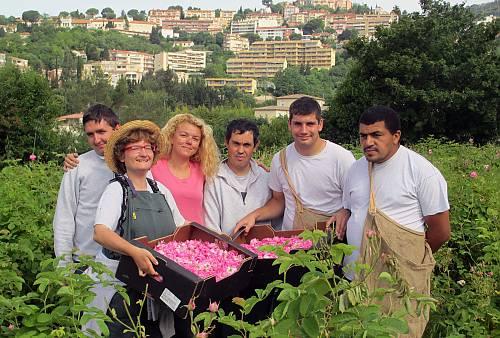}
      {\captionsetup{labelformat=empty}\captionof{figure}{\tiny\textit{Copyrigth: Esat Les Restanques 2013}}}
    \end{minipage}\hfill
    \begin{minipage}{0.18\linewidth}
      \centering
      \includegraphics[width=\linewidth]{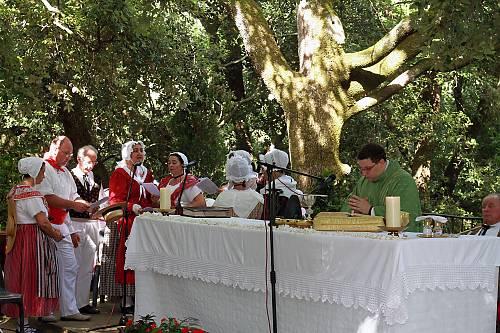}
      {\captionsetup{labelformat=empty}\captionof{figure}{\tiny\textit{Copyrigth: Forum des Associations Pays de Grasse 2014}}}
    \end{minipage}\hfill
    \begin{minipage}{0.18\linewidth}
      \centering
      \includegraphics[width=\linewidth]{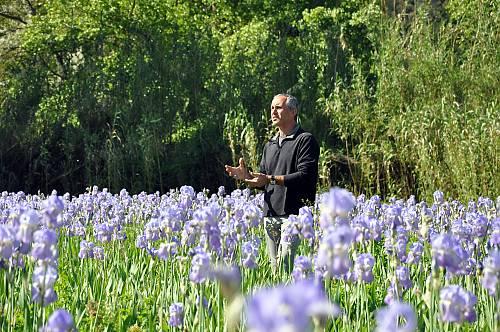}
      {\captionsetup{labelformat=empty}\captionof{figure}{\tiny\textit{Copyrigth: PH. Massé APVPG 2014}}}
    \end{minipage}\hfill
  \end{center}
\end{tcolorbox}
\end{figure}

%% file: examples/cef/eastern-european-states_0.tex
\begin{figure}[H]
\begin{tcolorbox}[colback=gray!5!white,colframe=black!75!black,fonttitle=\bfseries\scriptsize,fontupper=\ttfamily\footnotesize]
  {\large{Title:}} {\normalsize{Cultural Heritage of Boka Navy Kotor: a festive representation of a memory and cultural identity}}\\
  {\normalsize{Countries:}} Montenegro\\
  {\normalsize{Regions:}} Eastern European States\\
  {\normalsize{Description:}}\\
  Boka Navy is a traditional, non-governmental maritime organization founded in Kotor, Montenegro in 809. Its origin is linked to the arrival of the relics of St. Tryphon, the patron saint of the city of Kotor. Comprised of a community of seafarers with military, economic, educational and humanitarian functions, Boka Navy has played a memorial role for two centuries, preserving and promoting maritime history and tradition. Membership is voluntary and open to men, women and children of all ages. The organization is founded on the respect of human rights and of religious, national and cultural diversity. During formal celebrations, members wear colourful traditional uniforms, carry historic weapons and perform the traditional circle kolo dance. Boka Navy is the backbone of the annual St. Tryphon festivities, which take place from 13 January through 3 February and include a procession and a series of rituals in the cathedral. The external festivities begin with the Boka Navy’s traditional kolo circle dance and are followed by a procession carrying the relics of St. Tryphon through the main town squares and streets. Thousands of spectators attend the processions in the historic centre and observe the festive events. Hundreds of women, men and children also participate in preparations of the activities.\\[2mm]
  {\normalsize{UNESCO ICH URL:}} \href{https://ich.unesco.org/en/RL/cultural-heritage-of-boka-navy-kotor-a-festive-representation-of-a-memory-and-cultural-identity-01727}{https://ich.unesco.org/en/RL/cultural-heritage-of-boka-navy-...}
  \begin{center}
    \begin{minipage}{0.18\linewidth}
      \centering
      \includegraphics[width=\linewidth]{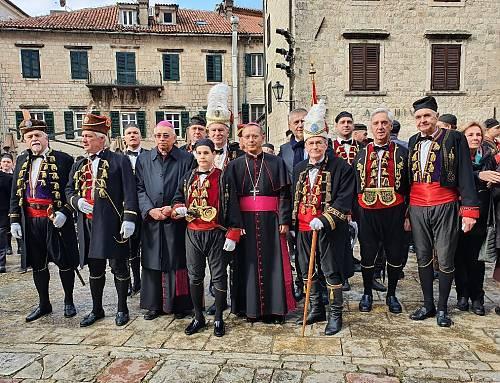}
      {\captionsetup{labelformat=empty}\captionof{figure}{\tiny\textit{Copyrigth: Ministry of Culture of Montenegro}}}
    \end{minipage}\hfill
    \begin{minipage}{0.18\linewidth}
      \centering
      \includegraphics[width=\linewidth]{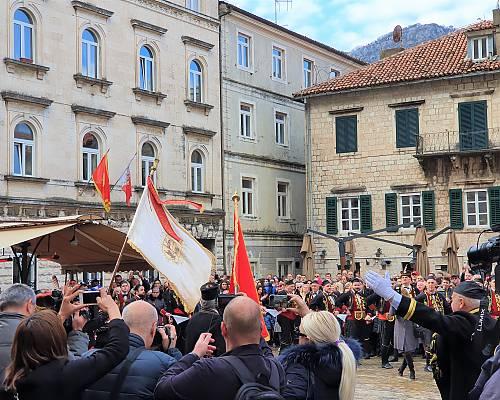}
      {\captionsetup{labelformat=empty}\captionof{figure}{\tiny\textit{Copyrigth: Ministry of Culture of Montenegro}}}
    \end{minipage}\hfill
    \begin{minipage}{0.18\linewidth}
      \centering
      \includegraphics[width=\linewidth]{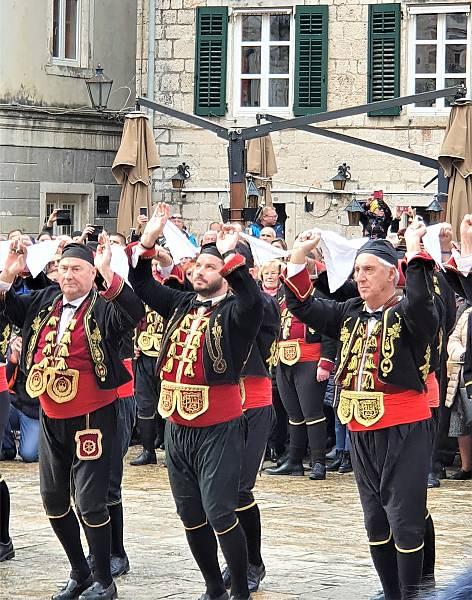}
      {\captionsetup{labelformat=empty}\captionof{figure}{\tiny\textit{Copyrigth: Ministry of Culture of Montenegro}}}
    \end{minipage}\hfill
    \begin{minipage}{0.18\linewidth}
      \centering
      \includegraphics[width=\linewidth]{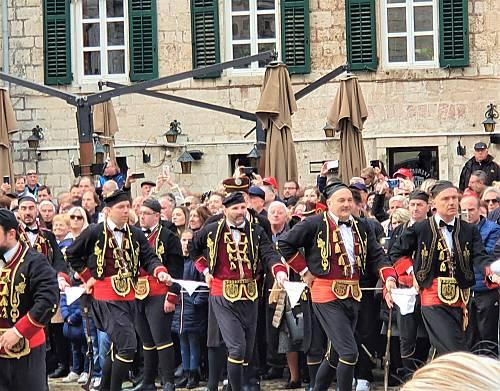}
      {\captionsetup{labelformat=empty}\captionof{figure}{\tiny\textit{Copyrigth: Ministry of Culture of Montenegro}}}
    \end{minipage}\hfill
    \begin{minipage}{0.18\linewidth}
      \centering
      \includegraphics[width=\linewidth]{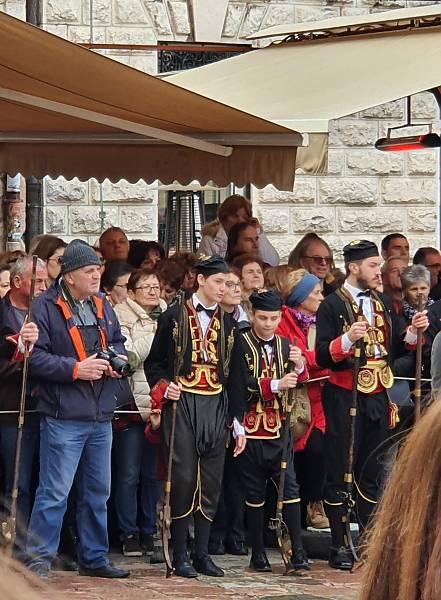}
      {\captionsetup{labelformat=empty}\captionof{figure}{\tiny\textit{Copyrigth: Ministry of Culture of Montenegro}}}
    \end{minipage}\hfill
  \\[4mm]
    \begin{minipage}{0.18\linewidth}
      \centering
      \includegraphics[width=\linewidth]{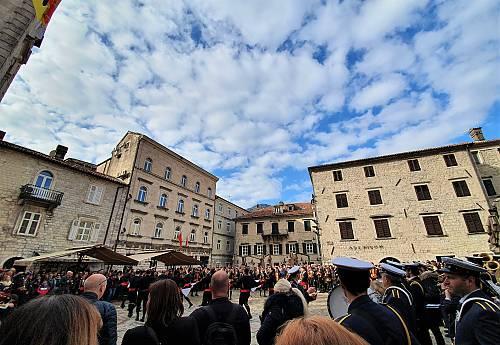}
      {\captionsetup{labelformat=empty}\captionof{figure}{\tiny\textit{Copyrigth: Ministry of Culture of Montenegro}}}
    \end{minipage}\hfill
    \begin{minipage}{0.18\linewidth}
      \centering
      \includegraphics[width=\linewidth]{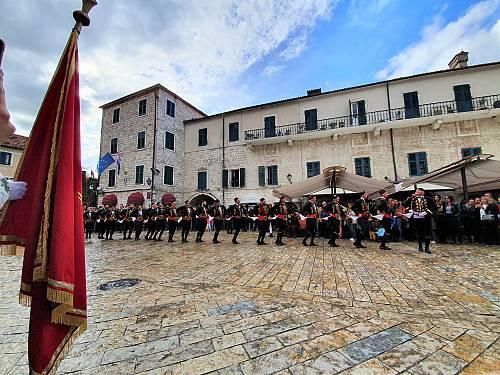}
      {\captionsetup{labelformat=empty}\captionof{figure}{\tiny\textit{Copyrigth: Ministry of Culture of Montenegro}}}
    \end{minipage}\hfill
    \begin{minipage}{0.18\linewidth}
      \centering
      \includegraphics[width=\linewidth]{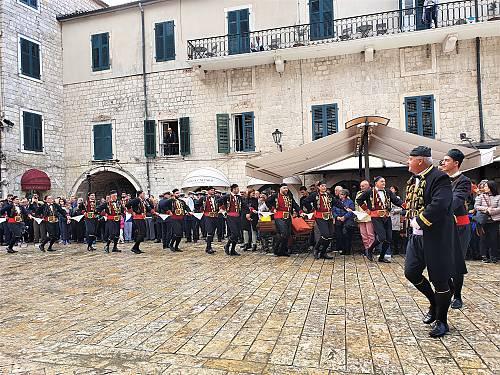}
      {\captionsetup{labelformat=empty}\captionof{figure}{\tiny\textit{Copyrigth: Ministry of Culture of Montenegro}}}
    \end{minipage}\hfill
    \begin{minipage}{0.18\linewidth}
      \centering
      \includegraphics[width=\linewidth]{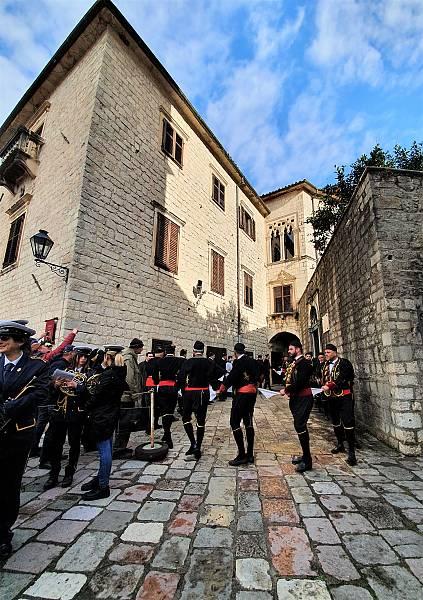}
      {\captionsetup{labelformat=empty}\captionof{figure}{\tiny\textit{Copyrigth: Ministry of Culture of Montenegro}}}
    \end{minipage}\hfill
    \begin{minipage}{0.18\linewidth}
      \centering
      \includegraphics[width=\linewidth]{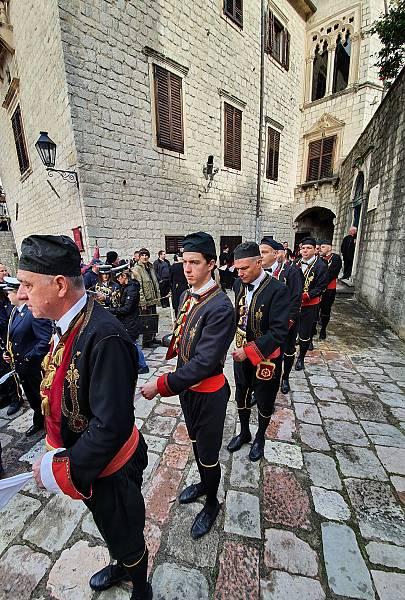}
      {\captionsetup{labelformat=empty}\captionof{figure}{\tiny\textit{Copyrigth: Ministry of Culture of Montenegro}}}
    \end{minipage}\hfill
  \end{center}
\end{tcolorbox}
\end{figure}

%% file: examples/cef/arab-states_0.tex
\begin{figure}[H]
\begin{tcolorbox}[colback=gray!5!white,colframe=black!75!black,fonttitle=\bfseries\scriptsize,fontupper=\ttfamily\footnotesize]
  {\large{Title:}} {\normalsize{Arts, skills and practices associated with engraving on metals (gold, silver and copper)}}\\
  {\normalsize{Countries:}} Algeria, Saudi Arabia, Egypt, Iraq, Morocco, Mauritania, Palestine, Sudan, Tunisia, Yemen\\
  {\normalsize{Regions:}} Arab States\\
  {\normalsize{Description:}}\\
  Engraving on metals such as gold, silver and copper is a centuries-old practice that entails manually cutting words, symbols or patterns into the surfaces of decorative, utilitarian, religious or ceremonial objects. The craftsperson uses different tools to manually cut symbols, names, Quran verses, prayers and geometric patterns into the objects. Engravings can be concave (recessed) or convex (elevated), or the result of a combination of different types of metals, such as gold and silver. Their social and symbolic meanings and functions vary according to the communities concerned. Engraved objects, such as jewelry or household objects, are often presented as traditional gifts for weddings or used in religious rituals and alternative medicine. For instance, certain types of metals are believed to have healing properties. Engraving on metals is transmitted within families, through observation and hands-on practice. It is also transmitted through workshops organized by training centres, organizations and universities, among others. Publications, cultural events and social media further contribute to the transmission of the related knowledge and skills. Practised by people of all ages and genders, metal engraving and the use of engraved objects are means of expressing the cultural, religious and geographical identity and the socioeconomic status of the communities concerned.\\[2mm]
  {\normalsize{UNESCO ICH URL:}} \href{https://ich.unesco.org/en/RL/arts-skills-and-practices-associated-with-engraving-on-metals-gold-silver-and-copper-01951}{https://ich.unesco.org/en/RL/arts-skills-and-practices-assoc...}
  \begin{center}
    \begin{minipage}{0.18\linewidth}
      \centering
      \includegraphics[width=\linewidth]{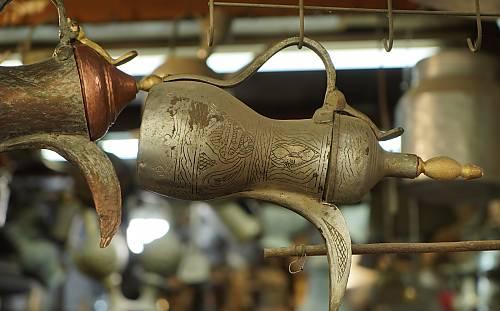}
      {\captionsetup{labelformat=empty}\captionof{figure}{\tiny\textit{Copyrigth: Huzaifa Ayad Bahaa El Din, Iraq, 2021}}}
    \end{minipage}\hfill
    \begin{minipage}{0.18\linewidth}
      \centering
      \includegraphics[width=\linewidth]{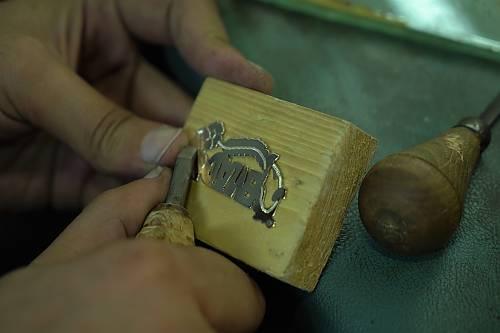}
      {\captionsetup{labelformat=empty}\captionof{figure}{\tiny\textit{Copyrigth: Huzaifa Ayad Bahaa El Din, Iraq, 2021}}}
    \end{minipage}\hfill
    \begin{minipage}{0.18\linewidth}
      \centering
      \includegraphics[width=\linewidth]{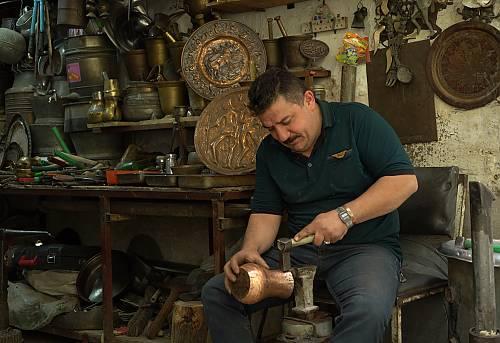}
      {\captionsetup{labelformat=empty}\captionof{figure}{\tiny\textit{Copyrigth: Huzaifa Ayad Bahaa El Din, Iraq, 2021}}}
    \end{minipage}\hfill
    \begin{minipage}{0.18\linewidth}
      \centering
      \includegraphics[width=\linewidth]{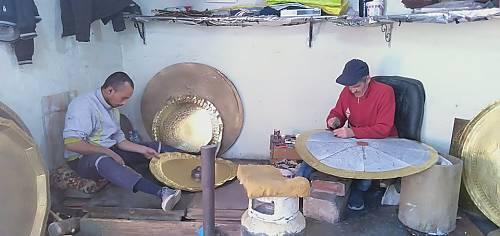}
      {\captionsetup{labelformat=empty}\captionof{figure}{\tiny\textit{Copyrigth: Zahia Benabdallah, Algeria, 2021}}}
    \end{minipage}\hfill
    \begin{minipage}{0.18\linewidth}
      \centering
      \includegraphics[width=\linewidth]{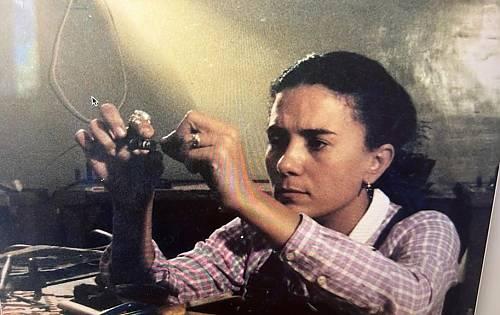}
      {\captionsetup{labelformat=empty}\captionof{figure}{\tiny\textit{Copyrigth: Azza Fahmi, Egypt, 2021}}}
    \end{minipage}\hfill
  \\[4mm]
    \begin{minipage}{0.18\linewidth}
      \centering
      \includegraphics[width=\linewidth]{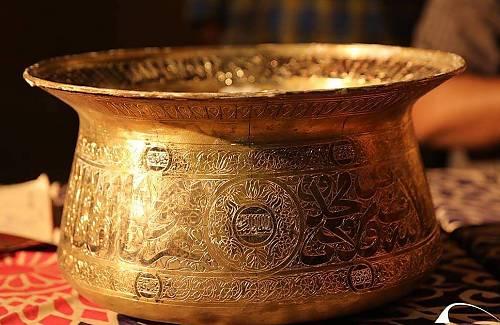}
      {\captionsetup{labelformat=empty}\captionof{figure}{\tiny\textit{Copyrigth: Mustafa Kamil, Egypt, 2021}}}
    \end{minipage}\hfill
    \begin{minipage}{0.18\linewidth}
      \centering
      \includegraphics[width=\linewidth]{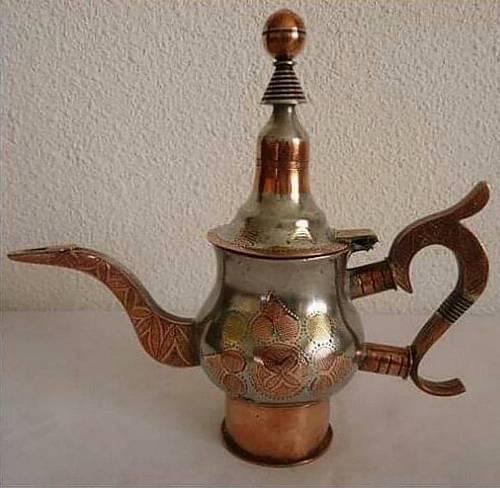}
      {\captionsetup{labelformat=empty}\captionof{figure}{\tiny\textit{Copyrigth: National Heritage Preservation, Ministry of Culture, Youth and Sport and Relations with the Parliament, Egypt, 2022}}}
    \end{minipage}\hfill
    \begin{minipage}{0.18\linewidth}
      \centering
      \includegraphics[width=\linewidth]{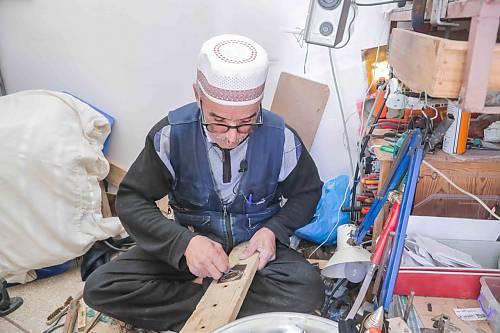}
      {\captionsetup{labelformat=empty}\captionof{figure}{\tiny\textit{Copyrigth: Direction du Patrimoine Culturel, Morocco, 2021}}}
    \end{minipage}\hfill
    \begin{minipage}{0.18\linewidth}
      \centering
      \includegraphics[width=\linewidth]{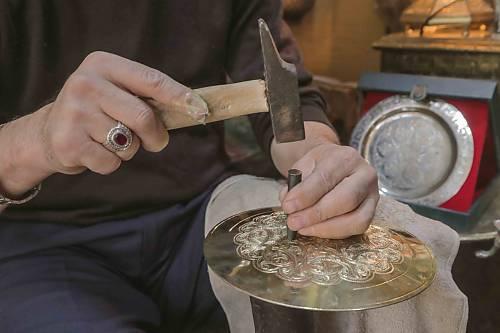}
      {\captionsetup{labelformat=empty}\captionof{figure}{\tiny\textit{Copyrigth: Direction du Patrimoine Culturel, Morocco, 2021}}}
    \end{minipage}\hfill
    \begin{minipage}{0.18\linewidth}
      \centering
      \includegraphics[width=\linewidth]{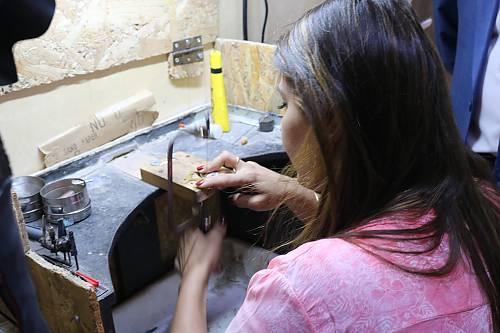}
      {\captionsetup{labelformat=empty}\captionof{figure}{\tiny\textit{Copyrigth: Ministry of Culture, Palestine, 2021}}}
    \end{minipage}\hfill
  \end{center}
\end{tcolorbox}
\end{figure}

%% file: examples/cef/asian-and-pacific-states_0.tex
\begin{figure}[H]
\begin{tcolorbox}[colback=gray!5!white,colframe=black!75!black,fonttitle=\bfseries\scriptsize,fontupper=\ttfamily\footnotesize]
  {\large{Title:}} {\normalsize{Tugging rituals and games}}\\
  {\normalsize{Countries:}} Cambodia, Korea, Philippines, Vietnam\\
  {\normalsize{Regions:}} Asian and Pacific States\\
  {\normalsize{Description:}}\\
  Tugging rituals and games in the rice-farming cultures of East Asia and Southeast Asia are enacted among communities to ensure abundant harvests and prosperity. They promote social solidarity, provide entertainment and mark the start of a new agricultural cycle. Many tugging rituals and games also have profound religious significance. Most variations include two teams, each of which pulls one end of a rope attempting to tug it from the other. The intentionally uncompetitive nature of the event removes the emphasis on winning or losing, affirming that these traditions are performed to promote the well-being of the community, and reminding members of the importance of cooperation. Many tugging games bear the traces of agricultural rituals, symbolizing the strength of natural forces, such as the sun and rain while also incorporating mythological elements or purification rites. Tugging rituals and games are often organized in front of a village’s communal house or shrine, preceded by commemorative rites to local protective deities. Village elders play active roles in leading and organizing younger people in playing the game and holding accompanying rituals. Tugging rituals and games also serve to strengthen unity and solidarity and sense of belonging and identity among community members.\\[2mm]
  {\normalsize{UNESCO ICH URL:}} \href{https://ich.unesco.org/en/RL/tugging-rituals-and-games-01080}{https://ich.unesco.org/en/RL/tugging-rituals-and-games-01080...}
  \begin{center}
    \begin{minipage}{0.18\linewidth}
      \centering
      \includegraphics[width=\linewidth]{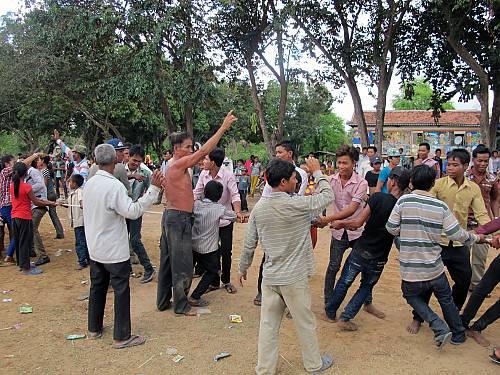}
      {\captionsetup{labelformat=empty}\captionof{figure}{\tiny\textit{Copyrigth: Siyonn Sophearith, 2013}}}
    \end{minipage}\hfill
    \begin{minipage}{0.18\linewidth}
      \centering
      \includegraphics[width=\linewidth]{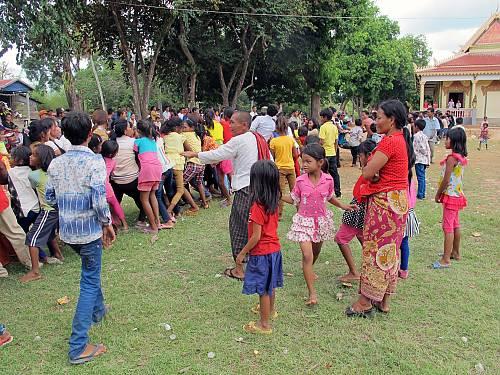}
      {\captionsetup{labelformat=empty}\captionof{figure}{\tiny\textit{Copyrigth: Siyonn Sophearith, 2013}}}
    \end{minipage}\hfill
    \begin{minipage}{0.18\linewidth}
      \centering
      \includegraphics[width=\linewidth]{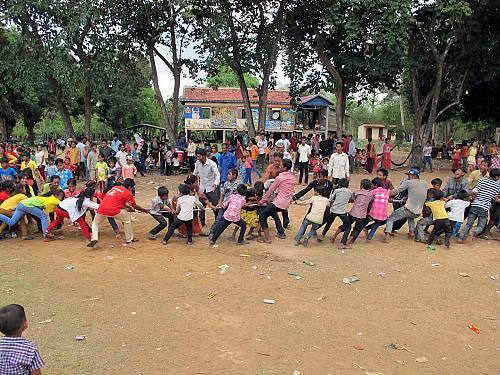}
      {\captionsetup{labelformat=empty}\captionof{figure}{\tiny\textit{Copyrigth: Siyonn Sophearith, 2013}}}
    \end{minipage}\hfill
    \begin{minipage}{0.18\linewidth}
      \centering
      \includegraphics[width=\linewidth]{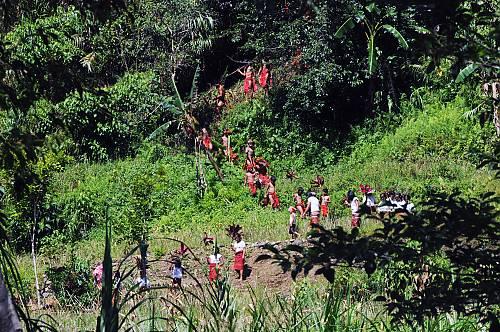}
      {\captionsetup{labelformat=empty}\captionof{figure}{\tiny\textit{Copyrigth: Renato S. Rastrollo, NCCA}}}
    \end{minipage}\hfill
    \begin{minipage}{0.18\linewidth}
      \centering
      \includegraphics[width=\linewidth]{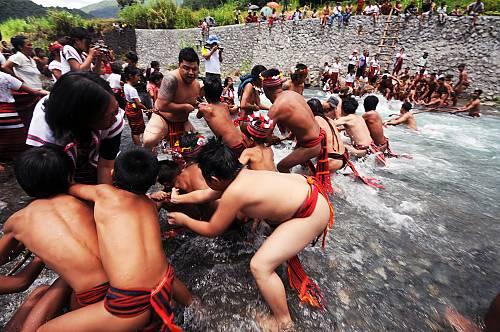}
      {\captionsetup{labelformat=empty}\captionof{figure}{\tiny\textit{Copyrigth: Renato S. Rastrollo, NCCA}}}
    \end{minipage}\hfill
  \\[4mm]
    \begin{minipage}{0.18\linewidth}
      \centering
      \includegraphics[width=\linewidth]{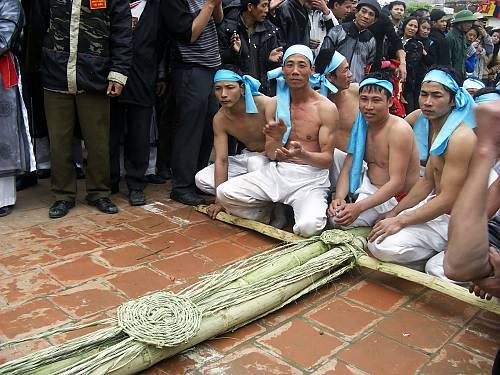}
      {\captionsetup{labelformat=empty}\captionof{figure}{\tiny\textit{Copyrigth: Vietnam Institute of Culture and Arts Studies, 2013}}}
    \end{minipage}\hfill
    \begin{minipage}{0.18\linewidth}
      \centering
      \includegraphics[width=\linewidth]{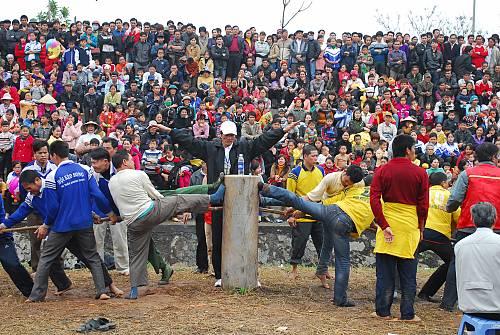}
      {\captionsetup{labelformat=empty}\captionof{figure}{\tiny\textit{Copyrigth: Vietnam Institute of Culture and Arts Studies, 2013}}}
    \end{minipage}\hfill
    \begin{minipage}{0.18\linewidth}
      \centering
      \includegraphics[width=\linewidth]{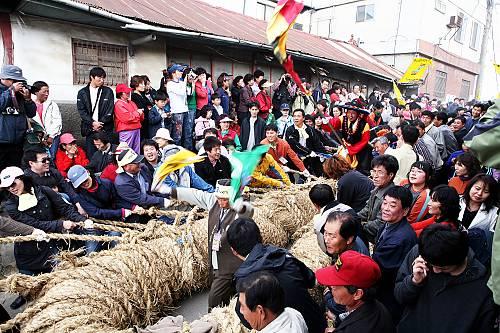}
      {\captionsetup{labelformat=empty}\captionof{figure}{\tiny\textit{Copyrigth: Joo Byung Soo, 2006}}}
    \end{minipage}\hfill
    \begin{minipage}{0.18\linewidth}
      \centering
      \includegraphics[width=\linewidth]{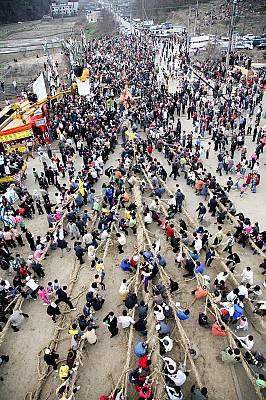}
      {\captionsetup{labelformat=empty}\captionof{figure}{\tiny\textit{Copyrigth: Joo Byung Soo, 2006}}}
    \end{minipage}\hfill
  \end{center}
\end{tcolorbox}
\end{figure}

%% file: examples/cef/latin-american-and-caribbean-states_0.tex
\begin{figure}[H]
\begin{tcolorbox}[colback=gray!5!white,colframe=black!75!black,fonttitle=\bfseries\scriptsize,fontupper=\ttfamily\footnotesize]
  {\large{Title:}} {\normalsize{Ancestral system of knowledge of the four indigenous peoples, Arhuaco, Kankuamo, Kogui and Wiwa of the Sierra Nevada de Santa Marta}}\\
  {\normalsize{Countries:}} Colombia\\
  {\normalsize{Regions:}} Latin-American and Caribbean States\\
  {\normalsize{Description:}}\\
  The Ancestral System of Knowledge of the Arhuaco, Kankuamo, Kogui and Wiwa peoples of the Sierra Nevada de Santa Marta is comprised of sacred mandates that keep the existence of the four peoples in harmony with the physical and spiritual universe. Through many years of dedication, the knowledgeable men (Mamos) and women (Sagas) acquire the necessary skills and sensitivity to communicate with the snow-capped peaks, connect with the knowledge of the rivers and decipher the messages of nature. Based on the Law of Origin, a philosophy that governs human relationships to nature and the universe, the Ancestral System of Knowledge entails caring for sacred sites and partaking in baptism rituals, marriage rites, traditional dances and songs, and retributions or offerings to spiritual powers. This ancestral wisdom is believed to play a fundamental role in protecting the Sierra Nevada ecosystem and avoiding the loss of the cultural identity of the four peoples of the region. The Ancestral System of Knowledge is transmitted from generation to generation through cultural practice, community activities, the use of the indigenous language and the implementation of the sacred mandates. The transmission process includes the understanding of physical and spiritual relationships with Mother Nature and sacred sites.\\[2mm]
  {\normalsize{UNESCO ICH URL:}} \href{https://ich.unesco.org/en/RL/ancestral-system-of-knowledge-of-the-four-indigenous-peoples-arhuaco-kankuamo-kogui-and-wiwa-of-the-sierra-nevada-de-santa-marta-01886}{https://ich.unesco.org/en/RL/ancestral-system-of-knowledge-o...}
  \begin{center}
    \begin{minipage}{0.18\linewidth}
      \centering
      \includegraphics[width=\linewidth]{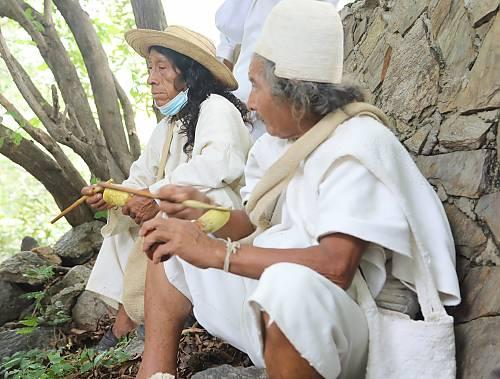}
      {\captionsetup{labelformat=empty}\captionof{figure}{\tiny\textit{Copyrigth: William Diaz, 2021}}}
    \end{minipage}\hfill
    \begin{minipage}{0.18\linewidth}
      \centering
      \includegraphics[width=\linewidth]{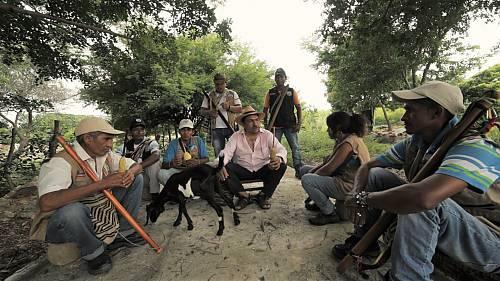}
      {\captionsetup{labelformat=empty}\captionof{figure}{\tiny\textit{Copyrigth: Jorge Mario Suarez/Government of Magdalena, 2017}}}
    \end{minipage}\hfill
    \begin{minipage}{0.18\linewidth}
      \centering
      \includegraphics[width=\linewidth]{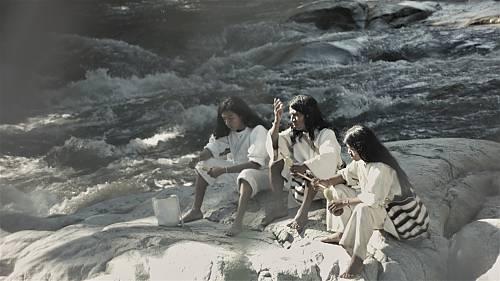}
      {\captionsetup{labelformat=empty}\captionof{figure}{\tiny\textit{Copyrigth: Jorge Mario Suarez/Government of Magdalena, 2017}}}
    \end{minipage}\hfill
    \begin{minipage}{0.18\linewidth}
      \centering
      \includegraphics[width=\linewidth]{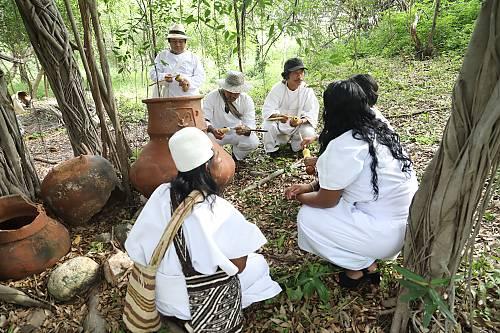}
      {\captionsetup{labelformat=empty}\captionof{figure}{\tiny\textit{Copyrigth: William Diaz, 2021}}}
    \end{minipage}\hfill
    \begin{minipage}{0.18\linewidth}
      \centering
      \includegraphics[width=\linewidth]{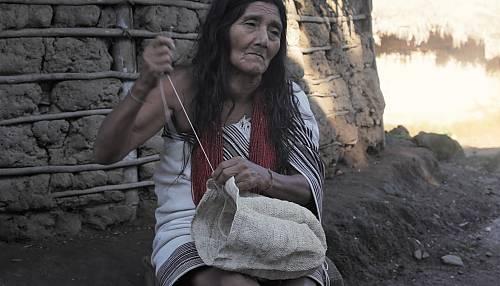}
      {\captionsetup{labelformat=empty}\captionof{figure}{\tiny\textit{Copyrigth: Jorge Mario Suarez/Government of Magdalena, 2017}}}
    \end{minipage}\hfill
  \\[4mm]
    \begin{minipage}{0.18\linewidth}
      \centering
      \includegraphics[width=\linewidth]{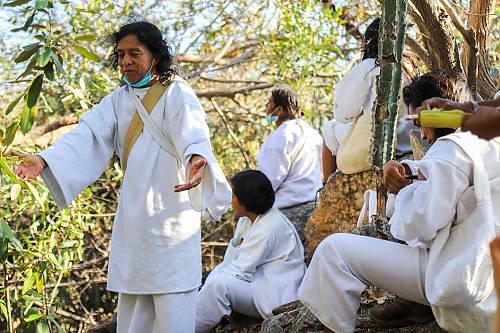}
      {\captionsetup{labelformat=empty}\captionof{figure}{\tiny\textit{Copyrigth: Jorge Mario Suarez/Government of Magdalena, 2017}}}
    \end{minipage}\hfill
    \begin{minipage}{0.18\linewidth}
      \centering
      \includegraphics[width=\linewidth]{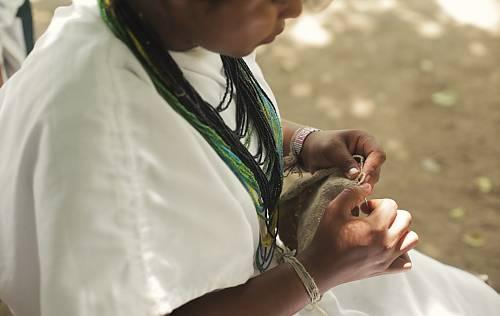}
      {\captionsetup{labelformat=empty}\captionof{figure}{\tiny\textit{Copyrigth: Jorge Mario Suarez/Government of Magdalena, 2017}}}
    \end{minipage}\hfill
    \begin{minipage}{0.18\linewidth}
      \centering
      \includegraphics[width=\linewidth]{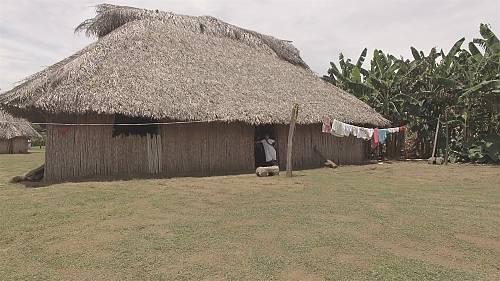}
      {\captionsetup{labelformat=empty}\captionof{figure}{\tiny\textit{Copyrigth: Jorge Mario Suarez/Government of Magdalena, 2017}}}
    \end{minipage}\hfill
    \begin{minipage}{0.18\linewidth}
      \centering
      \includegraphics[width=\linewidth]{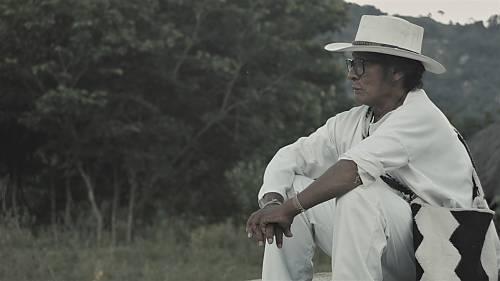}
      {\captionsetup{labelformat=empty}\captionof{figure}{\tiny\textit{Copyrigth: Jorge Mario Suarez/Government of Magdalena, 2017}}}
    \end{minipage}\hfill
    \begin{minipage}{0.18\linewidth}
      \centering
      \includegraphics[width=\linewidth]{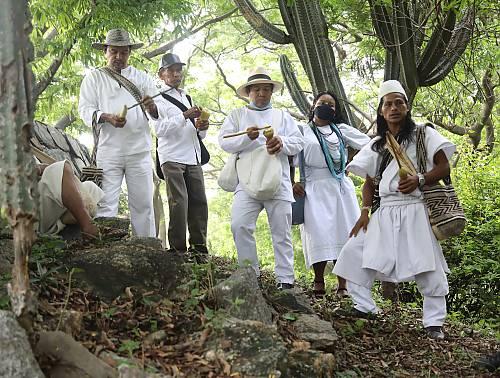}
      {\captionsetup{labelformat=empty}\captionof{figure}{\tiny\textit{Copyrigth: William Diaz, 2021}}}
    \end{minipage}\hfill
  \end{center}
\end{tcolorbox}
\end{figure}

%% file: examples/cef/subsaharian-african-states_0.tex
\begin{figure}[H]
\begin{tcolorbox}[colback=gray!5!white,colframe=black!75!black,fonttitle=\bfseries\scriptsize,fontupper=\ttfamily\footnotesize]
  {\large{Title:}} {\normalsize{Gada system, an indigenous democratic socio-political system of the Oromo}}\\
  {\normalsize{Countries:}} Ethiopia\\
  {\normalsize{Regions:}} Subsaharian African States\\
  {\normalsize{Description:}}\\
  Gada is a traditional system of governance used by the Oromo people in Ethiopia developed from knowledge gained by community experience over generations. The system regulates political, economic, social and religious activities of the community dealing with issues such as conflict resolution, reparation and protecting women’s rights. It serves as a mechanism for enforcing moral conduct, building social cohesion, and expressing forms of community culture. Gada is organized into five classes with one of these functioning as the ruling class consisting of a chairperson, officials and an assembly. Each class progresses through a series of grades before it can function in authority with the leadership changing on a rotational basis every eight years. Class membership is open to men, whose fathers are already members, while women are consulted for decision-making on protecting women’s rights. The classes are taught by oral historians covering history, laws, rituals, time reckoning, cosmology, myths, rules of conduct, and the function of the Gada system. Meetings and ceremonies take place under a sycamore tree (considered the Gada symbol) while major clans have established Gada centres and ceremonial spaces according to territory. Knowledge about the Gada system is transmitted to children in the home and at school.\\[2mm]
  {\normalsize{UNESCO ICH URL:}} \href{https://ich.unesco.org/en/RL/gada-system-an-indigenous-democratic-socio-political-system-of-the-oromo-01164}{https://ich.unesco.org/en/RL/gada-system-an-indigenous-democ...}
  \begin{center}
    \begin{minipage}{0.18\linewidth}
      \centering
      \includegraphics[width=\linewidth]{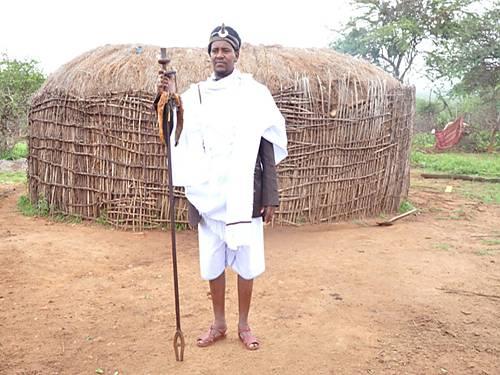}
      {\captionsetup{labelformat=empty}\captionof{figure}{\tiny\textit{Copyrigth: Authority for Research and Conservation of Cultural Heritage (ARCCH), Ethiopia, 2014}}}
    \end{minipage}\hfill
    \begin{minipage}{0.18\linewidth}
      \centering
      \includegraphics[width=\linewidth]{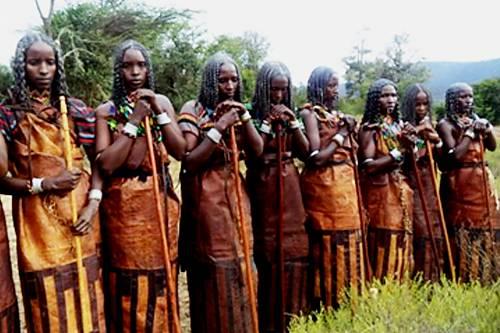}
      {\captionsetup{labelformat=empty}\captionof{figure}{\tiny\textit{Copyrigth: Authority for Research and Conservation of Cultural Heritage (ARCCH), Ethiopia, 2014}}}
    \end{minipage}\hfill
    \begin{minipage}{0.18\linewidth}
      \centering
      \includegraphics[width=\linewidth]{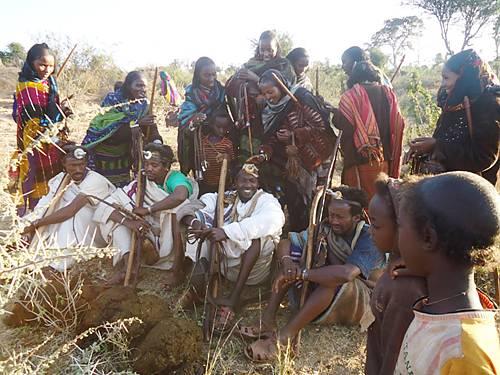}
      {\captionsetup{labelformat=empty}\captionof{figure}{\tiny\textit{Copyrigth: Authority for Research and Conservation of Cultural Heritage (ARCCH), Ethiopia, 2014}}}
    \end{minipage}\hfill
    \begin{minipage}{0.18\linewidth}
      \centering
      \includegraphics[width=\linewidth]{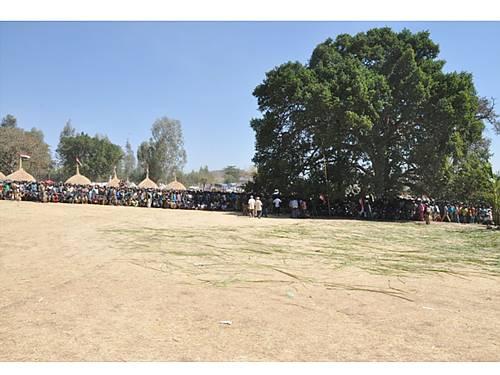}
      {\captionsetup{labelformat=empty}\captionof{figure}{\tiny\textit{Copyrigth: Authority for Research and Conservation of Cultural Heritage (ARCCH), Ethiopia, 2014}}}
    \end{minipage}\hfill
    \begin{minipage}{0.18\linewidth}
      \centering
      \includegraphics[width=\linewidth]{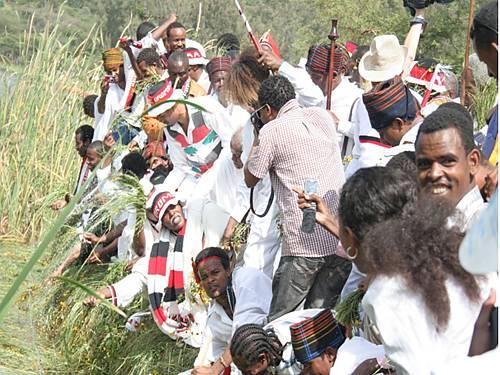}
      {\captionsetup{labelformat=empty}\captionof{figure}{\tiny\textit{Copyrigth: Authority for Research and Conservation of Cultural Heritage (ARCCH), Ethiopia, 2014}}}
    \end{minipage}\hfill
  \\[4mm]
    \begin{minipage}{0.18\linewidth}
      \centering
      \includegraphics[width=\linewidth]{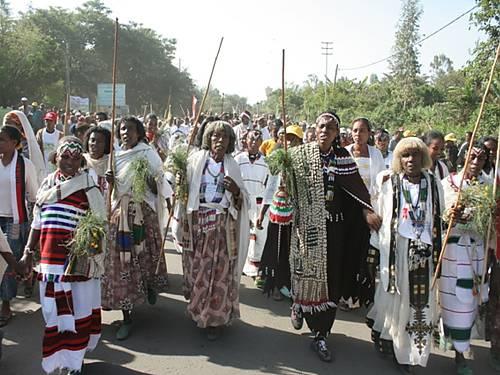}
      {\captionsetup{labelformat=empty}\captionof{figure}{\tiny\textit{Copyrigth: Authority for Research and Conservation of Cultural Heritage (ARCCH), Ethiopia, 2014}}}
    \end{minipage}\hfill
    \begin{minipage}{0.18\linewidth}
      \centering
      \includegraphics[width=\linewidth]{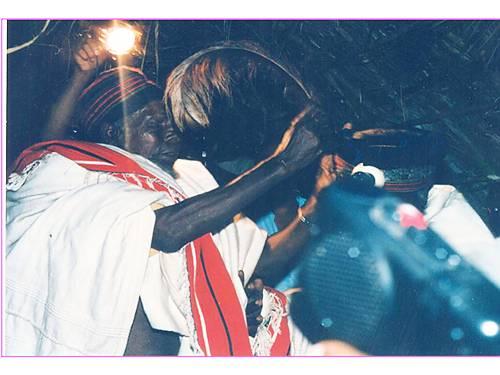}
      {\captionsetup{labelformat=empty}\captionof{figure}{\tiny\textit{Copyrigth: Authority for Research and Conservation of Cultural Heritage (ARCCH), Ethiopia, 2014}}}
    \end{minipage}\hfill
    \begin{minipage}{0.18\linewidth}
      \centering
      \includegraphics[width=\linewidth]{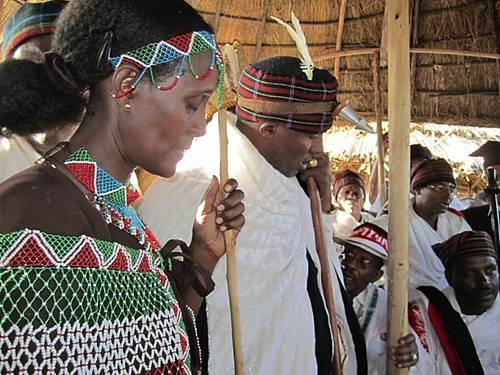}
      {\captionsetup{labelformat=empty}\captionof{figure}{\tiny\textit{Copyrigth: Authority for Research and Conservation of Cultural Heritage (ARCCH), Ethiopia, 2014}}}
    \end{minipage}\hfill
    \begin{minipage}{0.18\linewidth}
      \centering
      \includegraphics[width=\linewidth]{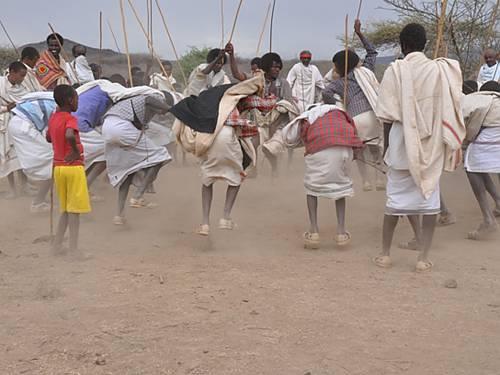}
      {\captionsetup{labelformat=empty}\captionof{figure}{\tiny\textit{Copyrigth: Authority for Research and Conservation of Cultural Heritage (ARCCH), Ethiopia, 2014}}}
    \end{minipage}\hfill
    \begin{minipage}{0.18\linewidth}
      \centering
      \includegraphics[width=\linewidth]{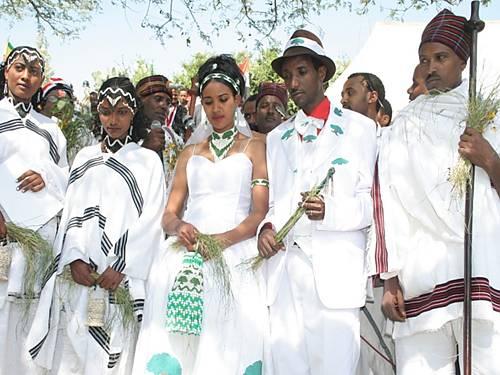}
      {\captionsetup{labelformat=empty}\captionof{figure}{\tiny\textit{Copyrigth: Authority for Research and Conservation of Cultural Heritage (ARCCH), Ethiopia, 2014}}}
    \end{minipage}\hfill
  \end{center}
\end{tcolorbox}
\end{figure}

%% file: src/991_1_0_appendix_sivqa.tex
\section{\sivqa Details}
\label{appendix:sec:sivqa}

\clearpage
\subsection{Examples}
\label{appendix:sec:sivqa:examples}
In the following, we provide one random sample per region for the \sivqa task.
Note that the lower part of the examples, where the related CEF is provided, is \emph{not} part of the actual sample.

\subsubsection*{\RegA}
\input{examples/sivqa/arab-states_0}
\subsubsection*{\RegAP}
\input{examples/sivqa/asian-and-pacific-states_0}
\subsubsection*{\RegE}
\input{examples/sivqa/eastern-european-states_0}
\subsubsection*{\RegLAC}
\input{examples/sivqa/latin-american-and-caribbean-states_0}
\subsubsection*{\RegSA}
\input{examples/sivqa/subsaharian-african-states_0}
\subsubsection*{\RegW}
\input{examples/sivqa/western-european-and-north-american-states_0}

\subsection{Cultural Aspects}
\label{appedix:sec:sivqa:aspects}
During the synthetic data generation phase of the \sivqa, we also obtained a ``target aspect'' per question (see \S\ref{appendix:sec:sivqa:synth} and \S\ref{appendix:sec:sivqa:synth:sys_prompt}).
We report these aspects in the following.
\begin{table*}[ht!]
    \centering
    \small
    \begin{minipage}[t]{0.29\textwidth}
        \centering
        \begin{tabular}{lr}
            \toprule
            Aspect & Questions \\
            \midrule
            traditions & 390 \\
            rituals & 241 \\
            art & 233 \\
            music & 210 \\
            craftsmanship & 177 \\
            instruments & 155 \\
            festivals & 151 \\
            dance & 150 \\
            tools & 108 \\
            food & 96 \\
            clothing & 93 \\
            architecture & 52 \\
            sports & 38 \\
            location & 28 \\
            symbols & 19 \\
            drinks & 14 \\
            customs & 13 \\
            cultural significance & 6 \\
            theatre & 4 \\
            \bottomrule
        \end{tabular}
    \end{minipage}%
    \begin{minipage}[t]{0.29\textwidth}
        \centering
        \begin{tabular}{lr}
            \toprule
            Aspect & Questions \\
            \midrule
            education & 3 \\
            culture & 3 \\
            games & 3 \\
            performing arts & 3 \\
            language & 3 \\
            performance & 3 \\
            characters & 2 \\
            practices & 2 \\
            skills & 2 \\
            origin & 2 \\
            cultural identity & 2 \\
            technology & 1 \\
            people & 1 \\
            community & 1 \\
            identity & 1 \\
            environment & 1 \\
            traditional medicine & 1 \\
            nature & 1 \\
            communication & 1 \\
            \bottomrule
        \end{tabular}
    \end{minipage}%
    \begin{minipage}[t]{0.29\textwidth}
        \centering
        \begin{tabular}{lr}
            \toprule
            Aspect & Questions \\
            \midrule
            jewelry & 1 \\
            objects & 1 \\
            animal & 1 \\
            plants & 1 \\
            process & 1 \\
            agriculture & 1 \\
            celebrations & 1 \\
            details & 1 \\
            historical & 1 \\
            function or usage & 1 \\
            symbolism & 1 \\
            healthcare & 1 \\
            knowledge & 1 \\
            social status & 1 \\
            religion & 1 \\
            cultural space & 1 \\
            social space & 1 \\
            cultural practice & 1 \\
            unknown & 1 \\
            \bottomrule
        \end{tabular}
    \end{minipage}
    \caption{Cultural aspects targeted by the questions within the \sivqa task.}
    \label{tab:sivqa:aspects}
\end{table*}

\subsection{External Hint Variations}
\label{appendix:sec:sivqa:hints}
For the \sivqa (and \vvqa) task, we ablate the effect of external cues or hints on the task performance of models.
In the following, we provide the Python pseudo-code snippet to generate the prompt for a given sample.
\begin{figure*}[ht!]
    \centering
    \begin{promptbox}{Python Pseudo-Code for the external cue settings of the \sivqa and \vvqa tasks.}
    \begin{minted}[breaklines]{python}
def apply_gimmick_prompt_template(
    sample: dict[str, Any],
    regions_hint: bool,
    countries_hint: bool,
) -> str:
    
    prompt_template = "{QUESTION}\n{HINTS}\n"
    hints = ""

    if regions_hint:
        hints += (
            "Hint: The question is related to a cultural event or facet from the following region(s): "
            f"{', '.join(sample['regions'])}\n"
        )

    if countries_hint:
        hints += (
            "Hint: The question is related to a cultural event or facet from the following country or countries: "
            f"{', '.join(sample['countries'])}\n"
        )

    return prompt_template.format(
        QUESTION=sample["prompt"],
        HINTS=hints,
    )
    \end{minted}
    \end{promptbox}
    \label{fig:sivqa:hints}
    \caption{Python Pseudo-Code to generate the prompt for a given \sivqa (or \vvqa) sample for the external cues settings.}
\end{figure*}

\subsection{Synthetic Data Generation}
\label{appendix:sec:sivqa:synth}
\include{src/991_1_2_appendix_sivqa_prompt}

\subsection{Annotation Project Details}
\label{appendix:sec:sivqa:anno}
We first conducted several internal pilot studies to iteratively create a straightforward annotation task, guidelines, and an intuitive interface for the final annotation project.
To find annotators, we advertised the task in our faculty research network, emphasizing our goal of creating a culturally diverse benchmark for assessing the cultural awareness of current AI models.
Therefore, we targeted primarily individuals from non-Western cultural backgrounds.
We found 18 volunteers who have spent most of their lives in 10 different countries from all six regions and thus cover diverse cultural backgrounds (see Table~\ref{tab:sivqa:anno:demographics}).
To train the annotators, we provided detailed annotation guidelines, followed by an oral introduction to the task.
For more details, refer to the (anonymized) original annotation guidelines we \href{https://drive.proton.me/urls/T6RHQCEW5G#5y0Itm2BdWYZ}{shared here}.

For the second annotation round, we hired 5 of the previous volunteering annotators (0, 1, 8, 15, 17) who assessed the kept samples from the first round to obtain two annotations (from distinct annotators) per sample.
We paid the second-round annotators a salary of roughly 12.5€ per hour.
\begin{table}[ht!]
    \centering
    \renewcommand{\arraystretch}{0.95}
    \resizebox{\linewidth}{!}{%
    \begin{tabular}{lrllllr}
        \toprule
        \textsc{ID} & \textsc{Age} & \textsc{Pronouns} & \textsc{Education} & \textsc{Country} & \textsc{Region} & \textsc{Round(s)} \\
        \midrule
        0 & 23 & she/her & Bachelor & Iran & \RegAP & 1, 2\\
        1 & 23 & she/her & Bachelor & Iran & \RegAP & 1, 2\\
        2 & 28 & she/her & PhD & Russia & \RegE & 1\\
        3 & 35 & he/him & Master & Germany & \RegW & 1 \\
        5 & 29 & he/him & Bachelor & Guatemala & \RegLAC & 1\\
        6 & 29 & he/him & Master & Germany & \RegW & 1\\
        7 & 42 & he/him & PhD & Ethiopia & \RegSA & 1\\
        8 & 23 & he/him & Bachelor & Egypt & \RegA & 1, 2\\
        9 & 33 & she/her & Master & Iran & \RegAP & 1\\
        10 & 29 & she/her & Bachelor & Afghanistan & \RegAP & 1\\
        11 & 23 & she/her & Bachelor & India & \RegAP & 1\\
        12 & 33 & he/him & Bachelor & Germany & \RegW & 1\\
        13 & 22 & she/her & Bachelor & Pakistan & \RegAP & 1\\
        14 & 27 & he/him & Master & China & \RegAP & 1\\
        15 & 29 & she/her & High School & Germany & \RegW & 1, 2\\
        16 & 22 & she/her & Bachelor & China & \RegAP & 1\\
        17 & 26 & he/him & High School & Germany & \RegW & 1, 2, 3\\
        \bottomrule
    \end{tabular}
    }%
    \caption{Demographics of the annotators who participated in our VQA annotation project. For the country, we asked the question, ``\textit{Where did you spend most of your life?}''. The Round(s) column indicates which annotation rounds the annotator participated in.}
    \label{tab:sivqa:anno:demographics}
\end{table}

\subsubsection{\sivqa Annotation Interface}
\label{appendix:sec:sivqa:anno:ui}
\begin{figure*}
    \centering
    \begin{subfigure}[b]{1.\textwidth}
         \centering
         \includegraphics[width=\textwidth]{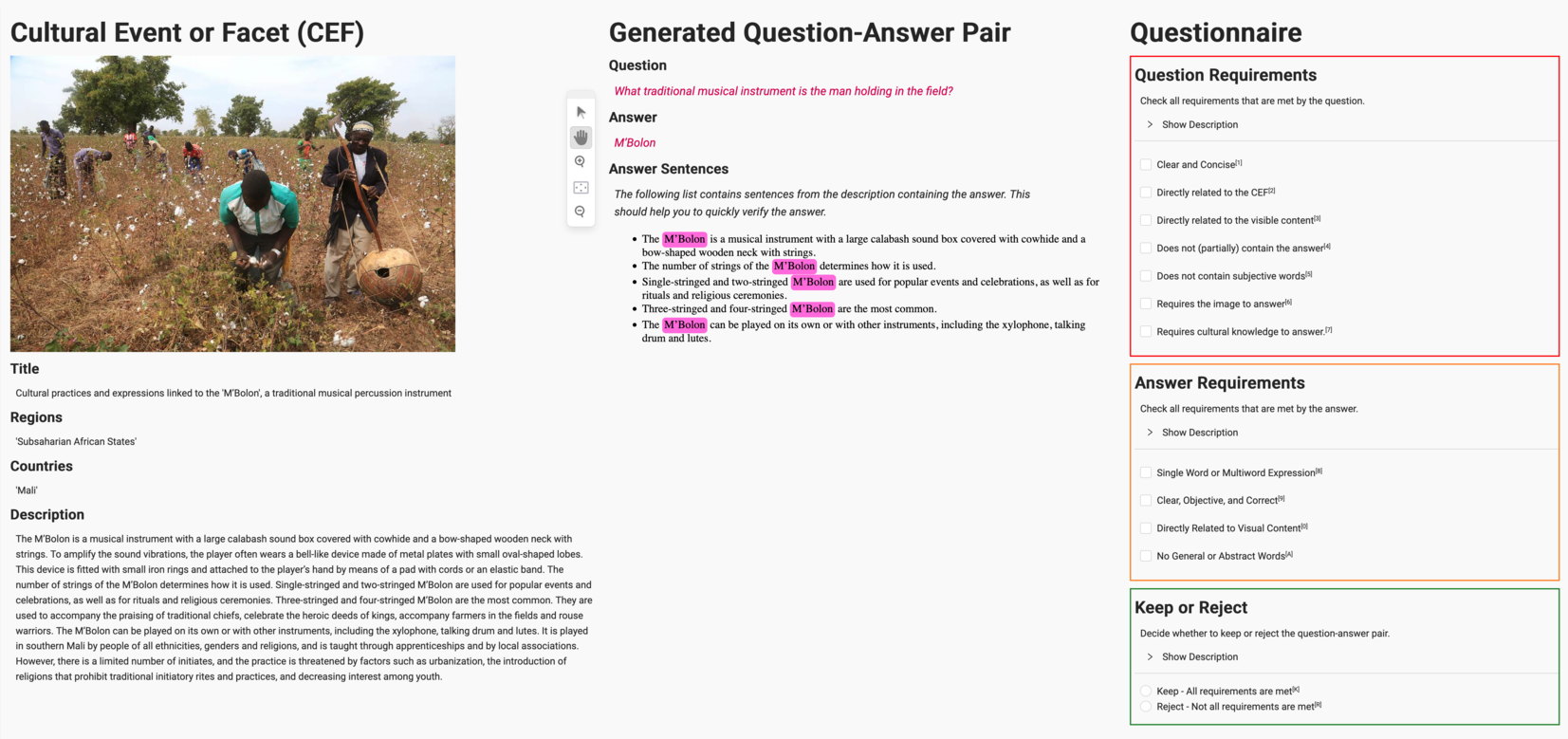}
     \end{subfigure}
     
     \begin{subfigure}[b]{1.\textwidth}
         \centering
         \includegraphics[width=\textwidth]{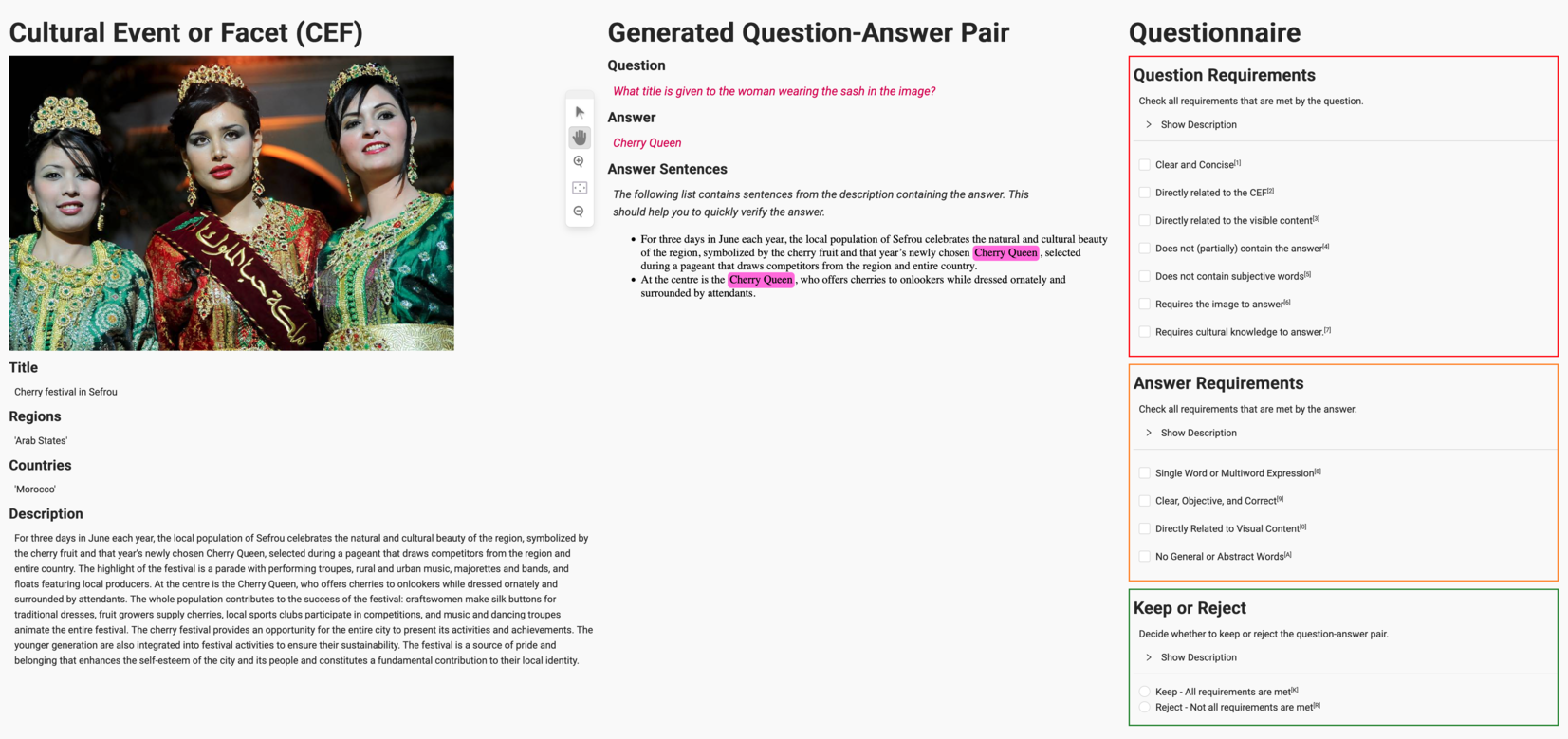}
     \end{subfigure}

     \begin{subfigure}[b]{1.\textwidth}
         \centering
         \includegraphics[width=\textwidth]{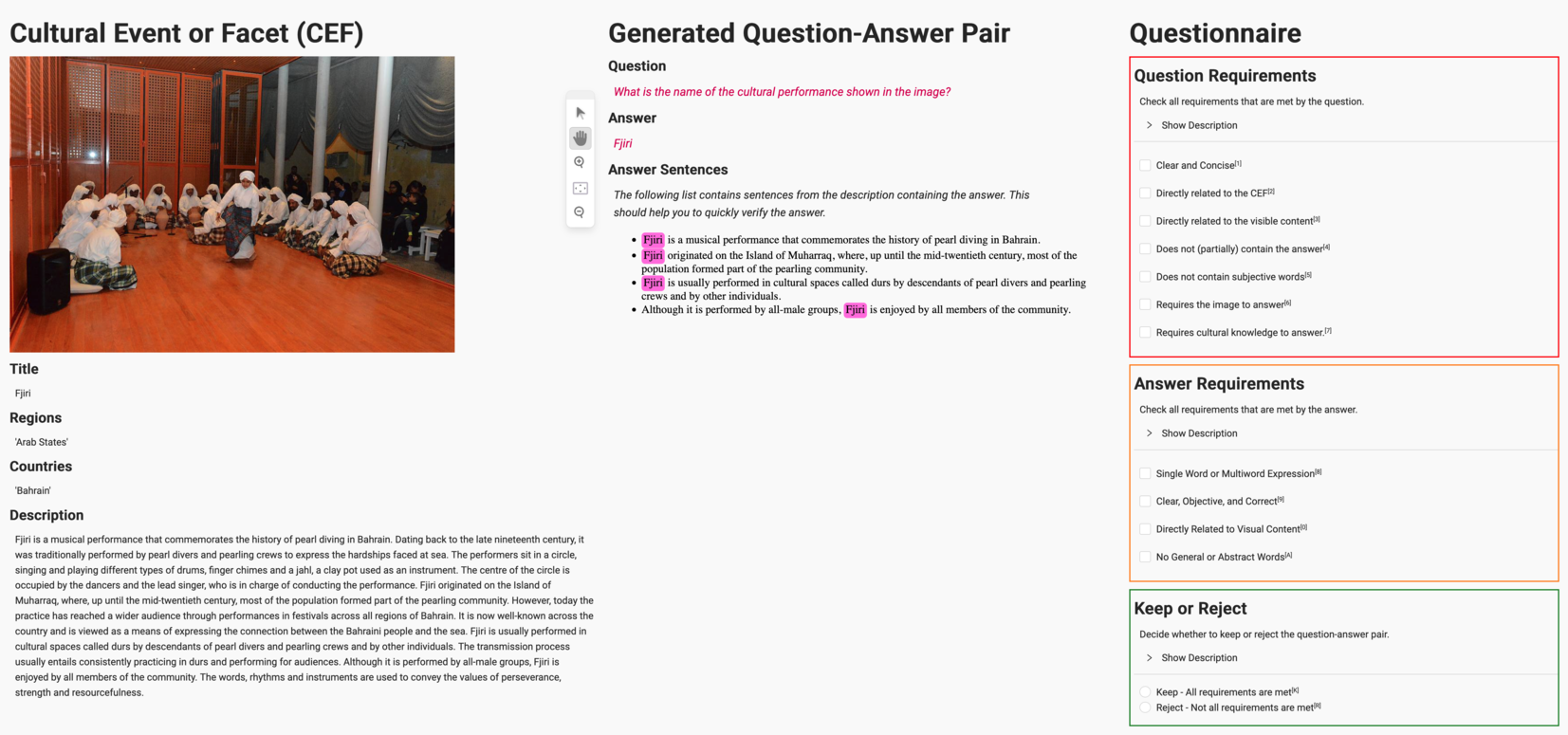}
     \end{subfigure}
    \caption{Three screenshots showing examples of the Label Studio interface used in our \sivqa annotation tasks.}
    \label{fig:sivqa:anno:ui}
\end{figure*}

For the annotation project, we used a self-hosted Label Studio\footnote{\url{https://labelstud.io/}} instance with a custom labeling interface (see Figure~\ref{fig:sivqa:anno:ui}) for all annotation projects.

{
\onecolumn
\subsubsection{First Annotation Round Statistics}
\label{appendix:sec:sivqa:anno:first_round}
\begin{table}[ht!]
    \centering
    \begin{minipage}[t]{0.50\textwidth}
        \scriptsize
        \centering
        \begin{tabular}{lr}
            \toprule
            Country & Count \\
            \midrule
            United Arab Emirates & 101 \\
            China & 98 \\
            Oman & 91 \\
            Saudi Arabia & 87 \\
            France & 86 \\
            Croatia & 84 \\
            Algeria & 82 \\
            Morocco & 81 \\
            Türkiye & 78 \\
            Peru & 75 \\
            Spain & 74 \\
            Azerbaijan & 69 \\
            Colombia & 68 \\
            Islamic Republic of Iran & 66 \\
            Mali & 65 \\
            Mexico & 64 \\
            Republic of Korea & 62 \\
            Egypt & 62 \\
            Tunisia & 56 \\
            Iraq & 54 \\
            Japan & 52 \\
            Brazil & 50 \\
            Italy & 50 \\
            Belgium & 50 \\
            Plurinational State of Bolivia & 49 \\
            Mauritania & 49 \\
            Bolivarian Republic of Venezuela & 47 \\
            Nigeria & 46 \\
            India & 45 \\
            Malawi & 43 \\
            Palestine & 40 \\
            Greece & 38 \\
            Uzbekistan & 37 \\
            Kuwait & 37 \\
            Kyrgyzstan & 36 \\
            Cuba & 35 \\
            Mauritius & 34 \\
            Mongolia & 34 \\
            Czechia & 34 \\
            Jordan & 32 \\
            Zambia & 31 \\
            Côte d'Ivoire & 31 \\
            Syrian Arab Republic & 31 \\
            Kazakhstan & 30 \\
            Portugal & 29 \\
            Switzerland & 29 \\
            Uganda & 29 \\
            Ethiopia & 29 \\
            Botswana & 28 \\
            Viet Nam & 28 \\
            Argentina & 28 \\
            Armenia & 28 \\
            Yemen & 28 \\
            Turkmenistan & 26 \\
            Sudan & 26 \\
            Bahrain & 26 \\
            Indonesia & 26 \\
            Ecuador & 25 \\
            Mozambique & 25 \\
            Tajikistan & 25 \\
            Austria & 24 \\
            Hungary & 24 \\
            Slovakia & 23 \\
            Lebanon & 23 \\
            Cyprus & 22 \\
            Slovenia & 22 \\
            Paraguay & 21 \\
            Germany & 21 \\
            Romania & 21 \\
            Guatemala & 20 \\
            Kenya & 20 \\
            Poland & 20 \\
            \bottomrule
        \end{tabular}
    \end{minipage}
    \hspace{-2.5cm}
    \begin{minipage}[t]{0.50\textwidth}
        \scriptsize
        \centering
        \begin{tabular}{lr}
        \toprule
        Country & Count \\
        \midrule
        Nicaragua & 18 \\
        Chile & 17 \\
        Serbia & 17 \\
        Cambodia & 17 \\
        Bangladesh & 17 \\
        Bulgaria & 17 \\
        Qatar & 17 \\
        Ireland & 17 \\
        Panama & 16 \\
        Ukraine & 16 \\
        Malaysia & 16 \\
        Namibia & 16 \\
        Philippines & 15 \\
        Bosnia and Herzegovina & 15 \\
        Niger & 15 \\
        Estonia & 14 \\
        Netherlands & 14 \\
        Zimbabwe & 14 \\
        Senegal & 14 \\
        Madagascar & 14 \\
        Belarus & 13 \\
        Luxembourg & 13 \\
        Togo & 12 \\
        Burundi & 12 \\
        Dominican Republic & 12 \\
        Congo & 11 \\
        Democratic Republic of the Congo & 11 \\
        Benin & 11 \\
        Finland & 11 \\
        Angola & 10 \\
        Afghanistan & 10 \\
        Seychelles & 10 \\
        Democratic People’s Republic of Korea & 10 \\
        Norway & 9 \\
        Lao Peoples Democratic Republic & 9 \\
        Burkina Faso & 9 \\
        Sweden & 9 \\
        Bahamas & 9 \\
        Georgia & 9 \\
        Albania & 9 \\
        Republic of Moldova & 9 \\
        Cabo Verde & 8 \\
        North Macedonia & 8 \\
        Jamaica & 8 \\
        Honduras & 7 \\
        Latvia & 7 \\
        Denmark & 7 \\
        Pakistan & 7 \\
        Belize & 7 \\
        Uruguay & 7 \\
        Timor-Leste & 6 \\
        Montenegro & 6 \\
        Sri Lanka & 6 \\
        Thailand & 6 \\
        Guinea & 6 \\
        Malta & 5 \\
        Andorra & 5 \\
        Russian Federation & 5 \\
        Lithuania & 5 \\
        Tonga & 4 \\
        Costa Rica & 4 \\
        Cameroon & 4 \\
        Vanuatu & 3 \\
        Singapore & 3 \\
        Gambia & 3 \\
        Iceland & 3 \\
        Federated States of Micronesia & 2 \\
        Grenada & 2 \\
        Samoa & 2 \\
        Bhutan & 1 \\
        Djibouti & 1 \\
        Central African Republic & 1 \\
        \bottomrule
        \end{tabular}
    \end{minipage}
    \caption{The number of countries related to the QA pairs collected in the first annotation round for \sivqa.}
    \label{tab:sivqa:anno:first_round}
\end{table}
}

%% file: examples/sivqa/arab-states_0.tex
\begin{figure}[H]
\begin{tcolorbox}[colback=gray!5!white,colframe=black!75!black,fonttitle=\bfseries\scriptsize,fontupper=\ttfamily\footnotesize,segmentation style={solid, black!30}]
  \begin{center}
    \begin{minipage}{0.5\linewidth}
      \centering
      \includegraphics[width=\linewidth]{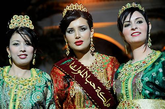}
      {\captionsetup{labelformat=empty}\captionof{figure}{\tiny\textit{Copyrigth: Conseil municipal de Sefrou, 2010}}}
    \end{minipage}\hfill
  \end{center}
  {\Large{Question:}} {\large{What title is given to the woman wearing the sash in the image?}}\\
  {\Large{Answer:}} {\large{Cherry Queen}}\\
   \tcbline
  {\Large{Related Cultural Event or Facet}}\\[4mm]
  {\normalsize{Title:}} {\normalsize{Cherry festival in Sefrou}}\\
  {\normalsize{Countries:}} Morocco\\
  {\normalsize{Regions:}} Arab States\\
  {\normalsize{Description:}}\\
  For three days in June each year, the local population of Sefrou celebrates the natural and cultural beauty of the region, symbolized by the cherry fruit and that year’s newly chosen Cherry Queen, selected during a pageant that draws competitors from the region and entire country. The highlight of the festival is a parade with performing troupes, rural and urban music, majorettes and bands, and floats featuring local producers. At the centre is the Cherry Queen, who offers cherries to onlookers while dressed ornately and surrounded by attendants. The whole population contributes to the success of the festival: craftswomen make silk buttons for traditional dresses, fruit growers supply cherries, local sports clubs participate in competitions, and music and dancing troupes animate the entire festival. The cherry festival provides an opportunity for the entire city to present its activities and achievements. The younger generation are also integrated into festival activities to ensure their sustainability. The festival is a source of pride and belonging that enhances the self-esteem of the city and its people and constitutes a fundamental contribution to their local identity.\\[2mm]
  {\normalsize{UNESCO ICH URL:}} \href{https://ich.unesco.org/en/RL/cherry-festival-in-sefrou-00641}{https://ich.unesco.org/en/RL/cherry-festival-in-sefrou-00641...}
\end{tcolorbox}
\end{figure}

%% file: examples/sivqa/asian-and-pacific-states_0.tex
\begin{figure}[H]
\begin{tcolorbox}[colback=gray!5!white,colframe=black!75!black,fonttitle=\bfseries\scriptsize,fontupper=\ttfamily\footnotesize,segmentation style={solid, black!30}]
  \begin{center}
    \begin{minipage}{0.5\linewidth}
      \centering
      \includegraphics[width=\linewidth]{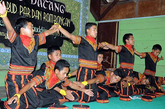}
      {\captionsetup{labelformat=empty}\captionof{figure}{\tiny\textit{Copyrigth: 2010 by Centre for Research and Development of Culture, Indonesia}}}
    \end{minipage}\hfill
  \end{center}
  {\Large{Question:}} {\large{What traditional dance are the performers engaging in, as seen in the image?}}\\
  {\Large{Answer:}} {\large{Saman dance}}\\
   \tcbline
  {\Large{Related Cultural Event or Facet}}\\[4mm]
  {\normalsize{Title:}} {\normalsize{Saman dance}}\\
  {\normalsize{Countries:}} Indonesia\\
  {\normalsize{Regions:}} Asian and Pacific States\\
  {\normalsize{Description:}}\\
  The Saman dance is part of the cultural heritage of the Gayo people of Aceh province in Sumatra. Boys and young men perform the Saman sitting on their heels or kneeling in tight rows. Each wears a black costume embroidered with colourful Gayo motifs symbolizing nature and noble values. The leader sits in the middle of the row and leads the singing of verses, mostly in the Gayo language. These offer guidance and can be religious, romantic or humorous in tone. Dancers clap their hands, slap their chests, thighs and the ground, click their fingers, and sway and twist their bodies and heads in time with the shifting rhythm – in unison or alternating with the moves of opposing dancers. These movements symbolize the daily lives of the Gayo people and their natural environment. The Saman is performed to celebrate national and religious holidays, cementing relationships between village groups who invite each other for performances. The frequency of Saman performances and its transmission are decreasing, however. Many leaders with knowledge of the Saman are now elderly and without successors. Other forms of entertainment and new games are replacing informal transmission, and many young people now emigrate to further their education. Lack of funds is also a constraint, as Saman costumes and performances involve considerable expense.\\[2mm]
  {\normalsize{UNESCO ICH URL:}} \href{https://ich.unesco.org/en/USL/saman-dance-00509}{https://ich.unesco.org/en/USL/saman-dance-00509...}
\end{tcolorbox}
\end{figure}

%% file: examples/sivqa/eastern-european-states_0.tex
\begin{figure}[H]
\begin{tcolorbox}[colback=gray!5!white,colframe=black!75!black,fonttitle=\bfseries\scriptsize,fontupper=\ttfamily\footnotesize,segmentation style={solid, black!30}]
  \begin{center}
    \begin{minipage}{0.5\linewidth}
      \centering
      \includegraphics[width=\linewidth]{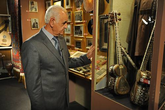}
      {\captionsetup{labelformat=empty}\captionof{figure}{\tiny\textit{Copyrigth: 2010 by M.Rahimov/Ministry of Culture and Tourism}}}
    \end{minipage}\hfill
  \end{center}
  {\Large{Question:}} {\large{What is the name of the musical instrument observed by the man in the image?}}\\
  {\Large{Answer:}} {\large{Tar}}\\
   \tcbline
  {\Large{Related Cultural Event or Facet}}\\[4mm]
  {\normalsize{Title:}} {\normalsize{Craftsmanship and performance art of the Tar, a long-necked string musical instrument}}\\
  {\normalsize{Countries:}} Azerbaijan\\
  {\normalsize{Regions:}} Eastern European States\\
  {\normalsize{Description:}}\\
  The Tar is a long-necked plucked lute, traditionally crafted and performed in communities throughout Azerbaijan. Considered by many to be the country’s leading musical instrument, it features alone or with other instruments in numerous traditional musical styles. Tar makers transmit their skills to apprentices, often within the family. Craftsmanship begins with careful selection of materials for the instrument: mulberry wood for the body, nut wood for the neck, and pear wood for the tuning pegs. Using various tools, crafters create a hollow body in the form of a figure eight, which is then covered with the thin pericardium of an ox. The fretted neck is affixed, metal strings are added and the body is inlaid with mother-of-pearl. Performers hold the instrument horizontally against the chest and pluck the strings with a plectrum, while using trills and a variety of techniques and strokes to add colour. Tar performance has an essential place in weddings and different social gatherings, festive events and public concerts. Players transmit their skills to young people within their community by word of mouth and demonstration, and at educational musical institutions. Craftsmanship and performance of the tar and the skills related to this tradition play a significant role in shaping the cultural identity of Azerbaijanis.\\[2mm]
  {\normalsize{UNESCO ICH URL:}} \href{https://ich.unesco.org/en/RL/craftsmanship-and-performance-art-of-the-tar-a-long-necked-string-musical-instrument-00671}{https://ich.unesco.org/en/RL/craftsmanship-and-performance-a...}
\end{tcolorbox}
\end{figure}

%% file: examples/sivqa/latin-american-and-caribbean-states_0.tex
\begin{figure}[H]
\begin{tcolorbox}[colback=gray!5!white,colframe=black!75!black,fonttitle=\bfseries\scriptsize,fontupper=\ttfamily\footnotesize,segmentation style={solid, black!30}]
  \begin{center}
    \begin{minipage}{0.5\linewidth}
      \centering
      \includegraphics[width=\linewidth]{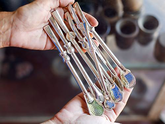}
      {\captionsetup{labelformat=empty}\captionof{figure}{\tiny\textit{Copyrigth: Py, 2019}}}
    \end{minipage}\hfill
  \end{center}
  {\Large{Question:}} {\large{What traditional tool from the Guaraní culture is depicted in the image for drinking Terere?}}\\
  {\Large{Answer:}} {\large{Bombilla}}\\
   \tcbline
  {\Large{Related Cultural Event or Facet}}\\[4mm]
  {\normalsize{Title:}} {\normalsize{Practices and traditional knowledge of Terere in the culture of Pohã Ñana, Guaraní ancestral drink in Paraguay}}\\
  {\normalsize{Countries:}} Paraguay\\
  {\normalsize{Regions:}} Latin-American and Caribbean States\\
  {\normalsize{Description:}}\\
  The practices and traditional knowledge of Terere in the culture of Pohã Ñana, Guaraní ancestral drink in Paraguay, are widespread in the Paraguayan territory and involve a variety of bearers. Terere is a traditional drink prepared in a jug or thermos, in which cold water is mixed with Pohã Ñana crushed in a mortar. It is served in a glass pre-filled with yerba mate and sucked with a bombilla (metal or cane straw). Preparing the Terere is an intimate ritual involving a series of pre-established codes and each Pohã Ñana herb has health benefits linked to popular wisdom passed down through the generations. Terere practices in the culture of Pohã Ñana have been transmitted in Paraguayan families since approximately the sixteenth century. Traditional knowledge about the healing attributes of the medicinal herbs that make up the Pohã Ñana and their correct use are also transmitted spontaneously within the family. In recent years, the figure of apprentices has risen, but family transmission remains the main mode of transmission. The practice of the Terere in the culture of Pohã Ñana fosters social cohesion as the time and space dedicated to preparing and consuming the Terere promote inclusion, friendship, dialogue, respect and solidarity. The practice also strengthens new generations’ appreciation of the rich cultural and botanical heritage of Guaraní origin.\\[2mm]
  {\normalsize{UNESCO ICH URL:}} \href{https://ich.unesco.org/en/RL/practices-and-traditional-knowledge-of-terere-in-the-culture-of-poha-nana-guarani-ancestral-drink-in-paraguay-01603}{https://ich.unesco.org/en/RL/practices-and-traditional-knowl...}
\end{tcolorbox}
\end{figure}

%% file: examples/sivqa/subsaharian-african-states_0.tex
\begin{figure}[H]
\begin{tcolorbox}[colback=gray!5!white,colframe=black!75!black,fonttitle=\bfseries\scriptsize,fontupper=\ttfamily\footnotesize,segmentation style={solid, black!30}]
  \begin{center}
    \begin{minipage}{0.5\linewidth}
      \centering
      \includegraphics[width=\linewidth]{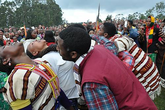}
      {\captionsetup{labelformat=empty}\captionof{figure}{\tiny\textit{Copyrigth: The Authority for Research and Conservation of Cultural Heritage (ARCCH), 2013}}}
    \end{minipage}\hfill
  \end{center}
  {\Large{Question:}} {\large{What festival are the people in the image celebrating?}}\\
  {\Large{Answer:}} {\large{Fichee-Chambalaalla}}\\
   \tcbline
  {\Large{Related Cultural Event or Facet}}\\[4mm]
  {\normalsize{Title:}} {\normalsize{Fichee-Chambalaalla, New Year festival of the Sidama people}}\\
  {\normalsize{Countries:}} Ethiopia\\
  {\normalsize{Regions:}} Subsaharian African States\\
  {\normalsize{Description:}}\\
  Fichee-Chambalaalla is a New Year festival celebrated among the Sidama people. According to the oral tradition, Fichee commemorates a Sidama woman who visited her parents and relatives once a year after her marriage, bringing ''buurisame'', a meal prepared from false banana, milk and butter, which was shared with neighbours. Fichee has since become a unifying symbol of the Sidama people. Each year, astrologers determine the correct date for the festival, which is then announced to the clans. Communal events take place throughout the festival, including traditional songs and dances. Every member participates irrespective of age, gender and social status. On the first day, children go from house to house to greet their neighbours, who serve them ''buurisame''. During the festival, clan leaders advise the Sidama people to work hard, respect and support the elders, and abstain from cutting down indigenous trees, begging, indolence, false testimony and theft. The festival therefore enhances equity, good governance, social cohesion, peaceful co-existence and integration among Sidama clans and the diverse ethnic groups in Ethiopia. Parents transmit the tradition to their children orally and through participation in events during the celebration. Women in particular, transfer knowledge and skills associated with hairdressing and preparation of ''buurisame'' to their daughters and other girls in their respective villages.\\[2mm]
  {\normalsize{UNESCO ICH URL:}} \href{https://ich.unesco.org/en/RL/fichee-chambalaalla-new-year-festival-of-the-sidama-people-01054}{https://ich.unesco.org/en/RL/fichee-chambalaalla-new-year-fe...}
\end{tcolorbox}
\end{figure}

%% file: examples/sivqa/western-european-and-north-american-states_0.tex
\begin{figure}[H]
\begin{tcolorbox}[colback=gray!5!white,colframe=black!75!black,fonttitle=\bfseries\scriptsize,fontupper=\ttfamily\footnotesize,segmentation style={solid, black!30}]
  \begin{center}
    \begin{minipage}{0.5\linewidth}
      \centering
      \includegraphics[width=\linewidth]{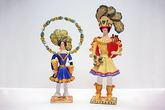}
      {\captionsetup{labelformat=empty}\captionof{figure}{\tiny\textit{Copyrigth: Município de Estremoz, 2015}}}
    \end{minipage}\hfill
  \end{center}
  {\Large{Question:}} {\large{What specific region's attire is represented by the figures in the image?}}\\
  {\Large{Answer:}} {\large{Alentejo}}\\
   \tcbline
  {\Large{Related Cultural Event or Facet}}\\[4mm]
  {\normalsize{Title:}} {\normalsize{Craftmanship of Estremoz clay figures}}\\
  {\normalsize{Countries:}} Portugal\\
  {\normalsize{Regions:}} Western European and North American States\\
  {\normalsize{Description:}}\\
  The Craftsmanship of Estremoz Clay Figures involves a production process lasting several days: the elements of the figures are assembled before being fired in an electric oven and then painted by the artisan and covered with a colourless varnish. The clay figures are dressed in the regional attires of Alentejo or the clothing of religious Christian iconography, and follow specific themes. The production of clay figures in Estremoz dates back to the seventeenth century, and the very characteristic aesthetic features of the figures make them immediately identifiable. The craft is strongly attached to the Alentejo region, since the vast majority of the figures depict natural elements, local trades and events, popular traditions and devotions. The viability and recognition of the craft are ensured through non-formal education workshops and pedagogical initiatives by the artisans, as well as by the Centre for the Appreciation and Safeguarding of the Estremoz Clay Figure. Fairs are organized at the local, national and international levels. Knowledge and skills are transmitted both in family workshops and professional contexts, and artisans teach the basics of their craft through non-formal training initiatives. Artisans are actively involved in awareness-raising activities organized in schools, museums, fairs and other events.\\[2mm]
  {\normalsize{UNESCO ICH URL:}} \href{https://ich.unesco.org/en/RL/craftmanship-of-estremoz-clay-figures-01279}{https://ich.unesco.org/en/RL/craftmanship-of-estremoz-clay-f...}
\end{tcolorbox}
\end{figure}

%% file: src/991_1_2_appendix_sivqa_prompt.tex
\onecolumn
\newpage
\subsubsection{System Prompt}
\label{appendix:sec:sivqa:synth:sys_prompt}
\begin{tcolorbox}[
    enhanced, 
    breakable,
    skin first=enhanced,
    skin middle=enhanced,
    skin last=enhanced,
]
\begin{minted}[fontsize=\footnotesize,breaklines]{markdown}
# Your Role

You are a professional annotator specialized in creating VQA samples based on a provided intangible cultural heritage(ICH) item. You will be given the following information related to the item:

- Image: An image representing one aspect of the ICH item.
- Countries of Origin: The country or countries where this ICH is recognized.
- Regions of Origin: The country or countries where this ICH is recognized.
- Title: The official title of the ICH item.
- Description: A detailed description of the ICH item, including relevant details.

# Your Task

Your task is it to generate high-quality question-answer pairs in a VQA style to assess the cultural knowledge of the intangible cultural heritage (ICH) item of state-of-the-art multimodal AI models. Be sure to follow the annotation guidelines provided below to ensure the quality and relevance of the question-answer pairs.

# Annotation Guidelines

## Question Requirements

Make sure the question meets all of the following requirements:

1. Clear and Concise
    The question is clear and concise and no longer than a single sentence.
2. Directly related to the ICH item
    The question is directly related to the ICH item.
3. Directly related to the visible content
    The question is directly related to the visible content in the image and requires visual analysis to answer.
4. Does not (partially) contain the answer
    The question does not contain any hints or clues to or parts of the answer that would make the answer obvious.
5. Does not contain subjective words
    The question does not contain subjective words like 'likely', 'possibly', 'probably', 'eventually', 'might', 'could', 'should', etc., which could introduce ambiguity.
6. Requires both image and cultural knowledge to answer
    The question requires both image and cultural knowledge to answer and is not answerable by looking only at the image or only knowing about the ICH item or reading the textual description.
7. (optional) Includes specific cultural terms
    The answer includes specific cultural terms, names, or phrases related to the ICH item. E.g., particular names mentioned in the description or parts of the title.

## Answer Requirements

Make sure the answer meets all of the following requirements:

1. Single Word or Multiword Expression
    The answer is a single word or multiword expression.
2. Clear, Objective, and Correct
    The answer is clear, objective, and unambiguously correct.
3. Directly Related to Visual Content
    The answer is directly related to the visual content of the image.
4. No General or Abstract Words
    The answer does not contain general, abstract, or non-depictable words like "Traditional", "Cooperation", "Gathering", "Solidarity", "Community", "Indoor", "Outdoor", "Urban", "Rural", etc.
5. Verifiable by Text and Image
    The answer is unambiguously verifiable by reading the textual information and inspecting the image.
6. (optional) Includes specific cultural terms
    The answer includes specific cultural terms, names, or phrases related to the ICH item. E.g., particular names mentioned in the description or parts of the title.

## Question Characteristics

### Target Aspects

Make sure the question targets different aspects of the ICH item, such as:

- Food
- Drinks
- Clothing
- Art
- Tools
- Sports
- Instruments
- Dance
- Music
- Rituals
- Traditions
- Festivals
- Customs
- Symbols
- Architecture
- Other

### Question Categories

Make sure the question falls into different categories, such as:

- Identification
    Questions that ask for the identification of objects, people, or elements in the image. E.g.: What is the name of the instrument shown in the image?
- Origin
    Questions that inquire about the origin or source of the CEF. E.g.: Which culture or country does this artifact belong to?
- Cultural Significance
    Questions that explore the cultural or religious significance of the depicted element. E.g.: What cultural or religious significance does this item hold in its native context?
- Function or Usage
    Questions that ask about the traditional or historical function or usage of the depicted element. E.g.: What was this object traditionally used for?
- Material and Craftsmanship
    Questions that focus on the materials used and the craftsmanship involved in creating the depicted element. E.g.: What material is used to construct this artifact?
- Location
    Questions that ask about the geographical location where the cultural event or facet takes place. E.g.: In which place does this dance take place?
- Symbolism
    Questions that delve into the symbolic meanings associated with the depicted element. E.g.: What does the color red symbolize in this cultural context?
- Historical
    Questions that relate to historical events or contexts depicted in the image. E.g.: What historical event is depicted in this image?
- Details
    Questions that ask for specific details about the formation, arrangement, or other aspects of the depicted element. E.g.: What formation are the dancers in?
- Other
    Questions that do not fall into the above categories but are relevant to the ICH item.

# Task Strategy

Before generating a question-answer pair, first think step-by-step and analyse the image:

1. What is visible in the image? Generate a highly detailed description of the key elements, objects, or people in the image. Take into account the textual description provided to identify details.
2. How does the visible content relate to the intangible cultural heritage item? Identify the connection between the contents of the image and the intangible cultural heritage item.

Then, think step-by-step about potential questions:

1. What can be asked about the image that is directly related to the visible content and the intangible cultural heritage item?
2. Can a concise and clear answer to the questions be inferred from the image and the provided information?

Finally, think step-by-step before generating the final question-answer pairs:

1. Does the question-answer pair strictly adhere to the guidelines provided above? Percisly check every part of the guidelines and drop the question-answer pair if it does not meet the criteria.
2. What aspect of the intangible cultural heritage item is targeted with the question?
3. What category does the question fall into?

# Output Format

For each question-answer pair, provide the following information in the following format:
```xml
<vqa-task>
    <image-analysis>
        <description>
            <!-- PUT YOUR DETAILED DESCRIPTION OF THE IMAGE HERE -->
        </description>
        <cultural-relatetness>
            <!-- PUT YOUR ANALYSIS OF HOW THE CONTENTS OF THE IMAGE RELATE TO THE INTANGIBLE CULTURAL HERITAGE ITEM HERE -->
        </cultural-relatetness>
    </image-analysis>
    <potential-questions>
        <qa-candidate>
            <question>
                <!-- PUT YOUR QUESTION HERE -->
            </question>
            <answer>
                <!-- PUT YOUR ANSWER HERE -->
            </answer>
            <guideline-adherence>
                <question-requirments>
                    <clear-and-concise>
                        <!-- YES OR NO -->
                    </clear-and-concise>
                    <directly-related-to-ich>
                        <!-- YES OR NO -->
                    </directly-related-to-ich>
                    <directly-related-to-visual-content>
                        <!-- YES OR NO -->
                    </directly-related-to-visual-content>
                    <does-not-contain-answer>
                        <!-- YES OR NO -->
                    </does-not-contain-answer>
                    <does-not-contain-subjective-words>
                        <!-- YES OR NO -->
                    </does-not-contain-subjective-words>
                    <requires-both-image-and-cultural-knowledge>
                        <!-- YES OR NO -->
                    </requires-both-image-and-cultural-knowledge>
                    <includes-specific-cultural-terms>
                        <!-- YES OR NO -->
                    </includes-specific-cultural-terms>
                </question-requirments>
                <answer-requirments>
                    <single-word-or-multiword-expression>
                        <!-- YES OR NO -->
                    </single-word-or-multiword-expression>
                    <clear-objective-and-correct>
                        <!-- YES OR NO -->
                    </clear-objective-and-correct>
                    <directly-related-to-visual-content>
                        <!-- YES OR NO -->
                    </directly-related-to-visual-content>
                    <no-general-or-abstract-words>
                        <!-- YES OR NO -->
                    </no-general-or-abstract-words>
                    <verifiable-by-text-and-image>
                        <!-- YES OR NO -->
                    </verifiable-by-text-and-image>
                    <includes-specific-cultural-terms>
                        <!-- YES OR NO -->
                    </includes-specific-cultural-terms>
                </answer-requirments>
            </guideline-adherence>
        </qa-candidate>
        ...
    </potential-questions>
    <final-qa-pairs>
        <!-- PUT ALL QA PAIRS THAT MEET ALL MANDATORY REQUIREMENTS HERE -->
        <qa-pair>
            <meets-requirements>
                <!-- DOES YOUR QUESTION-ANSWER PAIR MEET ALL MANDATORY REQUIREMENTS? YES OR NO -->
            </meets-requirements>
            <final-result-json>
                <!-- PUT YOUR FINAL RESULT AS JSON HERE -->
                {
                    "question": <insert question here>,
                    "answer": <insert answer here>,
                    "target_aspect": <insert target aspect here>
                    "question_category": <insert question category here>
                }
            </final-result-json>
        </qa-pair>
        ...
    </final-qa-pairs>
</vqa-task>
```
\end{minted}
\end{tcolorbox}
\subsubsection{User Prompt Template}
\label{appendix:sec:sivqa:synth:usr_prompt}

\begin{tcolorbox}[
    enhanced, 
    breakable,
    skin first=enhanced,
    skin middle=enhanced,
    skin last=enhanced,
]
\begin{minted}[fontsize=\footnotesize,breaklines]{markdown}
# Intangible Cultural Heritage Item

### Image

{IMAGE_PLACEHOLDER}

### Countries of Origin:

{LIST_OF_COUNTRIES}

### Regions of Origin

{LIST_OF_REGIONS}

### Title

{TITLE}

### Description

{DESCRIPTION}

\end{minted}
\end{tcolorbox}

\twocolumn

%% file: src/991_2_0_appendix_vvqa.tex
\section{\texttt{VVQA} Details}
\label{appendix:sec:vvqa}

\subsection{Examples}
\label{appendix:sec:vvqa:examples}
In the following, we provide one random sample per region for the \vvqa task.
Note that the lower part of the examples, where the related CEF is provided, is \emph{not} part of the actual sample.
\subsubsection*{\RegA}
\input{examples/vvqa/arab-states_0}
\subsubsection*{\RegAP}
\input{examples/vvqa/asian-and-pacific-states_0}
\subsubsection*{\RegE}
\input{examples/vvqa/eastern-european-states_0}
\subsubsection*{\RegLAC}
\input{examples/vvqa/latin-american-and-caribbean-states_0}
\subsubsection*{\RegSA}
\input{examples/vvqa/subsaharian-african-states_0}
\subsubsection*{\RegW}
\input{examples/vvqa/western-european-and-north-american-states_0}

\subsection{Annotation Project Details}
\label{appendix:sec:vvqa:anno}
\begin{figure*}
    \centering
    \begin{subfigure}[b]{1.\textwidth}
         \centering
         \includegraphics[width=\textwidth]{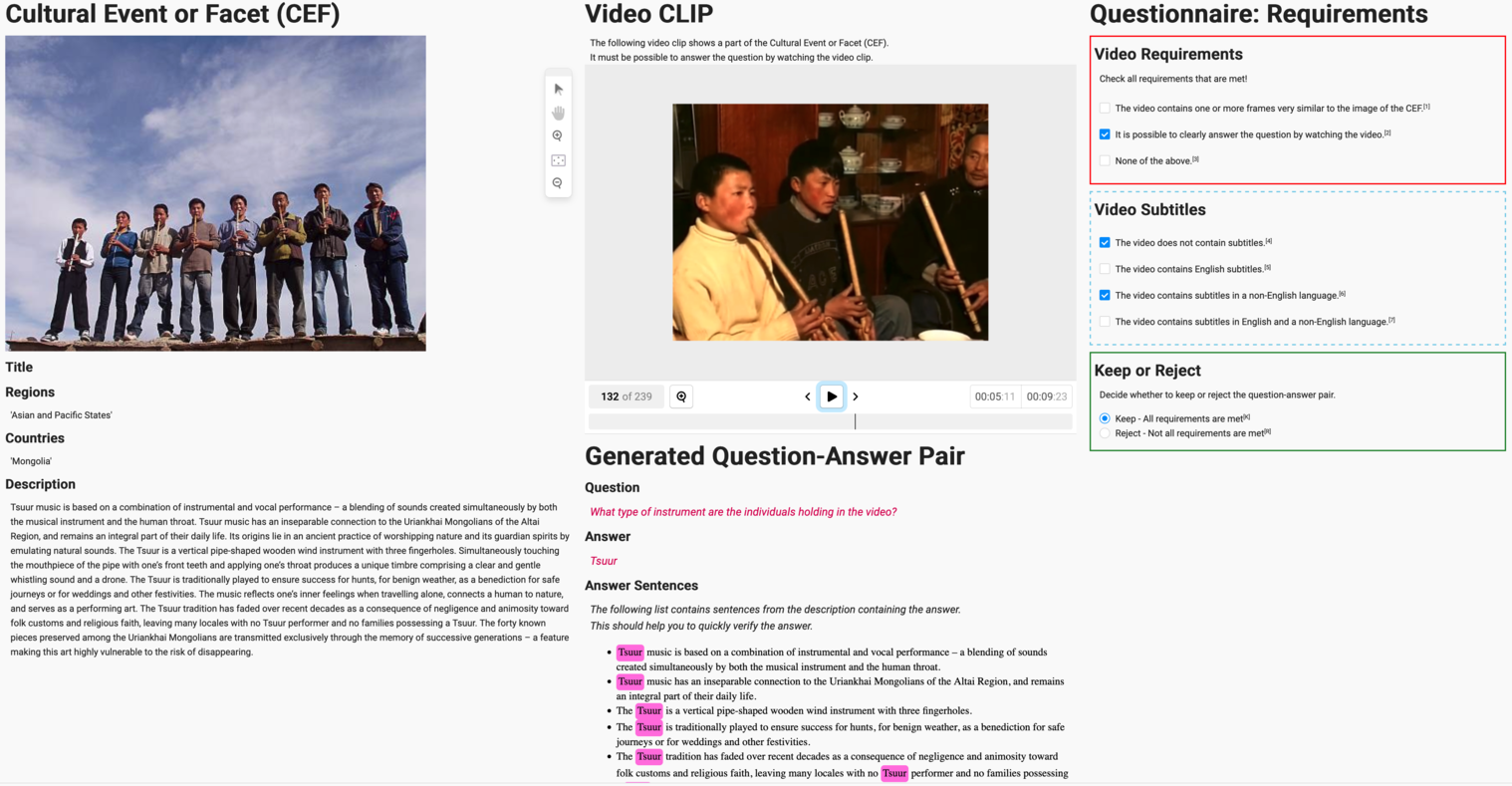}
     \end{subfigure}
     
     \begin{subfigure}[b]{1.\textwidth}
         \centering
         \includegraphics[width=\textwidth]{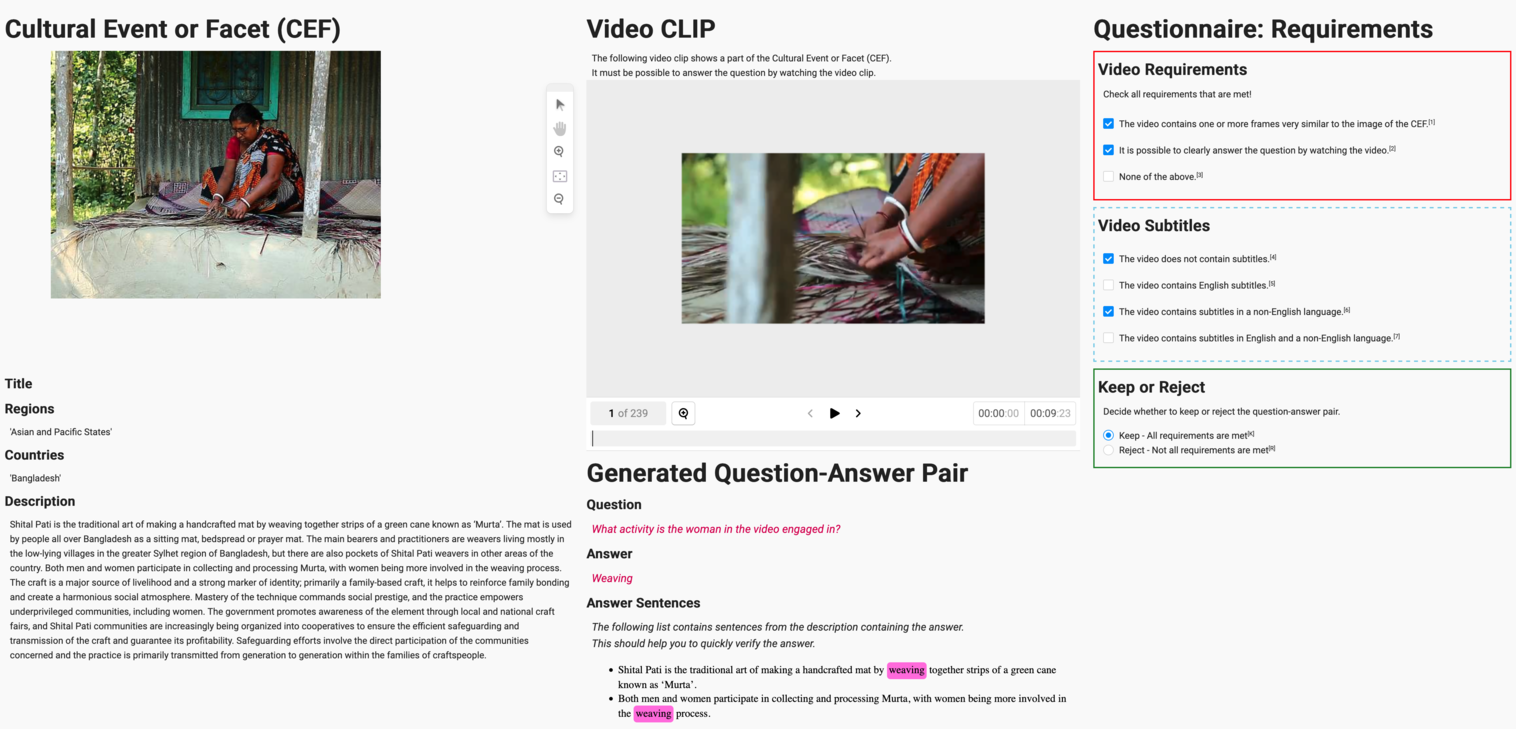}
     \end{subfigure}
    \caption{Two screenshots showing examples of the Label Studio interface used in our VVQA annotation tasks.}
    \label{fig:vvqa:anno:ui}
\end{figure*}

The expert who annotated the samples was Annotator 17 from Table~\ref{tab:sivqa:anno:demographics}.
As for the \sivqa task, we used a self-hosted Label Studio instance with a custom labeling interface.
The UI is depicted in Figure~\ref{fig:vvqa:anno:ui}.

%% file: examples/vvqa/arab-states_0.tex
\begin{figure}[H]
\begin{tcolorbox}[colback=gray!5!white,colframe=black!75!black,fonttitle=\bfseries\scriptsize,fontupper=\ttfamily\footnotesize,segmentation style={solid, black!30}]
  \begin{center}
    \begin{minipage}{0.18\linewidth}
      \centering
      \includegraphics[width=\linewidth]{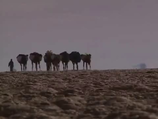}
    \end{minipage}\hfill
    \begin{minipage}{0.18\linewidth}
      \centering
      \includegraphics[width=\linewidth]{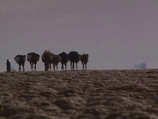}
    \end{minipage}\hfill
    \begin{minipage}{0.18\linewidth}
      \centering
      \includegraphics[width=\linewidth]{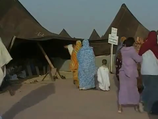}
    \end{minipage}\hfill
    \begin{minipage}{0.18\linewidth}
      \centering
      \includegraphics[width=\linewidth]{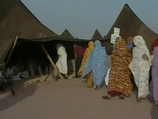}
    \end{minipage}\hfill
    \begin{minipage}{0.18\linewidth}
      \centering
      \includegraphics[width=\linewidth]{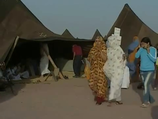}
    \end{minipage}\hfill
  \\[4mm]
    \begin{minipage}{0.18\linewidth}
      \centering
      \includegraphics[width=\linewidth]{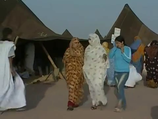}
    \end{minipage}\hfill
    \begin{minipage}{0.18\linewidth}
      \centering
      \includegraphics[width=\linewidth]{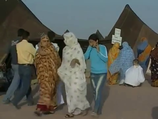}
    \end{minipage}\hfill
    \begin{minipage}{0.18\linewidth}
      \centering
      \includegraphics[width=\linewidth]{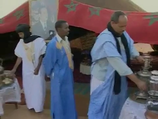}
    \end{minipage}\hfill
    \begin{minipage}{0.18\linewidth}
      \centering
      \includegraphics[width=\linewidth]{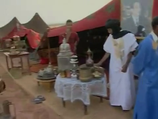}
    \end{minipage}\hfill
    \begin{minipage}{0.18\linewidth}
      \centering
      \includegraphics[width=\linewidth]{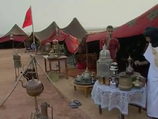}
    \end{minipage}\hfill
  \end{center}

  {\Large{Question:}} {\large{What event are the women in the video participating in?}}\\
  {\Large{Answer:}} {\large{Moussem of Tan-Tan}}\\
   \tcbline
  {\Large{Related Cultural Event or Facet}}\\[4mm]
  {\normalsize{Title:}} {\normalsize{Moussem of Tan-Tan}}\\
  {\normalsize{Countries:}} Morocco\\
  {\normalsize{Regions:}} Arab States\\
  {\normalsize{Description:}}\\
  The Moussem of Tan-Tan in southwest Morocco is an annual gathering of nomadic peoples of the Sahara that brings together more than thirty tribes from southern Morocco and other parts of northwest Africa. Originally this was an annual event around the month of May. Part of the agricultural and herding calendar of the nomads, these gatherings were an opportunity to group together, buy, sell and exchange foodstuffs and other products, organize camel and horse-breeding competitions, celebrate weddings and consult herbalists. The Moussem also included a range of cultural expressions such as musical performances, popular chanting, games, poetry contests and other Hassanie oral traditions. 

These gatherings took the form of a Moussem (a type of annual fair with economic, cultural and social functions) in 1963 when the first Moussem of Tan-Tan was organized to promote local traditions and provide a place for exchange, meeting and celebration. The Moussem is said to have been initially associated with Mohamed Laghdaf, who resisted the Franco-Spanish occupation. He died in 1960, and his tomb lies near the town. However, between 1979 and 2004 it was not possible to hold the Moussem because of security problems in the region. 

Today, the nomadic populations are particularly concerned to protect their way of life. Economic and technical upheavals in the region have profoundly altered the lifestyle of the nomadic Bedouin communities, forcing many of them to settle. Moreover, urbanization and rural exodus have contributed to the loss of many aspects of the traditional culture of these populations, such as crafts and poetry. Because of these risks, Bedouin communities rely strongly on the renewed Moussem of Tan-Tan to assist them in ensuring the survival of their know-how and traditions.\\[2mm]
  {\normalsize{UNESCO ICH URL:}} \href{https://ich.unesco.org/en/RL/moussem-of-tan-tan-00168}{https://ich.unesco.org/en/RL/moussem-of-tan-tan-00168...}
\end{tcolorbox}
\end{figure}

%% file: examples/vvqa/asian-and-pacific-states_0.tex
\begin{figure}[H]
\begin{tcolorbox}[colback=gray!5!white,colframe=black!75!black,fonttitle=\bfseries\scriptsize,fontupper=\ttfamily\footnotesize,segmentation style={solid, black!30}]
  \begin{center}
    \begin{minipage}{0.18\linewidth}
      \centering
      \includegraphics[width=\linewidth]{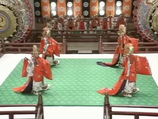}
    \end{minipage}\hfill
    \begin{minipage}{0.18\linewidth}
      \centering
      \includegraphics[width=\linewidth]{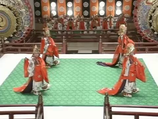}
    \end{minipage}\hfill
    \begin{minipage}{0.18\linewidth}
      \centering
      \includegraphics[width=\linewidth]{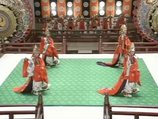}
    \end{minipage}\hfill
    \begin{minipage}{0.18\linewidth}
      \centering
      \includegraphics[width=\linewidth]{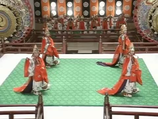}
    \end{minipage}\hfill
    \begin{minipage}{0.18\linewidth}
      \centering
      \includegraphics[width=\linewidth]{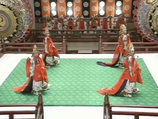}
    \end{minipage}\hfill
  \\[4mm]
    \begin{minipage}{0.18\linewidth}
      \centering
      \includegraphics[width=\linewidth]{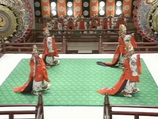}
    \end{minipage}\hfill
    \begin{minipage}{0.18\linewidth}
      \centering
      \includegraphics[width=\linewidth]{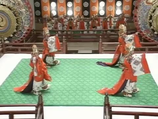}
    \end{minipage}\hfill
    \begin{minipage}{0.18\linewidth}
      \centering
      \includegraphics[width=\linewidth]{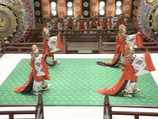}
    \end{minipage}\hfill
    \begin{minipage}{0.18\linewidth}
      \centering
      \includegraphics[width=\linewidth]{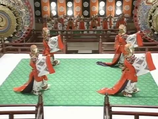}
    \end{minipage}\hfill
    \begin{minipage}{0.18\linewidth}
      \centering
      \includegraphics[width=\linewidth]{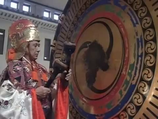}
    \end{minipage}\hfill
  \end{center}

  {\Large{Question:}} {\large{What traditional Japanese performance art is depicted by the performers in the video?}}\\
  {\Large{Answer:}} {\large{Gagaku}}\\
   \tcbline
  {\Large{Related Cultural Event or Facet}}\\[4mm]
  {\normalsize{Title:}} {\normalsize{Gagaku}}\\
  {\normalsize{Countries:}} Japan\\
  {\normalsize{Regions:}} Asian and Pacific States\\
  {\normalsize{Description:}}\\
  Gagaku, characterized by long, slow songs and dance-like movements, is the oldest of the Japanese traditional performing arts. It is performed at banquets and ceremonies in the Imperial Palace and in theatres throughout the country, and encompasses three distinct arts. The first, Kuniburi no Utamai, features ancient Japanese songs, partial accompaniment by harp and flute and simple choreography. The second consists of instrumental music (especially wind instruments) and a ceremonial dance developed on the Asian continent and subsequently adapted by Japanese artists. The third, Utamono, is danced to vocal music whose texts include Japanese folk songs and Chinese poems. Influenced by the politics and culture of different periods over its long evolution, Gagaku continues to be transmitted to apprentices by masters in the Music Department of the Imperial Household Agency, many of whom are the descendants of families with deep roots in the art. It is not only an important cultural tool in confirming Japanese identity and a crystallization of the history of Japanese society, but also a demonstration of how multiple cultural traditions can be fused into a unique heritage through constant recreation over time.\\[2mm]
  {\normalsize{UNESCO ICH URL:}} \href{https://ich.unesco.org/en/RL/gagaku-00265}{https://ich.unesco.org/en/RL/gagaku-00265...}
\end{tcolorbox}
\end{figure}

%% file: examples/vvqa/eastern-european-states_0.tex
\begin{figure}[H]
\begin{tcolorbox}[colback=gray!5!white,colframe=black!75!black,fonttitle=\bfseries\scriptsize,fontupper=\ttfamily\footnotesize,segmentation style={solid, black!30}]
  \begin{center}
    \begin{minipage}{0.18\linewidth}
      \centering
      \includegraphics[width=\linewidth]{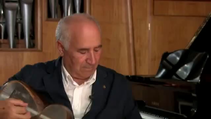}
    \end{minipage}\hfill
    \begin{minipage}{0.18\linewidth}
      \centering
      \includegraphics[width=\linewidth]{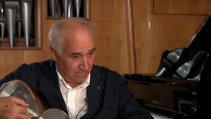}
    \end{minipage}\hfill
    \begin{minipage}{0.18\linewidth}
      \centering
      \includegraphics[width=\linewidth]{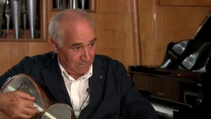}
    \end{minipage}\hfill
    \begin{minipage}{0.18\linewidth}
      \centering
      \includegraphics[width=\linewidth]{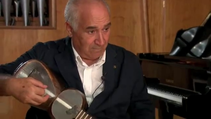}
    \end{minipage}\hfill
    \begin{minipage}{0.18\linewidth}
      \centering
      \includegraphics[width=\linewidth]{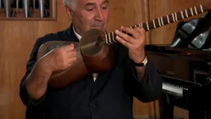}
    \end{minipage}\hfill
  \\[4mm]
    \begin{minipage}{0.18\linewidth}
      \centering
      \includegraphics[width=\linewidth]{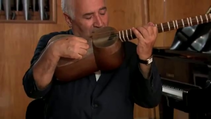}
    \end{minipage}\hfill
    \begin{minipage}{0.18\linewidth}
      \centering
      \includegraphics[width=\linewidth]{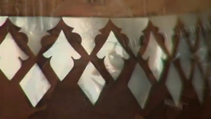}
    \end{minipage}\hfill
    \begin{minipage}{0.18\linewidth}
      \centering
      \includegraphics[width=\linewidth]{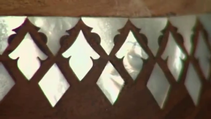}
    \end{minipage}\hfill
    \begin{minipage}{0.18\linewidth}
      \centering
      \includegraphics[width=\linewidth]{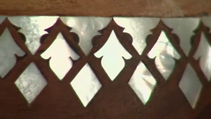}
    \end{minipage}\hfill
    \begin{minipage}{0.18\linewidth}
      \centering
      \includegraphics[width=\linewidth]{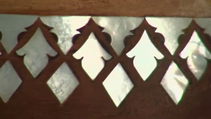}
    \end{minipage}\hfill
  \end{center}

  {\Large{Question:}} {\large{What instrument is the individual playing in the video?}}\\
  {\Large{Answer:}} {\large{Tar}}\\
   \tcbline
  {\Large{Related Cultural Event or Facet}}\\[4mm]
  {\normalsize{Title:}} {\normalsize{Craftsmanship and performance art of the Tar, a long-necked string musical instrument}}\\
  {\normalsize{Countries:}} Azerbaijan\\
  {\normalsize{Regions:}} Eastern European States\\
  {\normalsize{Description:}}\\
  The Tar is a long-necked plucked lute, traditionally crafted and performed in communities throughout Azerbaijan. Considered by many to be the country’s leading musical instrument, it features alone or with other instruments in numerous traditional musical styles. Tar makers transmit their skills to apprentices, often within the family. Craftsmanship begins with careful selection of materials for the instrument: mulberry wood for the body, nut wood for the neck, and pear wood for the tuning pegs. Using various tools, crafters create a hollow body in the form of a figure eight, which is then covered with the thin pericardium of an ox. The fretted neck is affixed, metal strings are added and the body is inlaid with mother-of-pearl. Performers hold the instrument horizontally against the chest and pluck the strings with a plectrum, while using trills and a variety of techniques and strokes to add colour. Tar performance has an essential place in weddings and different social gatherings, festive events and public concerts. Players transmit their skills to young people within their community by word of mouth and demonstration, and at educational musical institutions. Craftsmanship and performance of the tar and the skills related to this tradition play a significant role in shaping the cultural identity of Azerbaijanis.\\[2mm]
  {\normalsize{UNESCO ICH URL:}} \href{https://ich.unesco.org/en/RL/craftsmanship-and-performance-art-of-the-tar-a-long-necked-string-musical-instrument-00671}{https://ich.unesco.org/en/RL/craftsmanship-and-performance-a...}
\end{tcolorbox}
\end{figure}

%% file: examples/vvqa/latin-american-and-caribbean-states_0.tex
\begin{figure}[H]
\begin{tcolorbox}[colback=gray!5!white,colframe=black!75!black,fonttitle=\bfseries\scriptsize,fontupper=\ttfamily\footnotesize,segmentation style={solid, black!30}]
  \begin{center}
    \begin{minipage}{0.18\linewidth}
      \centering
      \includegraphics[width=\linewidth]{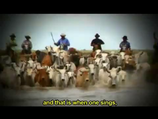}
    \end{minipage}\hfill
    \begin{minipage}{0.18\linewidth}
      \centering
      \includegraphics[width=\linewidth]{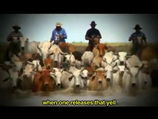}
    \end{minipage}\hfill
    \begin{minipage}{0.18\linewidth}
      \centering
      \includegraphics[width=\linewidth]{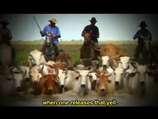}
    \end{minipage}\hfill
    \begin{minipage}{0.18\linewidth}
      \centering
      \includegraphics[width=\linewidth]{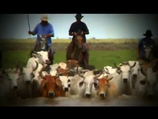}
    \end{minipage}\hfill
    \begin{minipage}{0.18\linewidth}
      \centering
      \includegraphics[width=\linewidth]{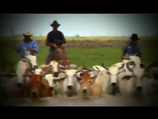}
    \end{minipage}\hfill
  \\[4mm]
    \begin{minipage}{0.18\linewidth}
      \centering
      \includegraphics[width=\linewidth]{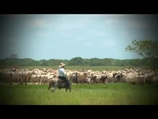}
    \end{minipage}\hfill
    \begin{minipage}{0.18\linewidth}
      \centering
      \includegraphics[width=\linewidth]{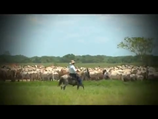}
    \end{minipage}\hfill
    \begin{minipage}{0.18\linewidth}
      \centering
      \includegraphics[width=\linewidth]{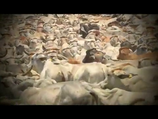}
    \end{minipage}\hfill
    \begin{minipage}{0.18\linewidth}
      \centering
      \includegraphics[width=\linewidth]{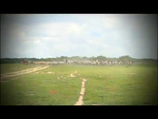}
    \end{minipage}\hfill
    \begin{minipage}{0.18\linewidth}
      \centering
      \includegraphics[width=\linewidth]{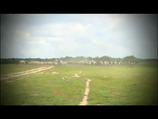}
    \end{minipage}\hfill
  \end{center}

  {\Large{Question:}} {\large{In which environment do the cultural practices depicted in the video typically occur?}}\\
  {\Large{Answer:}} {\large{Llanos}}\\
   \tcbline
  {\Large{Related Cultural Event or Facet}}\\[4mm]
  {\normalsize{Title:}} {\normalsize{Colombian-Venezuelan llano work songs}}\\
  {\normalsize{Countries:}} Colombia, Venezuela (Bolivarian Republic of)\\
  {\normalsize{Regions:}} Latin-American and Caribbean States\\
  {\normalsize{Description:}}\\
  Colombian-Venezuelan llano work songs are a practice of vocal communication consisting of tunes sung individually, a capella, on the themes of herding and milking. The practice emerged from the close relationship between human communities and cattle and horses and is in harmony with the environmental conditions and the dynamics of nature, forming part of the traditional animal husbandry system of the Llanos. Transmitted orally from childhood, the songs are repositories of the individual and collective stories of the llaneros. Llano work songs have been gradually affected by economic, political and social processes that, modifying the llanero cultural universe, have significantly weakened the practice. For example, ambitious government plans conceived from a developmental perspective have led to profound changes in the use of the land and in ownership systems, and the modification of the social, cultural and natural sites of the songs have resulted in a loss of interest in the values and techniques of llano work. Llanero work songs thus face various threats to their viability. Efforts to safeguard the element are nonetheless widespread, including a pedagogical strategy involving more than twenty meetings for bearers and young people in the region, training projects for schoolteachers and a proliferation of festivals.\\[2mm]
  {\normalsize{UNESCO ICH URL:}} \href{https://ich.unesco.org/en/USL/colombian-venezuelan-llano-work-songs-01285}{https://ich.unesco.org/en/USL/colombian-venezuelan-llano-wor...}
\end{tcolorbox}
\end{figure}

%% file: examples/vvqa/subsaharian-african-states_0.tex
\begin{figure}[H]
\begin{tcolorbox}[colback=gray!5!white,colframe=black!75!black,fonttitle=\bfseries\scriptsize,fontupper=\ttfamily\footnotesize,segmentation style={solid, black!30}]
  \begin{center}
    \begin{minipage}{0.18\linewidth}
      \centering
      \includegraphics[width=\linewidth]{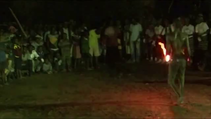}
    \end{minipage}\hfill
    \begin{minipage}{0.18\linewidth}
      \centering
      \includegraphics[width=\linewidth]{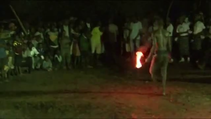}
    \end{minipage}\hfill
    \begin{minipage}{0.18\linewidth}
      \centering
      \includegraphics[width=\linewidth]{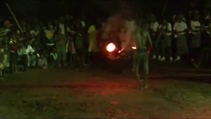}
    \end{minipage}\hfill
    \begin{minipage}{0.18\linewidth}
      \centering
      \includegraphics[width=\linewidth]{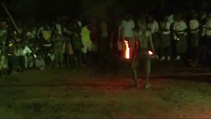}
    \end{minipage}\hfill
    \begin{minipage}{0.18\linewidth}
      \centering
      \includegraphics[width=\linewidth]{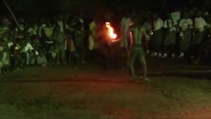}
    \end{minipage}\hfill
  \\[4mm]
    \begin{minipage}{0.18\linewidth}
      \centering
      \includegraphics[width=\linewidth]{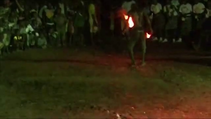}
    \end{minipage}\hfill
    \begin{minipage}{0.18\linewidth}
      \centering
      \includegraphics[width=\linewidth]{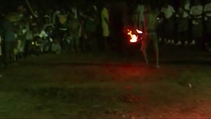}
    \end{minipage}\hfill
    \begin{minipage}{0.18\linewidth}
      \centering
      \includegraphics[width=\linewidth]{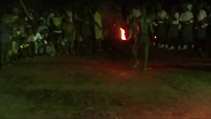}
    \end{minipage}\hfill
    \begin{minipage}{0.18\linewidth}
      \centering
      \includegraphics[width=\linewidth]{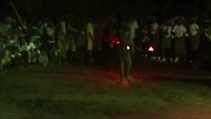}
    \end{minipage}\hfill
    \begin{minipage}{0.18\linewidth}
      \centering
      \includegraphics[width=\linewidth]{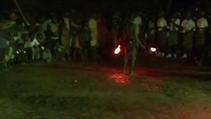}
    \end{minipage}\hfill
  \end{center}

  {\Large{Question:}} {\large{What type of theatre is depicted in the video, known for using elaborate costumes and performances?}}\\
  {\Large{Answer:}} {\large{Kwagh-Hir}}\\
   \tcbline
  {\Large{Related Cultural Event or Facet}}\\[4mm]
  {\normalsize{Title:}} {\normalsize{Kwagh-Hir theatrical performance}}\\
  {\normalsize{Countries:}} Nigeria\\
  {\normalsize{Regions:}} Subsaharian African States\\
  {\normalsize{Description:}}\\
  Kwagh-Hir theatrical performance is a composite art form encompassing a spectacle that is both visually stimulating and culturally edifying. Kwagh-hir has its roots in the story-telling tradition of the Tiv people called ‘kwagh-alom’, a practice where the family was treated to a storytelling session by creative storytellers, usually in the early hours of the night after the day’s farming work. With time, creative storytellers began to dramatize these stories, culminating in the present stage and status of Kwagh-hir. The practice is a social performance with the potential to entertain and teach moral lessons through the dramatization and performance of past and current social realities. As a form of total theatre, Kwagh-hir incorporates puppetry, masquerading, poetry, music, dance and animated narratives in articulating the reality of the Tiv people. People’s daily struggles, aspirations, successes and failures are all given expression through creative dramatization. Khwagh-hir theatre is owned by the community, with knowledge and skills being transmitted through apprenticeship. People who indicate an interest in the troupe’s activities are trained and mentored until they reach a certain level of proficiency; they are then accepted into the troupe. Regular performances are held to ensure the art is kept alive and that the younger generation continues to identify with it.\\[2mm]
  {\normalsize{UNESCO ICH URL:}} \href{https://ich.unesco.org/en/RL/kwagh-hir-theatrical-performance-00683}{https://ich.unesco.org/en/RL/kwagh-hir-theatrical-performanc...}
\end{tcolorbox}
\end{figure}

%% file: examples/vvqa/western-european-and-north-american-states_0.tex
\begin{figure}[H]
\begin{tcolorbox}[colback=gray!5!white,colframe=black!75!black,fonttitle=\bfseries\scriptsize,fontupper=\ttfamily\footnotesize,segmentation style={solid, black!30}]
  \begin{center}
    \begin{minipage}{0.18\linewidth}
      \centering
      \includegraphics[width=\linewidth]{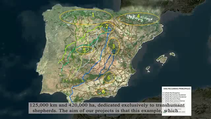}
    \end{minipage}\hfill
    \begin{minipage}{0.18\linewidth}
      \centering
      \includegraphics[width=\linewidth]{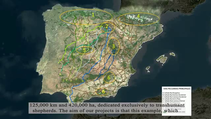}
    \end{minipage}\hfill
    \begin{minipage}{0.18\linewidth}
      \centering
      \includegraphics[width=\linewidth]{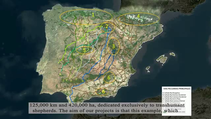}
    \end{minipage}\hfill
    \begin{minipage}{0.18\linewidth}
      \centering
      \includegraphics[width=\linewidth]{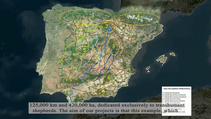}
    \end{minipage}\hfill
    \begin{minipage}{0.18\linewidth}
      \centering
      \includegraphics[width=\linewidth]{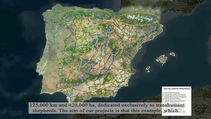}
    \end{minipage}\hfill
  \\[4mm]
    \begin{minipage}{0.18\linewidth}
      \centering
      \includegraphics[width=\linewidth]{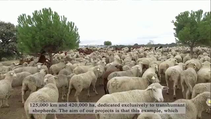}
    \end{minipage}\hfill
    \begin{minipage}{0.18\linewidth}
      \centering
      \includegraphics[width=\linewidth]{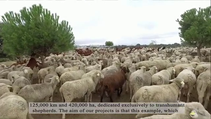}
    \end{minipage}\hfill
    \begin{minipage}{0.18\linewidth}
      \centering
      \includegraphics[width=\linewidth]{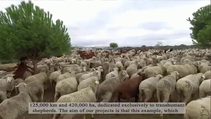}
    \end{minipage}\hfill
    \begin{minipage}{0.18\linewidth}
      \centering
      \includegraphics[width=\linewidth]{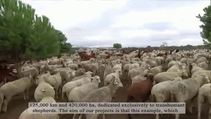}
    \end{minipage}\hfill
    \begin{minipage}{0.18\linewidth}
      \centering
      \includegraphics[width=\linewidth]{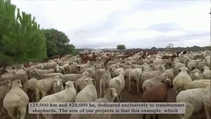}
    \end{minipage}\hfill
  \end{center}

  {\Large{Question:}} {\large{What traditional practice is depicted with the herders and sheep in the video?}}\\
  {\Large{Answer:}} {\large{Transhumance}}\\
   \tcbline
  {\Large{Related Cultural Event or Facet}}\\[4mm]
  {\normalsize{Title:}} {\normalsize{Transhumance, the seasonal droving of livestock}}\\
  {\normalsize{Countries:}} Albania, Andorra, Austria, Croatia, Spain, France, Greece, Italy, Luxembourg, Romania\\
  {\normalsize{Regions:}} Western European and North American States, Eastern European States\\
  {\normalsize{Description:}}\\
  Transhumance refers to the seasonal movement of people with their livestock between geographical or climatic regions. Each year, in spring and autumn, men and women herders organise the movement of thousands of animals along traditional pastoral paths. They move on foot or horseback, leading with their dogs and sometimes accompanied by their families. An ancestral practice, transhumance stems from a deep knowledge about the environment and entails social practices and rituals related to the care, breeding and training of animals and the management of natural resources. An entire socio-economic system has been developed around transhumance, from gastronomy to local handicrafts and festivities marking the beginning and end of a season. Families have been enacting and transmitting transhumance through observation and practice for many generations. Communities living along transhumance routes also play an important role in its transmission, such as by celebrating herd crossings and organising festivals. The practice is also transmitted through workshops organised by local communities, associations and networks of herders and farmers, as well as through universities and research institutes. Transhumance thus contributes to social inclusion, strengthening cultural identity and ties between families, communities and territories while counteracting the effects of rural depopulation.\\[2mm]
  {\normalsize{UNESCO ICH URL:}} \href{https://ich.unesco.org/en/RL/transhumance-the-seasonal-droving-of-livestock-01964}{https://ich.unesco.org/en/RL/transhumance-the-seasonal-drovi...}
\end{tcolorbox}
\end{figure}

%% file: src/991_3_0_appendix_coqa.tex
\section{\texttt{COQA} Details}
\label{appendix:sec:coqa}

\subsection{Prompts}
\label{appendix:sec:coqa:prompts}
In the following, the prompts for the \coqar and \coqac tasks are provided.
For the variations involving images, the image placeholder gets replaced $N$ times, where $N$ is the number of images related to the target CEF.
\begin{figure*}[ht]
    \centering
    \begin{promptbox}{Region --- Text-Only}
    \begin{minted}[breaklines]{markdown}
From which of the following regions does the cultural event or facet with the title `{TITLE}` originate?
Choose from the following options and output only the corresponding letter.

A. {REGION_OPTION_A}
B. {REGION_OPTION_B}
C. {REGION_OPTION_C}
D. {REGION_OPTION_D}

Your answer letter:
    \end{minted}
    \end{promptbox}
    \begin{promptbox}{Region --- Image-Only}
    \begin{minted}[breaklines]{markdown}
<IMAGE_PLACEHOLDER>

From which of the following countries does the cultural event or facet shown in the images originate?
Choose from the following options and output only the corresponding letter.

A. {REGION_OPTION_A}
B. {REGION_OPTION_B}
C. {REGION_OPTION_C}
D. {REGION_OPTION_D}

Your answer letter:
    \end{minted}
    \end{promptbox}
    \begin{promptbox}{Region --- Text-Image}
    \begin{minted}[breaklines]{markdown}
<IMAGE_PLACEHOLDER>

From which of the following regions does the cultural event or facet with the title `{TITLE}` shown in the images originate?
Choose from the following options and output only the corresponding letter.

A. {REGION_OPTION_A}
B. {REGION_OPTION_B}
C. {REGION_OPTION_C}
D. {REGION_OPTION_D}

Your answer letter:
    \end{minted}
    \end{promptbox}
    \label{fig:coqa:prompts_r}
    \caption{Prompts for the \coqar task.}
\end{figure*}

\begin{figure*}[ht]
    \centering
    \begin{promptbox}{Country --- Text-Only}
    \begin{minted}[breaklines]{markdown}
From which of the following countries does the cultural event or facet with the title `{TITLE}` originate?
Choose from the following options and output only the corresponding letter.

A. {COUNTRY_OPTION_A}
B. {COUNTRY_OPTION_B}
C. {COUNTRY_OPTION_C}
D. {COUNTRY_OPTION_D}

Your answer letter:
    \end{minted}
    \end{promptbox}
    \begin{promptbox}{Country --- Image-Only}
    \begin{minted}[breaklines]{markdown}
<IMAGE_PLACEHOLDER>

From which of the following countries does the cultural event or facet with the title `{TITLE}` originate?
Choose from the following options and output only the corresponding letter.

A. {COUNTRY_OPTION_A}
B. {COUNTRY_OPTION_B}
C. {COUNTRY_OPTION_C}
D. {COUNTRY_OPTION_D}

Your answer letter:
    \end{minted}
    \end{promptbox}
    \begin{promptbox}{Country --- Text-Image}
    \begin{minted}[breaklines]{markdown}
<IMAGE_PLACEHOLDER>

From which of the following countries does the cultural event or facet with the title `{TITLE}` shown in the images originate?
Choose from the following options and output only the corresponding letter.

A. {COUNTRY_OPTION_A}
B. {COUNTRY_OPTION_B}
C. {COUNTRY_OPTION_C}
D. {COUNTRY_OPTION_D}

Your answer letter:
    \end{minted}
    \end{promptbox}
    \label{fig:coqa:prompts_c}
    \caption{Prompts for the \coqac task.}
\end{figure*}

\clearpage
\subsection{Examples}
\label{appendix:sec:coqa:examples}
In the following, we provide one random sample per region for the \coqac task in the image-only setting.
For the other settings and the \coqa tasks, the same pattern applies using the respective prompts from above.
Note that the lower part of the examples, where the related CEF is provided, is \emph{not} part of the actual sample.

\subsubsection*{\RegA}
\input{examples/coqa/arab-states_0}
\subsubsection*{\RegAP}
\input{examples/coqa/asian-and-pacific-states_0}
\subsubsection*{\RegE}
\input{examples/coqa/eastern-european-states_0}
\subsubsection*{\RegLAC}
\input{examples/coqa/latin-american-and-caribbean-states_0}
\subsubsection*{\RegSA}
\input{examples/coqa/subsaharian-african-states_0}
\subsubsection*{\RegW}
\input{examples/coqa/western-european-and-north-american-states_0}

%% file: examples/coqa/arab-states_0.tex
\begin{figure}[H]
\begin{tcolorbox}[colback=gray!5!white,colframe=black!75!black,fonttitle=\bfseries\scriptsize,fontupper=\ttfamily\footnotesize,segmentation style={solid, black!30}]
  \begin{center}
    \begin{minipage}{0.18\linewidth}
      \centering
      \includegraphics[width=\linewidth]{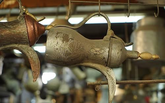}
      {\captionsetup{labelformat=empty}\captionof{figure}{\tiny\textit{Copyrigth: Huzaifa Ayad Bahaa El Din, Iraq, 2021}}}
    \end{minipage}\hfill
    \begin{minipage}{0.18\linewidth}
      \centering
      \includegraphics[width=\linewidth]{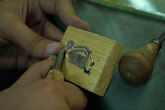}
      {\captionsetup{labelformat=empty}\captionof{figure}{\tiny\textit{Copyrigth: Huzaifa Ayad Bahaa El Din, Iraq, 2021}}}
    \end{minipage}\hfill
    \begin{minipage}{0.18\linewidth}
      \centering
      \includegraphics[width=\linewidth]{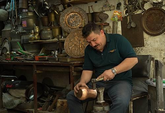}
      {\captionsetup{labelformat=empty}\captionof{figure}{\tiny\textit{Copyrigth: Huzaifa Ayad Bahaa El Din, Iraq, 2021}}}
    \end{minipage}\hfill
    \begin{minipage}{0.18\linewidth}
      \centering
      \includegraphics[width=\linewidth]{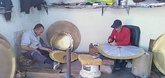}
      {\captionsetup{labelformat=empty}\captionof{figure}{\tiny\textit{Copyrigth: Zahia Benabdallah, Algeria, 2021}}}
    \end{minipage}\hfill
    \begin{minipage}{0.18\linewidth}
      \centering
      \includegraphics[width=\linewidth]{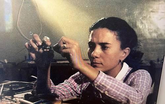}
      {\captionsetup{labelformat=empty}\captionof{figure}{\tiny\textit{Copyrigth: Azza Fahmi, Egypt, 2021}}}
    \end{minipage}\hfill
  \\[4mm]
    \begin{minipage}{0.18\linewidth}
      \centering
      \includegraphics[width=\linewidth]{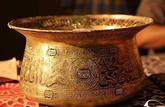}
      {\captionsetup{labelformat=empty}\captionof{figure}{\tiny\textit{Copyrigth: Mustafa Kamil, Egypt, 2021}}}
    \end{minipage}\hfill
    \begin{minipage}{0.18\linewidth}
      \centering
      \includegraphics[width=\linewidth]{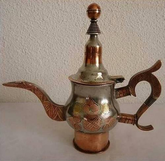}
      {\captionsetup{labelformat=empty}\captionof{figure}{\tiny\textit{Copyrigth: National Heritage Preservation, Ministry of Culture, Youth and Sport and Relations with the Parliament, Egypt, 2022}}}
    \end{minipage}\hfill
    \begin{minipage}{0.18\linewidth}
      \centering
      \includegraphics[width=\linewidth]{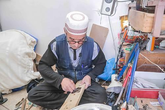}
      {\captionsetup{labelformat=empty}\captionof{figure}{\tiny\textit{Copyrigth: Direction du Patrimoine Culturel, Morocco, 2021}}}
    \end{minipage}\hfill
    \begin{minipage}{0.18\linewidth}
      \centering
      \includegraphics[width=\linewidth]{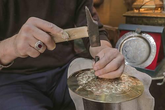}
      {\captionsetup{labelformat=empty}\captionof{figure}{\tiny\textit{Copyrigth: Direction du Patrimoine Culturel, Morocco, 2021}}}
    \end{minipage}\hfill
    \begin{minipage}{0.18\linewidth}
      \centering
      \includegraphics[width=\linewidth]{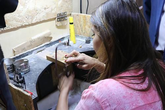}
      {\captionsetup{labelformat=empty}\captionof{figure}{\tiny\textit{Copyrigth: Ministry of Culture, Palestine, 2021}}}
    \end{minipage}\hfill
  \end{center}

  {\Large{Question:}} {\large{In which of the following countries does the event shown in the images take place? Choose from the following options and output only the corresponding letter.

A. Kuwait

B. Jordan

C. Egypt

D. United Arab Emirates

Your answer letter:}}\\
  {\Large{Answer:}} {\large{C}}\\
   \tcbline
  {\Large{Related Cultural Event or Facet}}\\[4mm]
  {\normalsize{Title:}} {\normalsize{Arts, skills and practices associated with engraving on metals (gold, silver and copper)}}\\
  {\normalsize{Countries:}} Algeria, Saudi Arabia, Egypt, Iraq, Morocco, Mauritania, Palestine, Sudan, Tunisia, Yemen\\
  {\normalsize{Regions:}} Arab States\\
  {\normalsize{Description:}}\\
  Engraving on metals such as gold, silver and copper is a centuries-old practice that entails manually cutting words, symbols or patterns into the surfaces of decorative, utilitarian, religious or ceremonial objects. The craftsperson uses different tools to manually cut symbols, names, Quran verses, prayers and geometric patterns into the objects. Engravings can be concave (recessed) or convex (elevated), or the result of a combination of different types of metals, such as gold and silver. Their social and symbolic meanings and functions vary according to the communities concerned. Engraved objects, such as jewelry or household objects, are often presented as traditional gifts for weddings or used in religious rituals and alternative medicine. For instance, certain types of metals are believed to have healing properties. Engraving on metals is transmitted within families, through observation and hands-on practice. It is also transmitted through workshops organized by training centres, organizations and universities, among others. Publications, cultural events and social media further contribute to the transmission of the related knowledge and skills. Practised by people of all ages and genders, metal engraving and the use of engraved objects are means of expressing the cultural, religious and geographical identity and the socioeconomic status of the communities concerned.\\[2mm]
  {\normalsize{UNESCO ICH URL:}} \href{https://ich.unesco.org/en/RL/arts-skills-and-practices-associated-with-engraving-on-metals-gold-silver-and-copper-01951}{https://ich.unesco.org/en/RL/arts-skills-and-practices-assoc...}
\end{tcolorbox}
\end{figure}

%% file: examples/coqa/asian-and-pacific-states_0.tex
\begin{figure}[H]
\begin{tcolorbox}[colback=gray!5!white,colframe=black!75!black,fonttitle=\bfseries\scriptsize,fontupper=\ttfamily\footnotesize,segmentation style={solid, black!30}]
  \begin{center}
    \begin{minipage}{0.18\linewidth}
      \centering
      \includegraphics[width=\linewidth]{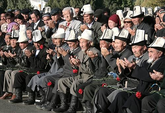}
      {\captionsetup{labelformat=empty}\captionof{figure}{\tiny\textit{Copyrigth: Public Foundation 'Min Kiyal', Kyrgyzstan, 2018}}}
    \end{minipage}\hfill
    \begin{minipage}{0.18\linewidth}
      \centering
      \includegraphics[width=\linewidth]{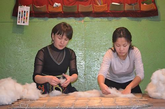}
      {\captionsetup{labelformat=empty}\captionof{figure}{\tiny\textit{Copyrigth: Public Foundation 'Min Kiyal', Kyrgyzstan, 2018}}}
    \end{minipage}\hfill
    \begin{minipage}{0.18\linewidth}
      \centering
      \includegraphics[width=\linewidth]{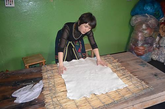}
      {\captionsetup{labelformat=empty}\captionof{figure}{\tiny\textit{Copyrigth: Public Foundation 'Min Kiyal', Kyrgyzstan, 2018}}}
    \end{minipage}\hfill
    \begin{minipage}{0.18\linewidth}
      \centering
      \includegraphics[width=\linewidth]{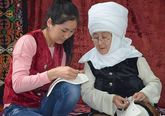}
      {\captionsetup{labelformat=empty}\captionof{figure}{\tiny\textit{Copyrigth: Public Foundation 'Min Kiyal', Kyrgyzstan, 2018}}}
    \end{minipage}\hfill
    \begin{minipage}{0.18\linewidth}
      \centering
      \includegraphics[width=\linewidth]{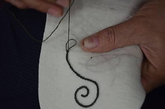}
      {\captionsetup{labelformat=empty}\captionof{figure}{\tiny\textit{Copyrigth: Public Foundation 'Min Kiyal', Kyrgyzstan, 2018}}}
    \end{minipage}\hfill
  \\[4mm]
    \begin{minipage}{0.18\linewidth}
      \centering
      \includegraphics[width=\linewidth]{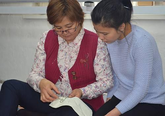}
      {\captionsetup{labelformat=empty}\captionof{figure}{\tiny\textit{Copyrigth: Public Foundation 'Min Kiyal', Kyrgyzstan, 2018}}}
    \end{minipage}\hfill
    \begin{minipage}{0.18\linewidth}
      \centering
      \includegraphics[width=\linewidth]{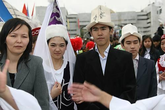}
      {\captionsetup{labelformat=empty}\captionof{figure}{\tiny\textit{Copyrigth: Public Foundation 'Min Kiyal', Kyrgyzstan, 2018}}}
    \end{minipage}\hfill
    \begin{minipage}{0.18\linewidth}
      \centering
      \includegraphics[width=\linewidth]{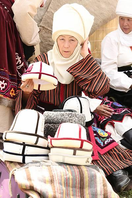}
      {\captionsetup{labelformat=empty}\captionof{figure}{\tiny\textit{Copyrigth: Public Foundation 'Min Kiyal', Kyrgyzstan, 2018}}}
    \end{minipage}\hfill
    \begin{minipage}{0.18\linewidth}
      \centering
      \includegraphics[width=\linewidth]{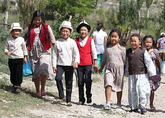}
      {\captionsetup{labelformat=empty}\captionof{figure}{\tiny\textit{Copyrigth: Public Foundation 'Min Kiyal', Kyrgyzstan, 2018}}}
    \end{minipage}\hfill
    \begin{minipage}{0.18\linewidth}
      \centering
      \includegraphics[width=\linewidth]{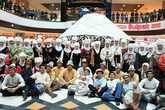}
      {\captionsetup{labelformat=empty}\captionof{figure}{\tiny\textit{Copyrigth: Public Foundation 'Min Kiyal', Kyrgyzstan, 2018}}}
    \end{minipage}\hfill
  \end{center}

  {\Large{Question:}} {\large{In which of the following countries does the event shown in the images take place? Choose from the following options and output only the corresponding letter.

A. Kyrgyzstan

B. Timor-Leste

C. Thailand

D. Turkmenistan

Your answer letter:}}\\
  {\Large{Answer:}} {\large{A}}\\
   \tcbline
  {\Large{Related Cultural Event or Facet}}\\[4mm]
  {\normalsize{Title:}} {\normalsize{Ak-kalpak craftsmanship, traditional knowledge and skills in making and wearing Kyrgyz men’s headwear}}\\
  {\normalsize{Countries:}} Kyrgyzstan\\
  {\normalsize{Regions:}} Asian and Pacific States\\
  {\normalsize{Description:}}\\
  Ak-kalpak craftsmanship is a traditional Kyrgyz handicraft. The Ak-kalpak is a traditional male hat made with white felt, which bears deep sacral meanings. Ak-kalpak craftsmanship is a cumulative, ever-evolving body of knowledge and skills passed down by craftswomen in the communities concerned comprising felting, cutting and sewing and pattern embroidery. Related knowledge and skills are transmitted via oral coaching, hands-on training and joint making in workshops. More than eighty kinds of Ak-kalpak can be distinguished, decorated with various patterns bearing a sacred meaning and history. Environmentally friendly and comfortable, the Ak-kalpak resembles a snow peak, with four sides representing the four elements: air, water, fire and earth. The four edging lines symbolize life, with the tassels on the top symbolizing ancestors’ posterity and memory, and the pattern symbolizing the family tree. Ak-kalpak unites different Kyrgyz tribes and communities and makes Kyrgyz people recognizable to other ethnic groups. It also fosters inclusivity when representatives of other ethnic groups wear it on holidays or days of mourning to express unity and sympathy. There are workshops all over the country where related knowledge and skills are passed down, and in 2013 a project entitled ‘From generation to generation’ was conducted on traditional Ak-kalpak-making techniques nationwide, resulting in an exhibition and published book.\\[2mm]
  {\normalsize{UNESCO ICH URL:}} \href{https://ich.unesco.org/en/RL/ak-kalpak-craftsmanship-traditional-knowledge-and-skills-in-making-and-wearing-kyrgyz-men-s-headwear-01496}{https://ich.unesco.org/en/RL/ak-kalpak-craftsmanship-traditi...}
\end{tcolorbox}
\end{figure}

%% file: examples/coqa/eastern-european-states_0.tex
\begin{figure}[H]
\begin{tcolorbox}[colback=gray!5!white,colframe=black!75!black,fonttitle=\bfseries\scriptsize,fontupper=\ttfamily\footnotesize,segmentation style={solid, black!30}]
  \begin{center}
    \begin{minipage}{0.18\linewidth}
      \centering
      \includegraphics[width=\linewidth]{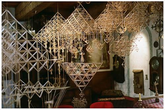}
      {\captionsetup{labelformat=empty}\captionof{figure}{\tiny\textit{Copyrigth: Lithuanian National Culture Centre, Archive, 2021}}}
    \end{minipage}\hfill
    \begin{minipage}{0.18\linewidth}
      \centering
      \includegraphics[width=\linewidth]{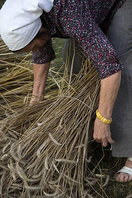}
      {\captionsetup{labelformat=empty}\captionof{figure}{\tiny\textit{Copyrigth: Vilnius Ethnic Culture Centre, Archive, 2021}}}
    \end{minipage}\hfill
    \begin{minipage}{0.18\linewidth}
      \centering
      \includegraphics[width=\linewidth]{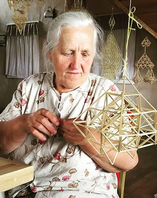}
      {\captionsetup{labelformat=empty}\captionof{figure}{\tiny\textit{Copyrigth: Vilnius Ethnic Culture Centre, Archive, 2021}}}
    \end{minipage}\hfill
    \begin{minipage}{0.18\linewidth}
      \centering
      \includegraphics[width=\linewidth]{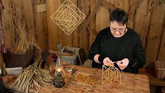}
      {\captionsetup{labelformat=empty}\captionof{figure}{\tiny\textit{Copyrigth: Lithuanian National Culture Centre, Archive, 2021}}}
    \end{minipage}\hfill
    \begin{minipage}{0.18\linewidth}
      \centering
      \includegraphics[width=\linewidth]{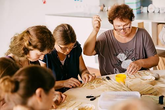}
      {\captionsetup{labelformat=empty}\captionof{figure}{\tiny\textit{Copyrigth: Vilnius Ethnic Culture Centre, Archive, 2021}}}
    \end{minipage}\hfill
  \\[4mm]
    \begin{minipage}{0.18\linewidth}
      \centering
      \includegraphics[width=\linewidth]{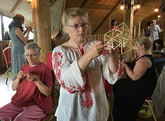}
      {\captionsetup{labelformat=empty}\captionof{figure}{\tiny\textit{Copyrigth: Vilnius Ethnic Culture Centre, Archive, 2021}}}
    \end{minipage}\hfill
    \begin{minipage}{0.18\linewidth}
      \centering
      \includegraphics[width=\linewidth]{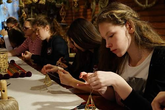}
      {\captionsetup{labelformat=empty}\captionof{figure}{\tiny\textit{Copyrigth: Vilnius Ethnic Culture Centre, Archive, 2021}}}
    \end{minipage}\hfill
    \begin{minipage}{0.18\linewidth}
      \centering
      \includegraphics[width=\linewidth]{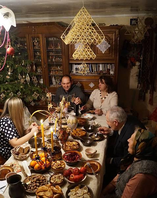}
      {\captionsetup{labelformat=empty}\captionof{figure}{\tiny\textit{Copyrigth: Lithuanian National Culture Centre, Archive, 2021}}}
    \end{minipage}\hfill
    \begin{minipage}{0.18\linewidth}
      \centering
      \includegraphics[width=\linewidth]{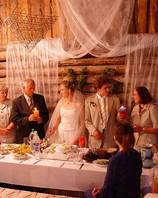}
      {\captionsetup{labelformat=empty}\captionof{figure}{\tiny\textit{Copyrigth: Marija Liugienė, Archive, 2003}}}
    \end{minipage}\hfill
    \begin{minipage}{0.18\linewidth}
      \centering
      \includegraphics[width=\linewidth]{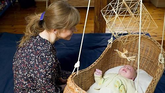}
      {\captionsetup{labelformat=empty}\captionof{figure}{\tiny\textit{Copyrigth: Lithuanian National Culture Centre, Archive, 2021}}}
    \end{minipage}\hfill
  \end{center}

  {\Large{Question:}} {\large{In which of the following countries does the event shown in the images take place? Choose from the following options and output only the corresponding letter.

A. Lithuania

B. Bosnia and Herzegovina

C. Russia

D. Poland

Your answer letter:}}\\
  {\Large{Answer:}} {\large{A}}\\
   \tcbline
  {\Large{Related Cultural Event or Facet}}\\[4mm]
  {\normalsize{Title:}} {\normalsize{Sodai straw garden making in Lithuania}}\\
  {\normalsize{Countries:}} Lithuania\\
  {\normalsize{Regions:}} Eastern European States\\
  {\normalsize{Description:}}\\
  Sodai straw gardens are hanging ornaments made from the stalks of grains. This practice involves the cultivation of grain (typically rye), the treatment of straw and the creation of geometric structures of varying sizes. The structures are then decorated with details symbolizing fertility and prosperity. Sodai gardens are believed to reflect the pattern of the universe and are associated with well-being and spirituality. They are hung over the cradles of babies and over a wedding or family table to wish happiness to newborns, fertility to newlyweds or harmony to the family. Lithuanian homes are also frequently decorated with sodai gardens for Easter and Christmas. Some sodai-making families have been practising the tradition for generations. Although most of the practitioners are women, workshops exist and are open to people of all ages and genders. The practice is passed on informally within families or during events such as festivals, exhibitions, conferences and summer camps. An integral part of traditional wooden home interiors, sodai gardens are viewed as spiritual gifts. They provide a sense of shared cultural heritage and continuity to the practising communities while strengthening communal partnerships, intergenerational bonds and cultural diversity.\\[2mm]
  {\normalsize{UNESCO ICH URL:}} \href{https://ich.unesco.org/en/RL/sodai-straw-garden-making-in-lithuania-01987}{https://ich.unesco.org/en/RL/sodai-straw-garden-making-in-li...}
\end{tcolorbox}
\end{figure}

%% file: examples/coqa/latin-american-and-caribbean-states_0.tex
\begin{figure}[H]
\begin{tcolorbox}[colback=gray!5!white,colframe=black!75!black,fonttitle=\bfseries\scriptsize,fontupper=\ttfamily\footnotesize,segmentation style={solid, black!30}]
  \begin{center}
    \begin{minipage}{0.18\linewidth}
      \centering
      \includegraphics[width=\linewidth]{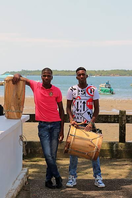}
      {\captionsetup{labelformat=empty}\captionof{figure}{\tiny\textit{Copyrigth: Gerson Fonseca/Ministry of Culture of Colombia, 2018}}}
    \end{minipage}\hfill
    \begin{minipage}{0.18\linewidth}
      \centering
      \includegraphics[width=\linewidth]{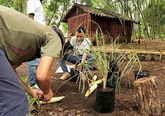}
      {\captionsetup{labelformat=empty}\captionof{figure}{\tiny\textit{Copyrigth: Gerson Fonseca/Ministry of Culture of Colombia, 2018}}}
    \end{minipage}\hfill
    \begin{minipage}{0.18\linewidth}
      \centering
      \includegraphics[width=\linewidth]{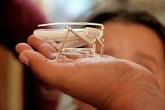}
      {\captionsetup{labelformat=empty}\captionof{figure}{\tiny\textit{Copyrigth: Gerson Fonseca/Ministry of Culture of Colombia, 2018}}}
    \end{minipage}\hfill
    \begin{minipage}{0.18\linewidth}
      \centering
      \includegraphics[width=\linewidth]{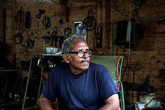}
      {\captionsetup{labelformat=empty}\captionof{figure}{\tiny\textit{Copyrigth: Gerson Fonseca/Ministry of Culture of Colombia, 2018}}}
    \end{minipage}\hfill
    \begin{minipage}{0.18\linewidth}
      \centering
      \includegraphics[width=\linewidth]{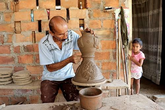}
      {\captionsetup{labelformat=empty}\captionof{figure}{\tiny\textit{Copyrigth: Gerson Fonseca/Ministry of Culture of Colombia, 2018}}}
    \end{minipage}\hfill
  \\[4mm]
    \begin{minipage}{0.18\linewidth}
      \centering
      \includegraphics[width=\linewidth]{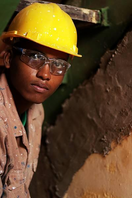}
      {\captionsetup{labelformat=empty}\captionof{figure}{\tiny\textit{Copyrigth: Gerson Fonseca/Ministry of Culture of Colombia, 2018}}}
    \end{minipage}\hfill
    \begin{minipage}{0.18\linewidth}
      \centering
      \includegraphics[width=\linewidth]{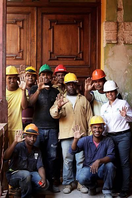}
      {\captionsetup{labelformat=empty}\captionof{figure}{\tiny\textit{Copyrigth: Gerson Fonseca/Ministry of Culture of Colombia, 2018}}}
    \end{minipage}\hfill
    \begin{minipage}{0.18\linewidth}
      \centering
      \includegraphics[width=\linewidth]{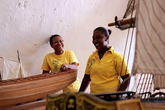}
      {\captionsetup{labelformat=empty}\captionof{figure}{\tiny\textit{Copyrigth: Gerson Fonseca/Ministry of Culture of Colombia, 2018}}}
    \end{minipage}\hfill
    \begin{minipage}{0.18\linewidth}
      \centering
      \includegraphics[width=\linewidth]{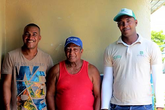}
      {\captionsetup{labelformat=empty}\captionof{figure}{\tiny\textit{Copyrigth: Gerson Fonseca/Ministry of Culture of Colombia, 2018}}}
    \end{minipage}\hfill
    \begin{minipage}{0.18\linewidth}
      \centering
      \includegraphics[width=\linewidth]{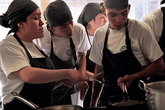}
      {\captionsetup{labelformat=empty}\captionof{figure}{\tiny\textit{Copyrigth: Gerson Fonseca/Ministry of Culture of Colombia, 2018}}}
    \end{minipage}\hfill
  \end{center}

  {\Large{Question:}} {\large{In which of the following countries does the event shown in the images take place? Choose from the following options and output only the corresponding letter.

A. Dominican Republic

B. Chile

C. Colombia

D. Grenada

Your answer letter:}}\\
  {\Large{Answer:}} {\large{C}}\\
   \tcbline
  {\Large{Related Cultural Event or Facet}}\\[4mm]
  {\normalsize{Title:}} {\normalsize{Safeguarding strategy of traditional crafts for peace building}}\\
  {\normalsize{Countries:}} Colombia\\
  {\normalsize{Regions:}} Latin-American and Caribbean States\\
  {\normalsize{Description:}}\\
  The safeguarding strategy of traditional crafts for peace building addresses the weakening of traditional crafts through a system of intergenerational transmission of knowledge between master and apprentice based on the non-formal ‘learning by doing’ method. The safeguarding strategy aims to train different sectors of the population, create labour connections and foster cultural entrepreneurship. It establishes a link between bearers of traditional crafts and skills who are recognized by their communities for their empirical knowledge of the peculiarities of their region and apprentices aged between fourteen and thirty-five who become builders of peace by learning a skill or craft, seeking to transform their situation of vulnerability. The safeguarding strategy is therefore geared at: allowing for the qualification of traditional crafts, thereby improving employment opportunities; implementing a Traditional Crafts Policy to guide and ensure continuity in the transmission and practice of these crafts; and enhancing the Workshop Schools Programme. Priority is accorded to young people who are exposed to the effects of armed conflict, a lack of opportunities, school desertion and unemployment. Training is also combined with work, guaranteeing apprentices’ future employability. The strategy thus aims to foster the safeguarding of traditional crafts as a tool for social inclusion, employment and cultural entrepreneurship. In turn, the community can recognize the cultural and societal value of safeguarding different traditional skills and crafts.\\[2mm]
  {\normalsize{UNESCO ICH URL:}} \href{https://ich.unesco.org/en/BSP/safeguarding-strategy-of-traditional-crafts-for-peace-building-01480}{https://ich.unesco.org/en/BSP/safeguarding-strategy-of-tradi...}
\end{tcolorbox}
\end{figure}

%% file: examples/coqa/subsaharian-african-states_0.tex
\begin{figure}[H]
\begin{tcolorbox}[colback=gray!5!white,colframe=black!75!black,fonttitle=\bfseries\scriptsize,fontupper=\ttfamily\footnotesize,segmentation style={solid, black!30}]
  \begin{center}
    \begin{minipage}{0.18\linewidth}
      \centering
      \includegraphics[width=\linewidth]{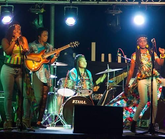}
      {\captionsetup{labelformat=empty}\captionof{figure}{\tiny\textit{Copyrigth: Etienne Kokolo, Kinshasa, République du Congo, 2018}}}
    \end{minipage}\hfill
    \begin{minipage}{0.18\linewidth}
      \centering
      \includegraphics[width=\linewidth]{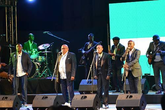}
      {\captionsetup{labelformat=empty}\captionof{figure}{\tiny\textit{Copyrigth: Etienne Kokolo, Kinshasa, République du Congo, 2019}}}
    \end{minipage}\hfill
    \begin{minipage}{0.18\linewidth}
      \centering
      \includegraphics[width=\linewidth]{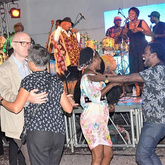}
      {\captionsetup{labelformat=empty}\captionof{figure}{\tiny\textit{Copyrigth: Etienne Kokolo, Kinshasa, République du Congo, 2018}}}
    \end{minipage}\hfill
    \begin{minipage}{0.18\linewidth}
      \centering
      \includegraphics[width=\linewidth]{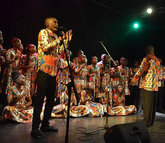}
      {\captionsetup{labelformat=empty}\captionof{figure}{\tiny\textit{Copyrigth: Etienne Kokolo, Kinshasa, République du Congo, 2018}}}
    \end{minipage}\hfill
    \begin{minipage}{0.18\linewidth}
      \centering
      \includegraphics[width=\linewidth]{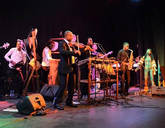}
      {\captionsetup{labelformat=empty}\captionof{figure}{\tiny\textit{Copyrigth: Etienne Kokolo, Kinshasa, République du Congo, 2018}}}
    \end{minipage}\hfill
  \\[4mm]
    \begin{minipage}{0.18\linewidth}
      \centering
      \includegraphics[width=\linewidth]{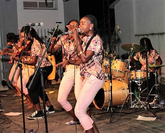}
      {\captionsetup{labelformat=empty}\captionof{figure}{\tiny\textit{Copyrigth: Etienne Kokolo, Kinshasa, République du Congo, 2017}}}
    \end{minipage}\hfill
    \begin{minipage}{0.18\linewidth}
      \centering
      \includegraphics[width=\linewidth]{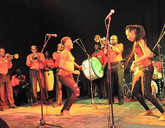}
      {\captionsetup{labelformat=empty}\captionof{figure}{\tiny\textit{Copyrigth: Etienne Kokolo, Kinshasa, République du Congo, 2018}}}
    \end{minipage}\hfill
    \begin{minipage}{0.18\linewidth}
      \centering
      \includegraphics[width=\linewidth]{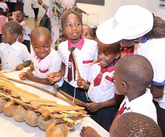}
      {\captionsetup{labelformat=empty}\captionof{figure}{\tiny\textit{Copyrigth: Etienne Kokolo, Kinshasa, République du Congo, 2020}}}
    \end{minipage}\hfill
    \begin{minipage}{0.18\linewidth}
      \centering
      \includegraphics[width=\linewidth]{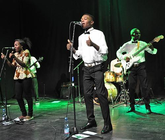}
      {\captionsetup{labelformat=empty}\captionof{figure}{\tiny\textit{Copyrigth: Etienne Kokolo, Kinshasa, République du Congo, 2017}}}
    \end{minipage}\hfill
    \begin{minipage}{0.18\linewidth}
      \centering
      \includegraphics[width=\linewidth]{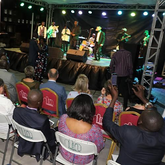}
      {\captionsetup{labelformat=empty}\captionof{figure}{\tiny\textit{Copyrigth: Etienne Kokolo, Kinshasa, République du Congo, 2020}}}
    \end{minipage}\hfill
  \end{center}

  {\Large{Question:}} {\large{In which of the following countries does the event shown in the images take place? Choose from the following options and output only the corresponding letter.

A. Congo

B. Togo

C. Namibia

D. Nigeria

Your answer letter:}}\\
  {\Large{Answer:}} {\large{A}}\\
   \tcbline
  {\Large{Related Cultural Event or Facet}}\\[4mm]
  {\normalsize{Title:}} {\normalsize{Congolese rumba}}\\
  {\normalsize{Countries:}} Congo, Democratic Republic of the Congo\\
  {\normalsize{Regions:}} Subsaharian African States\\
  {\normalsize{Description:}}\\
  Congolese rumba is a musical genre and a dance common in urban areas of the Democratic Republic of the Congo and the Republic of the Congo. Generally danced by a male-female couple, it is a multicultural form of expression originating from an ancient dance called nkumba (meaning ‘waist’ in Kikongo). The rumba is used for celebration and mourning, in private, public and religious spaces. It is performed by professional and amateur orchestras, choirs, dancers and individual musicians, and women have played a predominant role in the development of religious and romantic styles. The tradition of Congolese rumba is passed down to younger generations through neighbourhood clubs, formal training schools and community organisations. For instance, rumba musicians maintain clubs and apprentice artists to carry on the practice and the manufacture of instruments. The rumba also plays an important economic role, as orchestras are increasingly developing cultural entrepreneurship aimed at reducing poverty. The rumba is considered an essential and representative part of the identity of Congolese people and its diaspora. It is perceived as a means of conveying the social and cultural values of the region and of promoting intergenerational and social cohesion and solidarity.\\[2mm]
  {\normalsize{UNESCO ICH URL:}} \href{https://ich.unesco.org/en/RL/congolese-rumba-01711}{https://ich.unesco.org/en/RL/congolese-rumba-01711...}
\end{tcolorbox}
\end{figure}

%% file: examples/coqa/western-european-and-north-american-states_0.tex
\begin{figure}[H]
\begin{tcolorbox}[colback=gray!5!white,colframe=black!75!black,fonttitle=\bfseries\scriptsize,fontupper=\ttfamily\footnotesize,segmentation style={solid, black!30}]
  \begin{center}
    \begin{minipage}{0.18\linewidth}
      \centering
      \includegraphics[width=\linewidth]{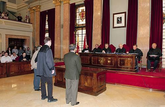}
      {\captionsetup{labelformat=empty}\captionof{figure}{\tiny\textit{Copyrigth: Servicio de Patrimonio Histórico de la Región de Murcia, 2005}}}
    \end{minipage}\hfill
    \begin{minipage}{0.18\linewidth}
      \centering
      \includegraphics[width=\linewidth]{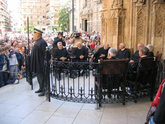}
      {\captionsetup{labelformat=empty}\captionof{figure}{\tiny\textit{Copyrigth: Generalitat Valenciana, 2005}}}
    \end{minipage}\hfill
    \begin{minipage}{0.18\linewidth}
      \centering
      \includegraphics[width=\linewidth]{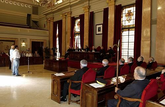}
      {\captionsetup{labelformat=empty}\captionof{figure}{\tiny\textit{Copyrigth: Servicio de Patrimonio Histórico de la Región de Murcia, 2005}}}
    \end{minipage}\hfill
    \begin{minipage}{0.18\linewidth}
      \centering
      \includegraphics[width=\linewidth]{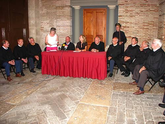}
      {\captionsetup{labelformat=empty}\captionof{figure}{\tiny\textit{Copyrigth: Generalitat Valenciana, 2005}}}
    \end{minipage}\hfill
    \begin{minipage}{0.18\linewidth}
      \centering
      \includegraphics[width=\linewidth]{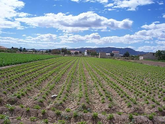}
      {\captionsetup{labelformat=empty}\captionof{figure}{\tiny\textit{Copyrigth: Servicio de Patrimonio Histórico de la Región de Murcia, 2005}}}
    \end{minipage}\hfill
  \\[4mm]
    \begin{minipage}{0.18\linewidth}
      \centering
      \includegraphics[width=\linewidth]{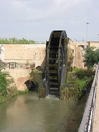}
      {\captionsetup{labelformat=empty}\captionof{figure}{\tiny\textit{Copyrigth: Servicio de Patrimonio Histórico de la Región de Murcia, 2005}}}
    \end{minipage}\hfill
    \begin{minipage}{0.18\linewidth}
      \centering
      \includegraphics[width=\linewidth]{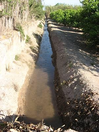}
      {\captionsetup{labelformat=empty}\captionof{figure}{\tiny\textit{Copyrigth: Servicio de Patrimonio Histórico de la Región de Murcia, 2005}}}
    \end{minipage}\hfill
    \begin{minipage}{0.18\linewidth}
      \centering
      \includegraphics[width=\linewidth]{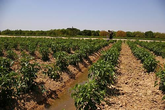}
      {\captionsetup{labelformat=empty}\captionof{figure}{\tiny\textit{Copyrigth: Servicio de Patrimonio Histórico de la Región de Murcia, 2005}}}
    \end{minipage}\hfill
    \begin{minipage}{0.18\linewidth}
      \centering
      \includegraphics[width=\linewidth]{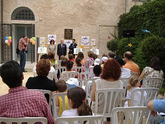}
      {\captionsetup{labelformat=empty}\captionof{figure}{\tiny\textit{Copyrigth: Servicio de Patrimonio Histórico de la Región de Murcia, 2005}}}
    \end{minipage}\hfill
    \begin{minipage}{0.18\linewidth}
      \centering
      \includegraphics[width=\linewidth]{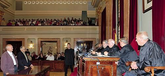}
      {\captionsetup{labelformat=empty}\captionof{figure}{\tiny\textit{Copyrigth: Servicio de Patrimonio Histórico de la Región de Murcia, 2005}}}
    \end{minipage}\hfill
  \end{center}

  {\Large{Question:}} {\large{In which of the following countries does the event shown in the images take place? Choose from the following options and output only the corresponding letter.

A. Austria

B. Spain

C. Cyprus

D. United Kingdom of Great Britain and Northern Ireland

Your answer letter:}}\\
  {\Large{Answer:}} {\large{B}}\\
   \tcbline
  {\Large{Related Cultural Event or Facet}}\\[4mm]
  {\normalsize{Title:}} {\normalsize{Irrigators’ tribunals of the Spanish Mediterranean coast: the Council of Wise Men of the plain of Murcia and the Water Tribunal of the plain of Valencia}}\\
  {\normalsize{Countries:}} Spain\\
  {\normalsize{Regions:}} Western European and North American States\\
  {\normalsize{Description:}}\\
  The irrigators’ tribunals of the Spanish Mediterranean coast are traditional law courts for water management that date back to the al-Andalus period (ninth to thirteenth centuries). The two main tribunals – the Council of Wise Men of the Plain of Murcia and the Water Tribunal of the Plain of Valencia – are recognized under Spanish law. Inspiring authority and respect, these two courts, whose members are elected democratically, settle disputes orally in a swift, transparent and impartial manner. The Council of Wise Men has seven geographically representative members, and has jurisdiction over a landowners’ assembly of 23,313 members. The Water Tribunal comprises eight elected administrators representing a total of 11,691 members from nine communities. In addition to their legal role the irrigators’ tribunals play a key part in the communities of which they are a visible symbol, as apparent from the rites performed when judgments are handed down and the fact that the tribunals often feature in local iconography. They provide cohesion among traditional communities and synergy between occupations (wardens, inspectors, pruners, etc.), contribute to the oral transmission of knowledge derived from centuries-old cultural exchanges, and have their own specialist vocabulary peppered with Arabic borrowings. In short, the courts are long-standing repositories of local and regional identity and are of special significance to local inhabitants.\\[2mm]
  {\normalsize{UNESCO ICH URL:}} \href{https://ich.unesco.org/en/RL/irrigators-tribunals-of-the-spanish-mediterranean-coast-the-council-of-wise-men-of-the-plain-of-murcia-and-the-water-tribunal-of-the-plain-of-valencia-00171}{https://ich.unesco.org/en/RL/irrigators-tribunals-of-the-spa...}
\end{tcolorbox}
\end{figure}

%% file: src/991_4_0_appendix_ckqa.tex
\section{\texttt{CKQA} Details}
\label{appendix:sec:ckqa}

\subsection{Prompts}
\label{appendix:sec:ckqa:prompts}
In the following, the prompts for the \ckqan and \ckqad tasks are provided.
For the variations involving images, the image placeholder gets replaced $N$ times, where $N$ is the number of images related to the target CEF.
Examples without the respective prompts, i.e., only the related CEFs, are provided in \S\ref{appendix:sec:benchmark:cef:examples}.

\begin{figure*}[ht!]
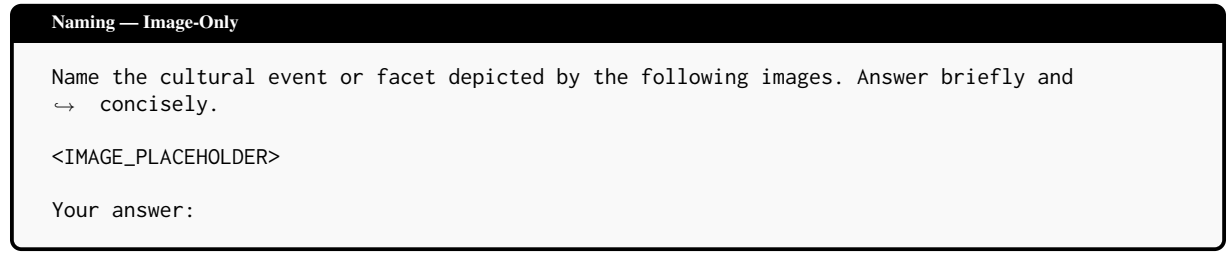

    \centering
    \begin{promptbox}{Naming --- Image-Only}
    \begin{minted}[breaklines]{markdown}
Name the cultural event or facet depicted by the following images. Answer briefly and concisely.

<IMAGE_PLACEHOLDER>

Your answer: 
    \end{minted}
    \end{promptbox}
    \label{fig:ckqa:prompts_n}
    \caption{Prompt for the \ckqan task.}
\end{figure*}

\begin{figure*}[ht!]
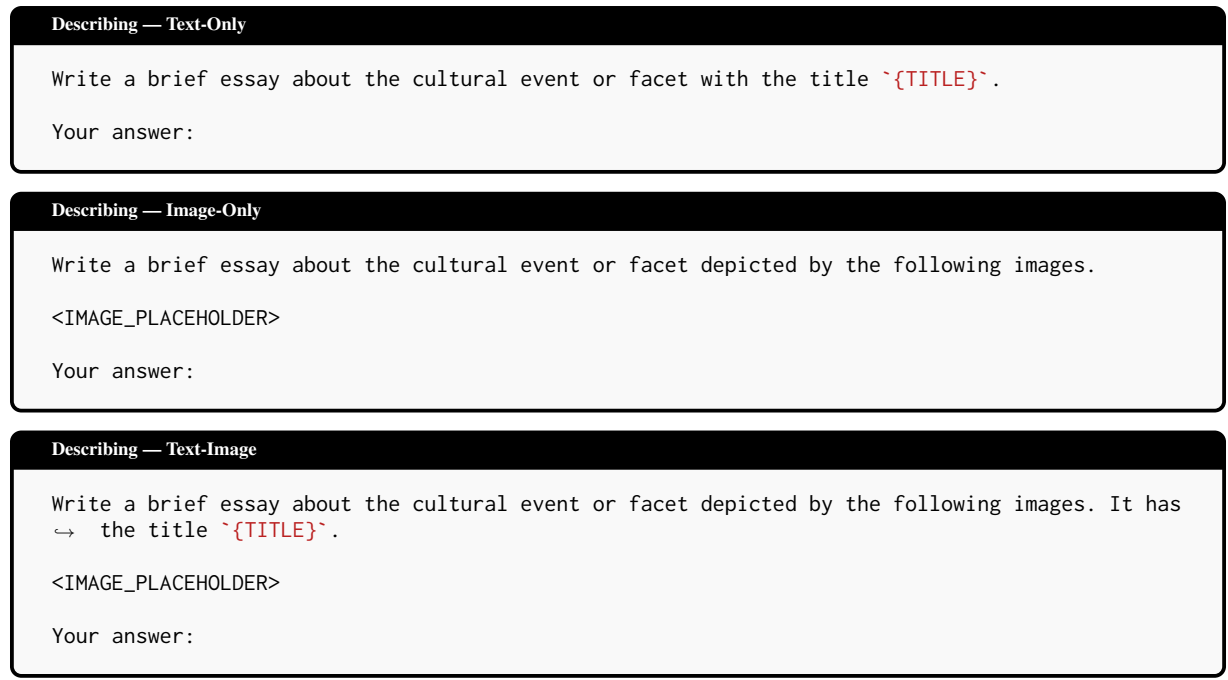

    \centering
    \begin{promptbox}{Describing --- Text-Only}
    \begin{minted}[breaklines]{markdown}
Write a brief essay about the cultural event or facet with the title `{TITLE}`.

Your answer: 
    \end{minted}
    \end{promptbox}
    \begin{promptbox}{Describing --- Image-Only}
    \begin{minted}[breaklines]{markdown}
Write a brief essay about the cultural event or facet depicted by the following images.

<IMAGE_PLACEHOLDER>

Your answer: 
    \end{minted}
    \end{promptbox}
    \begin{promptbox}{Describing --- Text-Image}
    \begin{minted}[breaklines]{markdown}
Write a brief essay about the cultural event or facet depicted by the following images. It has the title `{TITLE}`.

<IMAGE_PLACEHOLDER>

Your answer: 
    \end{minted}
    \end{promptbox}
    \label{fig:ckqa:prompts_c}
    \caption{Prompts for the \ckqad task.}
\end{figure*}

%% file: src/992_0_appendix_results.tex
\onecolumn
\section{Experimental Setup}
\label{appendix:sec:setup}
For inference, we load all models using the \textit{transformers} library (\texttt{v.4.48.0}) in 16-bit with Flash Attention 2~\cite{dao2022flashattention,dao2023flashattention2} (\texttt{v.2.7.3}), PyTorch (\texttt{v.2.4.0}), and CUDA (\texttt{v12.1}).
We used A40 (46GB) GPUs for models up to 26B parameters, A100 (80GB) GPUs for models up to 38B parameters, and two H100 (96GB) GPUs for 70B+ models in a multi-GPU setup.
To generate responses, we use greedy decoding, i.e., we use the following arguments for the generation method:
\begin{listing}[H]
\begin{minted}[fontsize=\footnotesize]{python}
generation_kwargs = {
    "max_new_tokens": 512,
    "do_sample": False,
    "temperature": None,
    "top_p": None,
    "top_k": None,
}
\end{minted}
\end{listing}

More details and exact hyperparameters are documented in the code base: \href{https://github.com/floschne/gimmick}{https://github.com/floschne/gimmick}.

\section{Results and Analyses}
\label{appendix:sec:analyses}
\input{src/992_1_appendix_results_sivqa}
\input{src/992_2_appendix_results_vvqa}
\input{src/992_3_appendix_results_coqa}
\input{src/992_4_appendix_results_ckqa}

%% file: src/992_1_appendix_results_sivqa.tex
\subsection{\sivqa}
\label{appendix:sec:analyses:sivqa}
\subsubsection{Results}
\label{appendix:sec:analyses:sivqa:results}
\subsubsection*{Relaxed Accuracy}
\label{appendix:sec:analyses:sivqa:results:acc}
\begin{table}[htbp]
  \centering
  \renewcommand{\arraystretch}{.97}
    \resizebox{\textwidth}{!}{%
  \begin{tabular}{l cccc cccc cccc cccc cccc cccc |cccc}
    \toprule
    \multirow{2}{*}{Model} 
      & \multicolumn{4}{c}{West EU \& North America} 
      & \multicolumn{4}{c}{Asia \& Pacific} 
      & \multicolumn{4}{c}{Subsaharian Africa} 
      & \multicolumn{4}{c}{Arab} 
      & \multicolumn{4}{c}{East EU} 
      & \multicolumn{4}{c}{Latin-America \& Caribbean} 
      & \multicolumn{4}{c}{Average} \\
    \cmidrule(lr){2-5} \cmidrule(lr){6-9} \cmidrule(lr){10-13} \cmidrule(lr){14-17} \cmidrule(lr){18-21} \cmidrule(lr){22-25} \cmidrule(lr){26-29}
      & N & R & C & B 
      & N & R & C & B 
      & N & R & C & B 
      & N & R & C & B 
      & N & R & C & B 
      & N & R & C & B 
      & N & R & C & B \\
\midrule
\rowcolor{gray!20}
GPT-4o & 31.58 & 34.39 & 41.05 & 40.70 & 29.89 & 31.37 & 36.63 & 37.68 & 17.38 & 17.63 & 32.49 & 31.74 & 25.70 & 30.53 & 39.19 & 39.95 & 26.80 & 32.56 & 42.94 & 41.79 & 23.17 & 25.77 & 30.26 & 32.39 & 25.44 & 28.17 & 36.59 & 37.08 \\
Gemini Pro & 27.02 & 30.53 & 31.23 & 32.28 & 22.53 & 26.11 & 31.16 & 29.68 & 16.84 & 14.61 & 26.20 & 24.18 & 19.85 & 22.39 & 28.50 & 28.50 & 25.07 & 25.94 & 31.12 & 32.56 & 22.46 & 19.86 & 24.35 & 27.19 & 21.50 & 22.84 & 28.30 & 28.30 \\
\rowcolor{gray!20}
GPT-4o Mini & 23.86 & 25.26 & 30.18 & 29.82 & 21.05 & 21.89 & 26.74 & 26.53 & 9.32 & 10.58 & 16.12 & 15.37 & 17.30 & 19.85 & 25.45 & 25.95 & 19.02 & 19.02 & 28.53 & 28.24 & 16.31 & 17.73 & 23.64 & 22.93 & 17.38 & 18.54 & 24.81 & 24.59 \\
Gemini Flash & 22.81 & 25.96 & 27.02 & 24.91 & 18.95 & 20.21 & 26.11 & 25.89 & 12.91 & 10.83 & 20.40 & 18.89 & 15.27 & 17.56 & 20.36 & 20.61 & 20.17 & 19.31 & 24.78 & 24.50 & 14.66 & 16.55 & 22.22 & 20.57 & 16.85 & 18.00 & 23.29 & 22.44 \\
\rowcolor{gray!20}
InternVL2.5 78B & 25.61 & 23.86 & 29.82 & 29.12 & 20.21 & 19.79 & 26.32 & 27.58 & 10.33 & 11.08 & 20.40 & 20.40 & 17.81 & 19.85 & 27.74 & 27.99 & 19.02 & 17.58 & 24.50 & 23.63 & 13.95 & 15.13 & 20.80 & 21.51 & 16.75 & 16.97 & 24.45 & 24.72 \\
Qwen2 VL 72B & 22.46 & 22.81 & 29.82 & 29.12 & 17.47 & 19.16 & 21.47 & 23.37 & 8.31 & 8.56 & 12.85 & 13.10 & 13.99 & 16.28 & 20.10 & 19.85 & 21.04 & 20.46 & 28.53 & 29.39 & 13.00 & 14.66 & 19.86 & 19.62 & 15.32 & 16.26 & 21.45 & 21.59 \\
\rowcolor{gray!20}
InternVL2.5 38B & 23.86 & 23.16 & 28.77 & 29.82 & 17.26 & 17.89 & 22.32 & 23.16 & 9.07 & 8.82 & 17.88 & 16.62 & 14.25 & 17.30 & 23.16 & 22.65 & 16.14 & 17.29 & 24.78 & 23.92 & 11.82 & 12.29 & 17.97 & 17.49 & 14.55 & 15.41 & 21.99 & 21.63 \\
Claude 3.5 Sonnet & 19.65 & 17.19 & 22.11 & 24.21 & 16.42 & 12.84 & 18.11 & 22.95 & 6.30 & 4.53 & 10.58 & 11.59 & 13.99 & 11.20 & 17.81 & 20.61 & 16.71 & 14.12 & 21.90 & 21.61 & 13.48 & 13.00 & 17.97 & 22.93 & 14.02 & 11.64 & 17.60 & 20.24 \\
\rowcolor{gray!20}
InternVL2.5 26B & 20.00 & 19.65 & 25.61 & 25.96 & 13.26 & 14.95 & 18.95 & 18.74 & 6.30 & 6.80 & 11.59 & 12.34 & 12.98 & 14.76 & 20.61 & 21.12 & 15.56 & 14.41 & 21.04 & 21.04 & 13.00 & 14.89 & 19.62 & 19.39 & 13.03 & 14.15 & 19.44 & 19.61 \\
Llama 3.2 11B Vision & 16.49 & 18.95 & 20.70 & 20.35 & 13.26 & 12.84 & 15.79 & 16.84 & 5.29 & 5.54 & 9.07 & 8.31 & 7.89 & 7.89 & 10.18 & 10.18 & 12.68 & 13.54 & 17.29 & 19.02 & 11.82 & 11.82 & 13.95 & 14.66 & 10.61 & 11.06 & 13.97 & 14.20 \\
\rowcolor{gray!20}
InternVL2.5 8B & 19.30 & 17.89 & 23.16 & 23.51 & 11.79 & 12.00 & 16.42 & 16.84 & 5.04 & 6.30 & 10.58 & 9.57 & 9.41 & 9.67 & 14.50 & 14.25 & 9.80 & 9.80 & 15.27 & 15.56 & 9.46 & 9.69 & 13.71 & 14.89 & 10.34 & 10.39 & 15.41 & 15.41 \\
Qwen2 VL 7B & 17.19 & 17.19 & 20.35 & 18.95 & 9.47 & 9.47 & 12.00 & 11.37 & 5.79 & 6.30 & 8.56 & 8.31 & 8.91 & 9.92 & 11.45 & 11.45 & 10.95 & 12.10 & 15.56 & 14.41 & 9.69 & 11.11 & 13.00 & 13.24 & 9.63 & 10.26 & 12.76 & 12.36 \\
\rowcolor{gray!20}
Phi 3.5 Vision & 14.39 & 12.63 & 20.00 & 18.95 & 8.84 & 10.74 & 13.89 & 13.47 & 6.05 & 6.05 & 8.31 & 8.82 & 6.62 & 8.14 & 9.41 & 9.92 & 8.93 & 8.65 & 14.70 & 14.70 & 8.27 & 9.93 & 13.71 & 12.77 & 8.55 & 9.18 & 12.99 & 12.85 \\
MiniCPM V 2.6 & 12.98 & 14.39 & 14.39 & 17.19 & 10.74 & 10.32 & 13.68 & 14.74 & 2.52 & 3.27 & 6.55 & 6.05 & 6.36 & 6.36 & 9.67 & 9.67 & 10.09 & 9.80 & 13.26 & 14.70 & 9.46 & 9.22 & 12.77 & 13.24 & 8.11 & 8.15 & 11.60 & 11.96 \\
\rowcolor{gray!20}
InternVL2.5 4B & 14.04 & 16.49 & 16.84 & 14.39 & 9.47 & 14.53 & 13.47 & 9.05 & 3.53 & 7.05 & 7.56 & 4.03 & 7.89 & 9.16 & 9.16 & 6.87 & 8.07 & 11.53 & 10.66 & 8.07 & 8.04 & 11.58 & 11.82 & 7.33 & 7.97 & 11.42 & 11.29 & 7.79 \\
Qwen2 VL 2B & 13.33 & 12.28 & 13.68 & 14.39 & 9.68 & 9.47 & 11.79 & 10.95 & 4.03 & 3.78 & 5.54 & 4.28 & 6.11 & 5.09 & 6.11 & 6.11 & 8.36 & 8.93 & 12.97 & 12.10 & 7.33 & 8.27 & 10.40 & 9.69 & 7.97 & 7.88 & 9.94 & 9.49 \\
\rowcolor{gray!20}
Centurio Qwen & 11.23 & 9.12 & 14.39 & 14.39 & 9.05 & 8.42 & 10.32 & 9.68 & 3.02 & 1.76 & 6.05 & 6.05 & 6.87 & 5.34 & 9.92 & 8.91 & 6.34 & 5.48 & 11.24 & 11.24 & 6.62 & 5.67 & 9.22 & 9.46 & 6.81 & 5.69 & 9.85 & 9.76 \\
InternVL2.5 2B & 6.67 & 7.37 & 10.18 & 9.47 & 4.21 & 4.63 & 6.95 & 5.89 & 2.27 & 2.02 & 3.53 & 5.29 & 2.80 & 3.56 & 5.34 & 5.60 & 3.17 & 3.75 & 7.49 & 6.92 & 5.44 & 5.44 & 8.04 & 6.15 & 4.03 & 4.39 & 6.85 & 6.45 \\
\rowcolor{gray!20}
InternVL2.5 1B & 7.02 & 7.37 & 10.53 & 11.58 & 4.21 & 3.58 & 4.84 & 4.63 & 2.52 & 0.76 & 2.77 & 2.77 & 3.56 & 3.82 & 5.09 & 4.07 & 4.61 & 3.46 & 6.63 & 7.49 & 4.02 & 5.67 & 6.86 & 6.15 & 4.03 & 4.03 & 5.96 & 5.87 \\
Centurio Aya & 3.16 & 7.37 & 8.77 & 8.77 & 2.95 & 5.68 & 9.05 & 9.68 & 1.76 & 1.51 & 4.79 & 3.53 & 1.27 & 3.56 & 5.60 & 6.11 & 2.02 & 3.46 & 7.20 & 7.20 & 2.84 & 5.44 & 6.38 & 7.09 & 2.24 & 4.39 & 6.99 & 7.17 \\
\midrule
Average X-Large & 24.04 & 23.33 & 29.82 & 29.12 & 18.84 & 19.47 & 23.89 & 25.47 & 9.32 & 9.82 & 16.62 & 16.75 & 15.90 & 18.07 & 23.92 & 23.92 & 20.03 & 19.02 & 26.51 & 26.51 & 13.48 & 14.89 & 20.33 & 20.57 & 16.03 & 16.61 & 22.95 & 23.15 \\
\rowcolor{gray!20}
Average Large & 21.93 & 21.40 & 27.19 & 27.89 & 15.26 & 16.42 & 20.63 & 20.95 & 7.68 & 7.81 & 14.74 & 14.48 & 13.61 & 16.03 & 21.88 & 21.88 & 15.85 & 15.85 & 22.91 & 22.48 & 12.41 & 13.59 & 18.79 & 18.44 & 13.79 & 14.78 & 20.71 & 20.62 \\
Average Medium & 13.39 & 14.15 & 16.96 & 17.19 & 9.54 & 9.79 & 12.88 & 13.19 & 3.90 & 4.11 & 7.60 & 6.97 & 6.79 & 7.12 & 10.22 & 10.09 & 8.65 & 9.03 & 13.30 & 13.69 & 8.31 & 8.83 & 11.51 & 12.10 & 7.96 & 8.32 & 11.76 & 11.81 \\
\rowcolor{gray!20}
Average Small & 11.09 & 11.23 & 14.25 & 13.75 & 7.28 & 8.59 & 10.19 & 8.80 & 3.68 & 3.93 & 5.54 & 5.04 & 5.39 & 5.95 & 7.02 & 6.51 & 6.63 & 7.26 & 10.49 & 9.86 & 6.62 & 8.18 & 10.17 & 8.42 & 6.51 & 7.38 & 9.40 & 8.49 \\
Average Open & 15.18 & 15.37 & 19.13 & 19.06 & 10.79 & 11.56 & 14.48 & 14.40 & 5.05 & 5.31 & 9.07 & 8.63 & 8.45 & 9.38 & 12.54 & 12.32 & 10.45 & 10.68 & 15.41 & 15.29 & 8.98 & 10.06 & 13.21 & 12.84 & 9.33 & 9.97 & 13.66 & 13.39 \\
\rowcolor{gray!20}
Average Proprietary & 24.98 & 26.67 & 30.32 & 30.39 & 21.77 & 22.48 & 27.75 & 28.55 & 12.55 & 11.64 & 21.16 & 20.35 & 18.42 & 20.31 & 26.26 & 27.12 & 21.56 & 22.19 & 29.86 & 29.74 & 18.01 & 18.58 & 23.69 & 25.20 & 19.04 & 19.84 & 26.12 & 26.53 \\
Average & 17.63 & 18.19 & 21.93 & 21.89 & 13.54 & 14.29 & 17.80 & 17.94 & 6.93 & 6.89 & 12.09 & 11.56 & 10.94 & 12.11 & 15.97 & 16.02 & 13.23 & 13.56 & 19.02 & 18.90 & 11.24 & 12.19 & 15.83 & 15.93 & 11.76 & 12.44 & 16.78 & 16.67 \\
\bottomrule
  \end{tabular}
  }%
  \caption{Cultural Image Visual Question Answering (\sivqa) scores. The reported score is relaxed accuracy. The columns \textbf{N}, \textbf{R}, \textbf{C}, and \textbf{B} stand for the hints \textbf{``None''}, \textbf{``Region''}, \textbf{``Country''}, and \textbf{``Both''}, respectively.}
  \label{tab:sivqa:scores}
\end{table}
\subsubsection*{Judge Score}
\label{appendix:sec:analyses:sivqa:results:judge}
\begin{figure}[ht!]
    \centering
    \includegraphics[width=.5\linewidth]{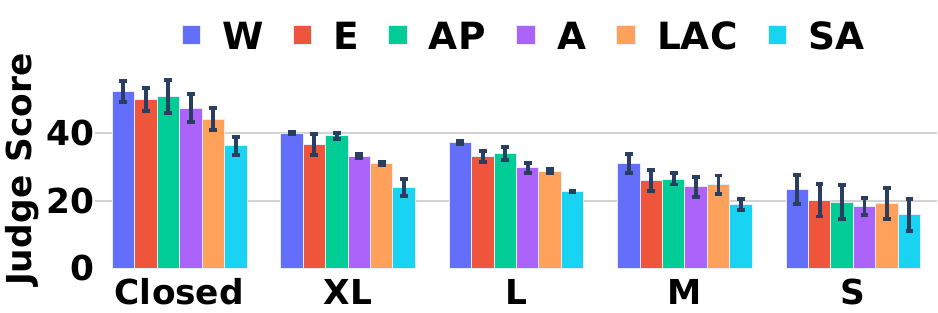}
    \caption{An overview of aggregated \sivqa Judge Score results.}
    \label{fig:analyses:a1_bias:scores:sivqa_judge}
\end{figure}
\begin{table}[htbp]
  \centering
  \renewcommand{\arraystretch}{.97}
    \resizebox{\textwidth}{!}{%
  \begin{tabular}{l cccc cccc cccc cccc cccc cccc |cccc}
    \toprule
    \multirow{2}{*}{Model} 
      & \multicolumn{4}{c}{West EU \& North America} 
      & \multicolumn{4}{c}{Asia \& Pacific} 
      & \multicolumn{4}{c}{Subsaharian Africa} 
      & \multicolumn{4}{c}{Arab} 
      & \multicolumn{4}{c}{East EU} 
      & \multicolumn{4}{c}{Latin-America \& Caribbean} 
      & \multicolumn{4}{c}{Average} \\
    \cmidrule(lr){2-5} \cmidrule(lr){6-9} \cmidrule(lr){10-13} \cmidrule(lr){14-17} \cmidrule(lr){18-21} \cmidrule(lr){22-25} \cmidrule(lr){26-29}
      & N & R & C & B 
      & N & R & C & B 
      & N & R & C & B 
      & N & R & C & B 
      & N & R & C & B 
      & N & R & C & B 
      & N & R & C & B \\
\midrule
\rowcolor{gray!20}
GPT-4o & 55.86 & 55.46 & 64.33 & 64.49 & 56.91 & 56.42 & 63.09 & 63.20 & 39.13 & 35.97 & 48.90 & 47.93 & 51.89 & 56.53 & 65.71 & 67.59 & 54.26 & 55.62 & 67.56 & 67.33 & 48.32 & 46.69 & 53.05 & 54.09 & 51.06 & 51.12 & 60.44 & 60.77 \\
Gemini Pro & 55.44 & 56.30 & 63.15 & 62.06 & 54.23 & 54.15 & 59.84 & 59.14 & 38.74 & 39.51 & 47.63 & 46.74 & 49.83 & 52.96 & 58.10 & 57.65 & 51.67 & 52.23 & 62.48 & 63.77 & 46.54 & 45.32 & 51.95 & 52.03 & 49.41 & 50.08 & 57.19 & 56.90 \\
\rowcolor{gray!20}
Claude 3.5 Sonnet & 50.65 & 51.35 & 65.33 & 64.14 & 50.11 & 51.07 & 63.70 & 63.59 & 35.06 & 39.90 & 57.56 & 56.52 & 48.99 & 55.34 & 67.49 & 67.86 & 50.42 & 51.80 & 70.12 & 70.20 & 41.49 & 46.10 & 57.85 & 59.53 & 46.12 & 49.26 & 63.68 & 63.64 \\
Gemini Flash & 50.49 & 49.93 & 56.29 & 56.11 & 48.19 & 47.42 & 53.20 & 53.11 & 35.60 & 31.32 & 39.58 & 38.54 & 44.04 & 44.25 & 50.85 & 49.03 & 46.66 & 47.34 & 57.24 & 55.22 & 43.59 & 41.58 & 47.74 & 46.88 & 44.76 & 43.64 & 50.82 & 49.81 \\
\rowcolor{gray!20}
GPT-4o Mini & 48.98 & 48.47 & 53.10 & 54.28 & 44.83 & 44.24 & 49.96 & 49.05 & 32.78 & 35.06 & 35.49 & 35.98 & 42.24 & 43.83 & 47.98 & 48.58 & 46.57 & 43.40 & 55.06 & 53.76 & 40.74 & 38.24 & 44.21 & 44.19 & 42.69 & 42.21 & 47.63 & 47.64 \\
Qwen2 VL 72B & 40.28 & 40.05 & 48.04 & 46.90 & 38.65 & 39.25 & 43.02 & 44.15 & 25.79 & 26.30 & 31.52 & 30.57 & 32.77 & 35.27 & 42.89 & 41.64 & 38.95 & 39.44 & 50.55 & 49.74 & 30.64 & 33.06 & 39.66 & 40.27 & 34.51 & 35.56 & 42.61 & 42.21 \\
\rowcolor{gray!20}
InternVL2.5 78B & 39.88 & 39.52 & 46.43 & 47.01 & 39.93 & 38.01 & 47.78 & 49.26 & 22.18 & 20.79 & 30.15 & 30.42 & 33.72 & 35.80 & 46.82 & 47.86 & 34.57 & 32.63 & 41.89 & 40.73 & 31.42 & 30.40 & 39.09 & 38.84 & 33.62 & 32.86 & 42.03 & 42.35 \\
InternVL2.5 26B & 37.00 & 34.65 & 39.75 & 41.00 & 32.64 & 32.97 & 39.10 & 39.47 & 22.63 & 21.71 & 29.40 & 27.22 & 30.89 & 31.39 & 38.10 & 38.81 & 34.34 & 32.38 & 41.14 & 41.53 & 29.34 & 29.69 & 37.05 & 37.78 & 31.14 & 30.47 & 37.42 & 37.64 \\
\rowcolor{gray!20}
InternVL2.5 38B & 37.55 & 37.51 & 45.58 & 45.49 & 35.45 & 36.26 & 42.65 & 43.88 & 22.98 & 22.52 & 29.11 & 28.35 & 28.71 & 31.78 & 38.96 & 38.63 & 32.08 & 31.69 & 41.98 & 41.21 & 28.39 & 29.15 & 36.46 & 35.18 & 30.86 & 31.48 & 39.12 & 38.79 \\
Qwen2 VL 7B & 33.36 & 34.84 & 38.64 & 38.12 & 28.19 & 28.97 & 31.23 & 31.13 & 21.31 & 25.25 & 25.09 & 26.26 & 28.72 & 28.45 & 32.00 & 32.28 & 29.19 & 31.13 & 35.53 & 37.11 & 27.84 & 28.61 & 31.45 & 32.87 & 28.10 & 29.54 & 32.33 & 32.96 \\
\rowcolor{gray!20}
Llama 3.2 11B Vision & 35.16 & 36.56 & 37.22 & 37.81 & 27.06 & 27.59 & 31.14 & 33.09 & 19.24 & 17.97 & 24.38 & 26.42 & 25.09 & 26.53 & 31.43 & 30.47 & 28.34 & 27.88 & 33.96 & 36.73 & 26.89 & 28.82 & 32.14 & 32.88 & 26.96 & 27.56 & 31.71 & 32.90 \\
MiniCPM V 2.6 & 30.61 & 32.35 & 32.73 & 35.48 & 27.73 & 25.88 & 31.13 & 33.25 & 20.29 & 18.92 & 25.31 & 24.58 & 24.52 & 24.57 & 28.47 & 28.19 & 28.13 & 25.07 & 34.31 & 36.27 & 26.74 & 26.04 & 29.16 & 30.37 & 26.34 & 25.47 & 30.18 & 31.36 \\
\rowcolor{gray!20}
Qwen2 VL 2B & 28.86 & 28.30 & 28.94 & 30.85 & 25.18 & 24.06 & 26.32 & 26.31 & 21.02 & 19.32 & 23.06 & 21.94 & 20.92 & 20.32 & 22.98 & 23.30 & 25.10 & 26.01 & 32.90 & 31.34 & 23.91 & 24.07 & 26.73 & 25.85 & 24.16 & 23.68 & 26.82 & 26.60 \\
Centurio Aya & 29.84 & 30.21 & 30.67 & 32.31 & 26.64 & 25.51 & 28.70 & 28.81 & 18.81 & 17.87 & 21.23 & 20.97 & 19.75 & 20.43 & 24.02 & 24.01 & 25.42 & 24.93 & 28.79 & 30.72 & 23.58 & 24.65 & 25.66 & 26.68 & 24.01 & 23.93 & 26.51 & 27.25 \\
\rowcolor{gray!20}
InternVL2.5 8B & 30.12 & 32.19 & 35.35 & 36.47 & 23.62 & 23.93 & 29.75 & 29.92 & 16.70 & 17.20 & 20.54 & 21.81 & 23.61 & 23.46 & 30.65 & 29.92 & 24.94 & 24.67 & 32.73 & 33.66 & 22.80 & 22.13 & 26.78 & 27.99 & 23.63 & 23.93 & 29.30 & 29.96 \\
Phi 3.5 Vision & 24.84 & 26.93 & 33.43 & 33.45 & 23.46 & 25.36 & 29.18 & 29.02 & 21.28 & 21.65 & 23.63 & 25.92 & 21.06 & 23.26 & 26.18 & 26.48 & 24.70 & 24.88 & 31.32 & 31.82 & 24.47 & 25.73 & 29.67 & 30.56 & 23.30 & 24.64 & 28.90 & 29.54 \\
\rowcolor{gray!20}
Centurio Qwen & 27.32 & 26.32 & 29.21 & 30.42 & 25.84 & 26.54 & 26.88 & 28.63 & 18.12 & 17.91 & 20.14 & 22.69 & 23.46 & 22.32 & 27.19 & 27.72 & 20.84 & 20.56 & 26.53 & 28.83 & 21.21 & 21.74 & 23.19 & 23.82 & 22.80 & 22.56 & 25.52 & 27.02 \\
InternVL2.5 4B & 25.18 & 26.06 & 26.71 & 29.67 & 20.67 & 22.04 & 26.29 & 27.53 & 12.32 & 14.45 & 14.99 & 17.91 & 18.42 & 21.62 & 24.43 & 25.80 & 20.22 & 22.07 & 26.43 & 28.43 & 17.56 & 20.93 & 23.45 & 24.29 & 19.06 & 21.19 & 23.72 & 25.61 \\
\rowcolor{gray!20}
InternVL2.5 1B & 19.67 & 20.32 & 20.90 & 23.91 & 14.46 & 13.92 & 14.95 & 17.03 & 12.05 & 13.50 & 16.60 & 15.86 & 16.48 & 16.31 & 16.88 & 17.42 & 16.10 & 14.90 & 18.27 & 20.82 & 14.94 & 15.64 & 16.75 & 17.59 & 15.62 & 15.76 & 17.39 & 18.77 \\
InternVL2.5 2B & 18.19 & 19.35 & 20.25 & 21.95 & 14.75 & 15.99 & 16.42 & 18.36 & 13.14 & 10.52 & 12.88 & 14.96 & 15.55 & 14.08 & 16.69 & 18.03 & 14.77 & 13.73 & 17.43 & 18.32 & 15.57 & 15.86 & 18.04 & 18.77 & 15.33 & 14.92 & 16.95 & 18.40 \\
\midrule
Average X-Large & 40.08 & 39.79 & 47.23 & 46.95 & 39.29 & 38.63 & 45.40 & 46.71 & 23.99 & 23.55 & 30.83 & 30.49 & 33.25 & 35.54 & 44.86 & 44.75 & 36.76 & 36.03 & 46.22 & 45.24 & 31.03 & 31.73 & 39.37 & 39.56 & 34.06 & 34.21 & 42.32 & 42.28 \\
\rowcolor{gray!20}
Average Large & 37.28 & 36.08 & 42.67 & 43.25 & 34.05 & 34.61 & 40.88 & 41.68 & 22.81 & 22.12 & 29.25 & 27.79 & 29.80 & 31.58 & 38.53 & 38.72 & 33.21 & 32.04 & 41.56 & 41.37 & 28.86 & 29.42 & 36.75 & 36.48 & 31.00 & 30.98 & 38.27 & 38.22 \\
Average Medium & 31.07 & 32.08 & 33.97 & 35.10 & 26.51 & 26.40 & 29.80 & 30.80 & 19.08 & 19.19 & 22.78 & 23.79 & 24.19 & 24.29 & 28.96 & 28.77 & 26.14 & 25.71 & 31.98 & 33.89 & 24.84 & 25.33 & 28.06 & 29.10 & 25.31 & 25.50 & 29.26 & 30.24 \\
\rowcolor{gray!20}
Average Small & 23.35 & 24.19 & 26.05 & 27.96 & 19.70 & 20.27 & 22.63 & 23.65 & 15.96 & 15.89 & 18.23 & 19.32 & 18.48 & 19.12 & 21.43 & 22.21 & 20.18 & 20.32 & 25.27 & 26.15 & 19.29 & 20.45 & 22.93 & 23.41 & 19.49 & 20.04 & 22.76 & 23.78 \\
Average Open & 30.52 & 31.01 & 34.26 & 35.39 & 26.95 & 27.08 & 30.97 & 31.99 & 19.19 & 19.06 & 23.20 & 23.73 & 24.24 & 25.04 & 29.85 & 30.04 & 26.51 & 26.13 & 32.92 & 33.82 & 24.35 & 25.10 & 29.02 & 29.58 & 25.30 & 25.57 & 30.03 & 30.76 \\
\rowcolor{gray!20}
Average Proprietary & 52.28 & 52.30 & 60.44 & 60.22 & 50.85 & 50.66 & 57.96 & 57.62 & 36.26 & 36.35 & 45.83 & 45.14 & 47.40 & 50.58 & 58.03 & 58.14 & 49.92 & 50.08 & 62.49 & 62.05 & 44.14 & 43.59 & 50.96 & 51.34 & 46.81 & 47.26 & 55.95 & 55.75 \\
Average & 35.96 & 36.33 & 40.80 & 41.59 & 32.93 & 32.98 & 37.72 & 38.40 & 23.46 & 23.38 & 28.86 & 29.08 & 30.03 & 31.42 & 36.89 & 37.06 & 32.36 & 32.12 & 40.31 & 40.88 & 29.30 & 29.72 & 34.50 & 35.02 & 30.67 & 30.99 & 36.51 & 37.01 \\
\bottomrule
  \end{tabular}
  }%
  \caption{Cultural Image Visual Question Answering (\sivqa) scores. The reported score is the average judge score. The columns \textbf{N}, \textbf{R}, \textbf{C}, and \textbf{B} stand for the hints \textbf{``None''}, \textbf{``Region''}, \textbf{``Country''}, and \textbf{``Both''}, respectively.}
  \label{tab:sivqa:judge_scores}
\end{table}

\subsubsection{Ground-Truth Answer Perplexity}
\label{appendix:sec:analyses:sivqa:results:ppl}

The perplexity for every sample is computed as follows:

\begin{equation}
\mathrm{PPL}(y \mid x) = \exp\left(-\frac{1}{N} \sum_{t=0}^{N} \log p\left(y_t \mid y_{t-1},\, x\right)\right)
\end{equation}

where $x = \{s, v\}$ are the textual ($s$) and visual ($v$) prompt (prefix) tokens and $y$ are the $N$ ground-truth answer tokens.

\subsubsection*{Results Per Cultural Aspect}
\label{sec:appendix:sec:analyses:sivqa:results:aspects}
We computed the average accuracy for questions targeting one of the ten most frequent cultural aspects (see \S\ref{appedix:sec:sivqa:aspects}), grouped by model size and region.
For better interpretation, Table~\ref{tab:sivqa:results:aspect_counts} reports the counts of questions associated with each cultural aspect per region.
As shown in Table~\ref{tab:sivqa:results:aspect_scores}, our results reveal a consistent trend: models perform significantly better on tangible cultural aspects (e.g., food) than on intangible ones.
For instance, across all regions, closed models achieve an average accuracy of 30\% for food-related questions, compared to only 8\% and 10\% for questions concerning rituals and festivals, respectively.
These findings highlight not only regional biases but also biases along the cultural dimension, the latter being particularly pronounced in non-Western contexts.
\begin{table}[ht]
    \centering
    \resizebox{\textwidth}{!}{%
\begin{tabular}{lrrrrrrrrrr}
\toprule
aspect & art & craftsmanship & dance & festivals & food & instruments & music & rituals & tools & traditions \\
\midrule
\RegA & 45 & 32 & 20 & 6 & 33 & 20 & 37 & 32 & 30 & 76 \\
\RegAP & 57 & 44 & 31 & 14 & 12 & 25 & 32 & 53 & 22 & 68 \\
\RegE & 53 & 36 & 18 & 19 & 10 & 19 & 26 & 18 & 20 & 49 \\
\RegLAC & 31 & 22 & 31 & 66 & 6 & 13 & 51 & 47 & 12 & 78 \\
\RegSA & 14 & 16 & 40 & 16 & 22 & 64 & 41 & 73 & 7 & 70 \\
\RegW & 33 & 27 & 10 & 30 & 13 & 14 & 23 & 18 & 17 & 49 \\
\bottomrule
\end{tabular}
}
    \caption{Number of questions targeting one of the top-10 cultural aspects per region in \sivqa.}
    \label{tab:sivqa:results:aspect_counts}
\end{table}

\begin{table}[ht!]
    \centering
    \resizebox{\textwidth}{!}{%
\begin{tabular}{lrrrrrrrrrrrrrrrrrrrrrrrrrrrrrrrrrrr}
\toprule
 & \multicolumn{5}{c}{\RegAP} & \multicolumn{5}{c}{\RegA} & \multicolumn{5}{c}{\RegSA} & \multicolumn{5}{c}{\RegW} & \multicolumn{5}{c}{\RegE} & \multicolumn{5}{c}{\RegLAC} & \multicolumn{5}{c}{\textsc{Overall}} \\
\cmidrule(lr){2-6} \cmidrule(lr){7-11} \cmidrule(lr){12-16} \cmidrule(lr){17-21} \cmidrule(lr){22-26} \cmidrule(lr){27-31} \cmidrule(lr){32-36}
 & \textsc{A} & \textsc{XL} & \textsc{L} & \textsc{M} & \textsc{S} & \textsc{A} & \textsc{XL} & \textsc{L} & \textsc{M} & \textsc{S} & \textsc{A} & \textsc{XL} & \textsc{L} & \textsc{M} & \textsc{S} & \textsc{A} & \textsc{XL} & \textsc{L} & \textsc{M} & \textsc{S} & \textsc{A} & \textsc{XL} & \textsc{L} & \textsc{M} & \textsc{S} & \textsc{A} & \textsc{XL} & \textsc{L} & \textsc{M} & \textsc{S} & \textsc{A} & \textsc{XL} & \textsc{L} & \textsc{M} & \textsc{S} \\
\midrule
food & 0.28 & 0.23 & 0.21 & 0.07 & 0.10 & 0.28 & 0.35 & 0.31 & 0.06 & 0.09 & 0.21 & 0.16 & 0.07 & 0.03 & 0.04 & 0.18 & 0.12 & 0.19 & 0.07 & 0.09 & 0.18 & 0.36 & 0.31 & 0.10 & 0.12 & 0.68 & 0.54 & 0.71 & 0.31 & 0.39 & 0.30 & 0.29 & 0.30 & 0.11 & 0.14 \\
instruments & 0.29 & 0.27 & 0.25 & 0.05 & 0.07 & 0.20 & 0.15 & 0.12 & 0.03 & 0.04 & 0.16 & 0.15 & 0.16 & 0.03 & 0.04 & 0.26 & 0.37 & 0.32 & 0.05 & 0.11 & 0.32 & 0.31 & 0.26 & 0.04 & 0.05 & 0.45 & 0.44 & 0.31 & 0.15 & 0.20 & 0.28 & 0.28 & 0.24 & 0.06 & 0.08 \\
craftsmanship & 0.15 & 0.17 & 0.15 & 0.04 & 0.07 & 0.18 & 0.24 & 0.22 & 0.08 & 0.08 & 0.11 & 0.09 & 0.08 & 0.04 & 0.02 & 0.14 & 0.19 & 0.12 & 0.10 & 0.08 & 0.26 & 0.28 & 0.27 & 0.13 & 0.17 & 0.12 & 0.12 & 0.12 & 0.06 & 0.07 & 0.16 & 0.18 & 0.16 & 0.08 & 0.08 \\
music & 0.20 & 0.32 & 0.26 & 0.07 & 0.09 & 0.10 & 0.09 & 0.12 & 0.03 & 0.04 & 0.13 & 0.11 & 0.13 & 0.03 & 0.03 & 0.25 & 0.27 & 0.28 & 0.10 & 0.13 & 0.10 & 0.10 & 0.05 & 0.02 & 0.02 & 0.19 & 0.22 & 0.22 & 0.08 & 0.12 & 0.16 & 0.19 & 0.18 & 0.06 & 0.07 \\
tools & 0.19 & 0.29 & 0.18 & 0.09 & 0.11 & 0.18 & 0.17 & 0.18 & 0.05 & 0.06 & 0.00 & 0.05 & 0.04 & 0.00 & 0.04 & 0.22 & 0.15 & 0.19 & 0.09 & 0.11 & 0.14 & 0.17 & 0.17 & 0.04 & 0.04 & 0.23 & 0.17 & 0.30 & 0.05 & 0.05 & 0.16 & 0.17 & 0.18 & 0.05 & 0.07 \\
traditions & 0.19 & 0.18 & 0.15 & 0.06 & 0.07 & 0.11 & 0.09 & 0.09 & 0.04 & 0.05 & 0.06 & 0.07 & 0.06 & 0.04 & 0.04 & 0.21 & 0.25 & 0.21 & 0.09 & 0.09 & 0.16 & 0.16 & 0.15 & 0.06 & 0.08 & 0.14 & 0.16 & 0.13 & 0.06 & 0.08 & 0.14 & 0.15 & 0.13 & 0.06 & 0.07 \\
art & 0.15 & 0.17 & 0.13 & 0.07 & 0.07 & 0.18 & 0.13 & 0.14 & 0.04 & 0.04 & 0.07 & 0.09 & 0.06 & 0.01 & 0.00 & 0.16 & 0.27 & 0.23 & 0.10 & 0.13 & 0.13 & 0.20 & 0.12 & 0.06 & 0.07 & 0.10 & 0.11 & 0.11 & 0.06 & 0.09 & 0.13 & 0.16 & 0.13 & 0.06 & 0.07 \\
dance & 0.07 & 0.07 & 0.04 & 0.01 & 0.00 & 0.02 & 0.00 & 0.00 & 0.00 & 0.00 & 0.05 & 0.04 & 0.02 & 0.01 & 0.01 & 0.26 & 0.39 & 0.35 & 0.17 & 0.11 & 0.22 & 0.23 & 0.17 & 0.04 & 0.03 & 0.14 & 0.11 & 0.08 & 0.04 & 0.03 & 0.13 & 0.14 & 0.11 & 0.05 & 0.03 \\
festivals & 0.18 & 0.20 & 0.18 & 0.09 & 0.08 & 0.02 & 0.06 & 0.00 & 0.00 & 0.00 & 0.02 & 0.00 & 0.00 & 0.00 & 0.00 & 0.10 & 0.10 & 0.10 & 0.04 & 0.06 & 0.13 & 0.11 & 0.05 & 0.02 & 0.01 & 0.13 & 0.11 & 0.11 & 0.04 & 0.04 & 0.10 & 0.10 & 0.07 & 0.03 & 0.03 \\
rituals & 0.09 & 0.08 & 0.09 & 0.02 & 0.03 & 0.06 & 0.06 & 0.05 & 0.02 & 0.03 & 0.06 & 0.07 & 0.04 & 0.02 & 0.04 & 0.12 & 0.10 & 0.08 & 0.02 & 0.02 & 0.13 & 0.16 & 0.08 & 0.01 & 0.01 & 0.04 & 0.06 & 0.03 & 0.01 & 0.00 & 0.08 & 0.09 & 0.06 & 0.02 & 0.02 \\
\midrule
\emph{Average} & 0.18 & 0.20 & 0.16 & 0.06 & 0.07 & 0.13 & 0.13 & 0.12 & 0.03 & 0.04 & 0.09 & 0.08 & 0.07 & 0.02 & 0.03 & 0.19 & 0.22 & 0.21 & 0.08 & 0.09 & 0.18 & 0.21 & 0.16 & 0.05 & 0.06 & 0.22 & 0.20 & 0.21 & 0.09 & 0.11 & 0.16 & 0.17 & 0.16 & 0.06 & 0.07 \\
\bottomrule
\end{tabular}
    }
    \caption{The averaged accuracy per region per model size group (A, XL, L, M, S) per target cultural aspect for samples in the \sivqa task.}
    \label{tab:sivqa:results:aspect_scores}
\end{table}

%% file: src/992_2_appendix_results_vvqa.tex
\subsection{\vvqa}
\label{appendix:sec:analyses:vvqa}
\subsubsection{Results}
\begin{table}[htbp]
  \centering
  \renewcommand{\arraystretch}{.97}
    \resizebox{\textwidth}{!}{%
  \begin{tabular}{@{}l cccc cccc cccc cccc cccc cccc |cccc@{}}
    \toprule
    \multirow{2}{*}{Model} 
      & \multicolumn{4}{c}{West EU \& North America} 
      & \multicolumn{4}{c}{Asia \& Pacific} 
      & \multicolumn{4}{c}{Subsaharian Africa} 
      & \multicolumn{4}{c}{Arab} 
      & \multicolumn{4}{c}{East EU} 
      & \multicolumn{4}{c}{Latin-America \& Caribbean} 
      & \multicolumn{4}{c}{Average} \\
    \cmidrule(lr){2-5} \cmidrule(lr){6-9} \cmidrule(lr){10-13} \cmidrule(lr){14-17} \cmidrule(lr){18-21} \cmidrule(lr){22-25} \cmidrule(lr){26-29}
      & N & R & C & B 
      & N & R & C & B 
      & N & R & C & B 
      & N & R & C & B 
      & N & R & C & B 
      & N & R & C & B 
      & N & R & C & B \\
\midrule
\rowcolor{gray!20}
GPT-4o & 38.97 & 39.91 & 41.31 & 44.13 & 34.56 & 36.41 & 35.71 & 39.17 & 23.67 & 26.33 & 36.67 & 36.00 & 29.18 & 32.46 & 36.72 & 36.39 & 37.59 & 40.43 & 47.16 & 45.74 & 31.98 & 32.56 & 38.95 & 39.24 & 32.67 & 34.49 & 39.19 & 39.97 \\
GPT-4o Mini & 38.06 & 31.58 & 34.01 & 38.87 & 29.45 & 25.64 & 25.64 & 29.66 & 20.32 & 13.33 & 15.56 & 20.63 & 28.61 & 24.40 & 25.60 & 29.52 & 35.37 & 29.27 & 30.18 & 38.72 & 25.13 & 21.20 & 23.04 & 25.65 & 28.69 & 23.89 & 24.84 & 29.69 \\
\rowcolor{gray!20}
Gemini Pro & 33.80 & 37.09 & 40.85 & 39.91 & 30.41 & 31.34 & 34.10 & 34.79 & 20.07 & 22.33 & 28.67 & 28.67 & 26.56 & 28.85 & 32.13 & 32.13 & 32.27 & 33.33 & 36.52 & 36.88 & 28.78 & 30.52 & 33.43 & 32.85 & 28.32 & 29.91 & 33.67 & 33.78 \\
Gemini Flash & 29.55 & 29.96 & 30.36 & 34.82 & 22.67 & 24.36 & 26.69 & 26.69 & 12.06 & 12.06 & 15.87 & 19.05 & 20.18 & 20.78 & 21.39 & 23.49 & 26.52 & 27.74 & 32.01 & 31.71 & 23.30 & 24.61 & 26.18 & 27.49 & 21.64 & 22.59 & 24.89 & 26.29 \\
\rowcolor{gray!20}
Claude 3.5 Sonnet & 21.86 & 19.84 & 25.91 & 24.29 & 22.46 & 19.92 & 25.21 & 25.85 & 9.21 & 6.03 & 12.38 & 11.11 & 16.87 & 14.16 & 16.87 & 18.37 & 23.17 & 20.12 & 26.52 & 24.70 & 19.11 & 15.45 & 21.47 & 22.25 & 18.74 & 15.89 & 21.44 & 21.24 \\
Qwen2 VL 72B & 25.35 & 27.23 & 33.33 & 34.27 & 18.43 & 19.12 & 23.73 & 23.73 & 9.00 & 10.00 & 16.33 & 17.00 & 17.05 & 18.36 & 22.62 & 21.64 & 25.53 & 25.53 & 32.27 & 31.91 & 16.57 & 20.93 & 23.55 & 24.42 & 18.13 & 19.62 & 24.27 & 24.65 \\
\rowcolor{gray!20}
InternVL2.5 78B & 23.94 & 29.11 & 31.92 & 31.46 & 19.12 & 24.19 & 28.11 & 29.49 & 7.33 & 12.33 & 18.67 & 19.67 & 13.44 & 22.30 & 25.90 & 25.90 & 19.50 & 24.82 & 30.14 & 29.43 & 15.41 & 21.22 & 24.42 & 26.45 & 15.75 & 21.56 & 25.98 & 26.70 \\
InternVL2.5 38B & 22.07 & 28.64 & 32.86 & 32.86 & 18.66 & 24.19 & 27.19 & 26.73 & 6.33 & 13.00 & 21.67 & 21.33 & 13.77 & 22.95 & 24.59 & 27.54 & 19.86 & 26.24 & 30.85 & 30.50 & 13.37 & 20.93 & 26.16 & 24.71 & 14.98 & 21.78 & 26.31 & 26.37 \\
\rowcolor{gray!20}
Qwen2 VL 2B & 19.72 & 18.78 & 21.13 & 23.00 & 13.13 & 14.75 & 16.59 & 15.67 & 6.67 & 4.67 & 7.00 & 6.67 & 13.11 & 11.15 & 13.44 & 12.46 & 16.67 & 16.67 & 17.38 & 17.38 & 15.70 & 15.12 & 16.28 & 16.28 & 13.88 & 13.27 & 15.26 & 14.98 \\
Qwen2 VL 7B & 18.78 & 18.78 & 22.07 & 21.60 & 14.06 & 14.06 & 17.05 & 16.59 & 5.00 & 6.00 & 7.67 & 7.33 & 13.11 & 15.08 & 17.05 & 17.70 & 15.25 & 17.73 & 18.79 & 19.50 & 15.70 & 18.60 & 19.48 & 20.06 & 13.54 & 14.76 & 16.86 & 16.92 \\
\rowcolor{gray!20}
InternVL2.5 26B & 20.66 & 25.35 & 28.64 & 29.11 & 16.36 & 19.35 & 23.27 & 24.88 & 3.33 & 7.33 & 9.33 & 10.33 & 11.80 & 15.41 & 19.67 & 20.00 & 17.73 & 21.63 & 24.82 & 24.11 & 13.66 & 18.31 & 21.51 & 22.97 & 13.32 & 17.30 & 20.78 & 21.61 \\
MiniCPM V 2.6 & 16.90 & 19.25 & 18.31 & 19.72 & 14.75 & 16.82 & 17.28 & 18.43 & 5.67 & 10.00 & 11.33 & 11.00 & 12.13 & 13.44 & 14.75 & 14.75 & 19.15 & 20.21 & 22.34 & 21.99 & 15.41 & 17.44 & 16.28 & 19.19 & 13.16 & 15.37 & 16.14 & 17.03 \\
\rowcolor{gray!20}
Phi 3.5 Vision & 16.43 & 14.55 & 16.90 & 16.90 & 13.82 & 14.06 & 17.51 & 17.28 & 8.67 & 8.33 & 10.67 & 10.33 & 9.84 & 10.16 & 11.15 & 10.82 & 15.60 & 15.25 & 19.15 & 19.86 & 13.95 & 15.12 & 18.60 & 18.90 & 12.82 & 12.88 & 16.09 & 15.87 \\
Centurio Qwen & 20.19 & 17.84 & 23.00 & 21.13 & 15.67 & 15.44 & 18.43 & 17.74 & 6.00 & 6.33 & 7.33 & 7.33 & 9.51 & 10.82 & 10.16 & 10.49 & 14.89 & 15.96 & 22.34 & 20.92 & 11.63 & 11.92 & 15.70 & 14.53 & 12.38 & 12.55 & 15.70 & 15.15 \\
\rowcolor{gray!20}
InternVL2.5 8B & 14.55 & 19.25 & 20.66 & 23.00 & 11.98 & 15.44 & 18.43 & 18.43 & 3.33 & 6.33 & 9.33 & 9.00 & 9.84 & 13.77 & 15.08 & 16.07 & 15.25 & 17.73 & 23.05 & 23.05 & 10.17 & 12.21 & 16.28 & 16.57 & 10.61 & 13.82 & 16.92 & 17.36 \\
InternVL2.5 4B & 14.55 & 15.96 & 18.78 & 18.31 & 12.67 & 14.29 & 17.74 & 16.59 & 5.67 & 6.33 & 9.00 & 9.00 & 8.52 & 9.84 & 13.11 & 12.46 & 11.35 & 15.25 & 19.50 & 18.09 & 11.34 & 14.24 & 15.70 & 15.12 & 10.45 & 12.38 & 15.70 & 14.70 \\
\rowcolor{gray!20}
Centurio Aya & 11.74 & 12.21 & 15.49 & 12.21 & 9.68 & 9.91 & 12.21 & 11.06 & 4.67 & 4.67 & 6.67 & 5.33 & 6.89 & 7.54 & 7.54 & 7.54 & 9.93 & 9.57 & 12.77 & 10.64 & 7.56 & 9.01 & 10.17 & 9.59 & 8.46 & 8.96 & 10.95 & 9.62 \\
InternVL2.5 1B & 8.45 & 9.86 & 11.27 & 12.21 & 5.76 & 8.29 & 9.22 & 7.60 & 1.67 & 2.33 & 4.00 & 2.67 & 5.90 & 7.87 & 8.85 & 8.85 & 6.74 & 7.45 & 10.99 & 10.64 & 7.27 & 8.72 & 9.01 & 9.01 & 5.86 & 7.46 & 8.90 & 8.35 \\
\midrule
Average X-Large & 24.65 & 28.17 & 32.63 & 32.86 & 18.78 & 21.66 & 25.92 & 26.61 & 8.17 & 11.17 & 17.50 & 18.33 & 15.25 & 20.33 & 24.26 & 23.77 & 22.52 & 25.18 & 31.21 & 30.67 & 15.99 & 21.08 & 23.98 & 25.44 & 16.94 & 20.59 & 25.12 & 25.68 \\
\rowcolor{gray!20}
Average Large & 21.36 & 27.00 & 30.75 & 30.99 & 17.51 & 21.77 & 25.23 & 25.81 & 4.83 & 10.17 & 15.50 & 15.83 & 12.79 & 19.18 & 22.13 & 23.77 & 18.79 & 23.94 & 27.84 & 27.30 & 13.52 & 19.62 & 23.84 & 23.84 & 14.15 & 19.54 & 23.55 & 23.99 \\
Average Medium & 16.43 & 17.46 & 19.91 & 19.53 & 13.23 & 14.33 & 16.68 & 16.45 & 4.93 & 6.67 & 8.47 & 8.00 & 10.30 & 12.13 & 12.92 & 13.31 & 14.89 & 16.24 & 19.86 & 19.22 & 12.09 & 13.84 & 15.58 & 15.99 & 11.63 & 13.09 & 15.31 & 15.21 \\
\rowcolor{gray!20}
Average Small & 14.79 & 14.79 & 17.02 & 17.61 & 11.35 & 12.85 & 15.26 & 14.29 & 5.67 & 5.42 & 7.67 & 7.17 & 9.34 & 9.75 & 11.64 & 11.15 & 12.59 & 13.65 & 16.76 & 16.49 & 12.06 & 13.30 & 14.90 & 14.83 & 10.75 & 11.50 & 13.99 & 13.47 \\
Average Open & 17.95 & 19.75 & 22.64 & 22.75 & 14.16 & 16.15 & 18.98 & 18.79 & 5.64 & 7.51 & 10.69 & 10.54 & 11.15 & 13.75 & 15.69 & 15.86 & 15.96 & 18.00 & 21.88 & 21.39 & 12.90 & 15.68 & 17.93 & 18.29 & 12.57 & 14.75 & 17.68 & 17.64 \\
\rowcolor{gray!20}
Average Proprietary & 32.45 & 31.67 & 34.49 & 36.40 & 27.91 & 27.53 & 29.47 & 31.23 & 17.06 & 16.02 & 21.83 & 23.09 & 24.28 & 24.13 & 26.54 & 27.98 & 30.98 & 30.18 & 34.48 & 35.55 & 25.66 & 24.87 & 28.61 & 29.50 & 26.01 & 25.35 & 28.80 & 30.19 \\
Average & 21.98 & 23.07 & 25.93 & 26.54 & 17.98 & 19.31 & 21.90 & 22.24 & 8.81 & 9.88 & 13.79 & 14.03 & 14.80 & 16.63 & 18.70 & 19.23 & 20.13 & 21.39 & 25.38 & 25.32 & 16.45 & 18.23 & 20.90 & 21.40 & 16.30 & 17.69 & 20.77 & 21.13 \\
\bottomrule
  \end{tabular}
  }%
  \caption{\dsname Video Visual Question Answering (VVQA) results. The reported score is relaxed accuracy. The columns \textbf{N}, \textbf{R}, \textbf{C}, and \textbf{B} stand for the hints \textbf{``None''}, \textbf{``Region''}, \textbf{``Country''}, and \textbf{``Both''}, respectively.}
  \label{tab:vvqa:scores}
\end{table}

%% file: src/992_3_appendix_results_coqa.tex
\newpage
\subsection{\coqa Details}
\label{appendix:sec:analyses:coqa}
\subsubsection{Results}
\begin{table}[htbp]
  \centering
  \renewcommand{\arraystretch}{.97}
    \resizebox{\textwidth}{!}{%
    \begin{tabular}{lccc ccc ccc ccc ccc ccc | cccc}
    \toprule
     & \multicolumn{3}{c}{\textsc{West EU \& North Am.}} &
     \multicolumn{3}{c}{\textsc{East EU}} &
     \multicolumn{3}{c}{\textsc{Asia \& Pacific}} &
     \multicolumn{3}{c}{\textsc{Lat. Am. \& Carib.}} &
     \multicolumn{3}{c}{\textsc{Arab}} &
     \multicolumn{3}{c}{\textsc{Subs. Africa}} &
     \multicolumn{4}{c}{\textsc{Average}} \\
    \cmidrule(lr){2-4} \cmidrule(lr){5-7} \cmidrule(lr){8-10} \cmidrule(lr){11-13} \cmidrule(lr){14-16} \cmidrule(lr){17-19} \cmidrule(lr){20-23}
    & I & T & I+T & I & T & I+T & I & T & I+T & I & T & I+T & I & T & I+T & I & T & I+T & I & T & I+T & Avg. \\
\midrule
\rowcolor{gray!20}
GPT-4o & 82.50 & 83.75 & 85.00 & 85.89 & 90.18 & 88.34 & 94.37 & 96.54 & 97.40 & 93.68 & 92.63 & 92.63 & 88.00 & 88.00 & 92.00 & 91.30 & 91.30 & 94.20 & 89.29 & 90.40 & 91.60 & 90.43 \\
Claude 3.5 Sonnet & 72.50 & 83.75 & 81.25 & 76.69 & 85.89 & 82.21 & 87.88 & 95.67 & 95.67 & 83.16 & 89.47 & 87.37 & 84.00 & 90.67 & 90.67 & 82.61 & 89.86 & 88.41 & 81.14 & 89.22 & 87.60 & 85.98 \\
\rowcolor{gray!20}
InternVL2.5 78B & 77.50 & 80.00 & 86.88 & 83.44 & 82.21 & 88.96 & 94.37 & 94.81 & 96.97 & 88.42 & 92.63 & 92.63 & 88.00 & 90.67 & 92.00 & 92.75 & 91.30 & 92.75 & 87.41 & 88.60 & 91.70 & 89.24 \\
Qwen2.5 72B & -- & 81.25 & -- & -- & 84.05 & -- & -- & 96.10 & -- & -- & 89.47 & -- & -- & 86.67 & -- & -- & 89.86 & -- & -- & 87.90 & -- & 87.90 \\
\rowcolor{gray!20}
GPT-4o Mini & 76.25 & 82.50 & 86.25 & 84.66 & 82.21 & 84.66 & 94.37 & 95.67 & 96.54 & 87.37 & 90.53 & 90.53 & 85.33 & 86.67 & 86.67 & 91.30 & 89.86 & 92.75 & 86.55 & 87.90 & 89.57 & 88.01 \\
InternVL2.5 38B & 81.25 & 81.25 & 84.38 & 85.89 & 84.66 & 85.28 & 90.04 & 95.24 & 92.64 & 86.32 & 86.32 & 92.63 & 89.33 & 90.67 & 92.00 & 89.86 & 86.96 & 91.30 & 87.11 & 87.51 & 89.70 & 88.11 \\
\rowcolor{gray!20}
Qwen2 VL 72B & 79.38 & 80.62 & 81.25 & 88.34 & 84.66 & 88.34 & 90.48 & 94.81 & 96.97 & 86.32 & 88.42 & 92.63 & 86.67 & 88.00 & 89.33 & 91.30 & 85.51 & 91.30 & 87.08 & 87.00 & 89.97 & 88.02 \\
Gemini Flash & 82.50 & 78.75 & 78.13 & 85.28 & 80.37 & 84.66 & 87.01 & 91.34 & 94.81 & 85.11 & 87.37 & 90.53 & 89.19 & 86.67 & 90.67 & 89.86 & 91.30 & 91.30 & 86.49 & 85.97 & 88.35 & 86.94 \\
\rowcolor{gray!20}
Qwen2.5 32B & -- & 76.88 & -- & -- & 79.75 & -- & -- & 94.37 & -- & -- & 87.37 & -- & -- & 84.00 & -- & -- & 89.86 & -- & -- & 85.37 & -- & 85.37 \\
Qwen2 VL 7B & 71.25 & 74.38 & 76.25 & 82.82 & 80.37 & 84.05 & 92.64 & 93.51 & 93.51 & 85.26 & 88.42 & 92.63 & 80.00 & 82.67 & 84.00 & 86.96 & 85.51 & 84.06 & 83.16 & 84.14 & 85.75 & 84.35 \\
\rowcolor{gray!20}
MiniCPM V 2.6 & 72.50 & 72.50 & 75.00 & 81.60 & 79.14 & 80.37 & 88.74 & 90.48 & 93.07 & 80.00 & 87.37 & 90.53 & 80.00 & 77.33 & 86.67 & 88.41 & 85.51 & 86.96 & 81.87 & 82.05 & 85.43 & 83.12 \\
InternVL2.5 26B & 77.50 & 74.38 & 80.62 & 87.12 & 75.46 & 87.12 & 91.77 & 91.77 & 96.54 & 88.42 & 84.21 & 93.68 & 84.00 & 85.33 & 88.00 & 91.30 & 79.71 & 86.96 & 86.69 & 81.81 & 88.82 & 85.77 \\
\rowcolor{gray!20}
Phi 3.5 Mini & -- & 74.38 & -- & -- & 72.39 & -- & -- & 88.31 & -- & -- & 83.16 & -- & -- & 81.33 & -- & -- & 86.96 & -- & -- & 81.09 & -- & 81.09 \\
InternLM2.5 7B & -- & 74.38 & -- & -- & 76.69 & -- & -- & 90.48 & -- & -- & 80.00 & -- & -- & 78.67 & -- & -- & 85.51 & -- & -- & 80.95 & -- & 80.95 \\
\rowcolor{gray!20}
Centurio Qwen & 75.63 & 74.38 & 80.00 & 79.75 & 76.69 & 82.82 & 86.58 & 92.64 & 92.21 & 83.16 & 86.32 & 89.47 & 78.67 & 77.33 & 88.00 & 86.96 & 76.81 & 89.86 & 81.79 & 80.69 & 87.06 & 83.18 \\
InternLM2.5 20B & -- & 74.38 & -- & -- & 75.46 & -- & -- & 89.18 & -- & -- & 86.32 & -- & -- & 76.00 & -- & -- & 82.61 & -- & -- & 80.66 & -- & 80.66 \\
\rowcolor{gray!20}
Qwen2.5 7B & -- & 71.88 & -- & -- & 72.39 & -- & -- & 93.51 & -- & -- & 85.26 & -- & -- & 77.33 & -- & -- & 81.16 & -- & -- & 80.26 & -- & 80.26 \\
Aya Expanse 8B & -- & 68.12 & -- & -- & 77.30 & -- & -- & 91.77 & -- & -- & 81.05 & -- & -- & 80.00 & -- & -- & 81.16 & -- & -- & 79.90 & -- & 79.90 \\
\rowcolor{gray!20}
InternVL2.5 8B & 68.12 & 72.50 & 75.63 & 83.44 & 76.07 & 83.44 & 87.88 & 89.61 & 94.37 & 84.21 & 83.16 & 92.63 & 84.00 & 73.33 & 89.33 & 88.41 & 81.16 & 92.75 & 82.68 & 79.31 & 88.03 & 83.34 \\
Centurio Aya & 80.62 & 68.12 & 78.75 & 82.21 & 75.46 & 80.37 & 90.91 & 85.71 & 92.21 & 84.21 & 82.11 & 85.26 & 81.33 & 82.67 & 85.33 & 85.51 & 81.16 & 91.30 & 84.13 & 79.21 & 85.54 & 82.96 \\
\rowcolor{gray!20}
Phi 3.5 Vision & 65.62 & 72.50 & 75.63 & 69.94 & 70.55 & 76.69 & 89.18 & 91.34 & 95.24 & 80.00 & 81.05 & 86.32 & 72.00 & 80.00 & 86.67 & 85.51 & 79.71 & 88.41 & 77.04 & 79.19 & 84.82 & 80.35 \\
InternVL2.5 4B & 66.88 & 66.88 & 76.25 & 84.66 & 75.46 & 84.05 & 87.01 & 86.15 & 93.07 & 83.16 & 78.95 & 87.37 & 80.00 & 82.67 & 86.67 & 86.96 & 84.06 & 89.86 & 81.44 & 79.03 & 86.21 & 82.23 \\
\rowcolor{gray!20}
Qwen2 VL 2B & 77.50 & 72.50 & 78.75 & 84.05 & 64.42 & 84.05 & 91.77 & 82.68 & 92.21 & 88.42 & 81.05 & 86.32 & 84.00 & 70.67 & 89.33 & 88.41 & 79.71 & 91.30 & 85.69 & 75.17 & 86.99 & 82.62 \\
Qwen2.5 3B & -- & 68.75 & -- & -- & 73.01 & -- & -- & 83.12 & -- & -- & 73.68 & -- & -- & 74.67 & -- & -- & 75.36 & -- & -- & 74.76 & -- & 74.76 \\
\rowcolor{gray!20}
Qwen2.5 1.5B & -- & 61.88 & -- & -- & 65.03 & -- & -- & 82.25 & -- & -- & 78.95 & -- & -- & 72.00 & -- & -- & 78.26 & -- & -- & 73.06 & -- & 73.06 \\
Qwen2.5 0.5B & -- & 68.12 & -- & -- & 72.39 & -- & -- & 67.53 & -- & -- & 65.26 & -- & -- & 70.67 & -- & -- & 55.07 & -- & -- & 66.51 & -- & 66.51 \\
\rowcolor{gray!20}
InternLM2.5 1.8B & -- & 56.25 & -- & -- & 65.03 & -- & -- & 65.37 & -- & -- & 60.00 & -- & -- & 66.67 & -- & -- & 66.67 & -- & -- & 63.33 & -- & 63.33 \\
InternVL2.5 2B & 70.62 & 51.88 & 72.50 & 76.69 & 58.28 & 71.78 & 77.92 & 72.29 & 82.68 & 83.16 & 62.11 & 83.16 & 73.33 & 66.67 & 82.67 & 84.06 & 60.87 & 89.86 & 77.63 & 62.02 & 80.44 & 73.36 \\
\rowcolor{gray!20}
InternVL2.5 1B & 63.75 & 58.75 & 66.88 & 62.58 & 60.74 & 74.23 & 64.50 & 61.90 & 80.09 & 77.89 & 57.89 & 87.37 & 62.67 & 68.00 & 82.67 & 75.36 & 59.42 & 82.61 & 67.79 & 61.12 & 78.97 & 69.29 \\
Gemini Pro & 76.25 & 59.38 & 78.13 & 68.10 & 55.21 & 82.21 & 82.25 & 56.28 & 89.61 & 79.79 & 61.05 & 85.11 & 79.73 & 61.33 & 84.00 & 72.46 & 65.22 & 95.65 & 76.43 & 59.75 & 85.78 & 73.99 \\
\midrule
Average X-Large LVLMs & 78.44 & 80.31 & 84.06 & 85.89 & 83.44 & 88.65 & 92.42 & 94.81 & 96.97 & 87.37 & 90.53 & 92.63 & 87.33 & 89.33 & 90.67 & 92.03 & 88.41 & 92.03 & 87.24 & 87.80 & 90.84 & 88.63 \\
\rowcolor{gray!20}
Average Large LVLMs & 79.38 & 77.81 & 82.50 & 86.50 & 80.06 & 86.20 & 90.91 & 93.51 & 94.59 & 87.37 & 85.26 & 93.16 & 86.67 & 88.00 & 90.00 & 90.58 & 83.33 & 89.13 & 86.90 & 84.66 & 89.26 & 86.94 \\
Average Medium LVLMs & 73.62 & 72.38 & 77.12 & 81.96 & 77.55 & 82.21 & 89.35 & 90.39 & 93.07 & 83.37 & 85.47 & 90.11 & 80.80 & 78.67 & 86.67 & 87.25 & 82.03 & 88.99 & 82.73 & 81.08 & 86.36 & 83.39 \\
\rowcolor{gray!20}
Average Small LVLMs & 68.88 & 64.50 & 74.00 & 75.58 & 65.89 & 78.16 & 82.08 & 78.87 & 88.66 & 82.53 & 72.21 & 86.11 & 74.40 & 73.60 & 85.60 & 84.06 & 72.75 & 88.41 & 77.92 & 71.31 & 83.49 & 77.57 \\
Average LVLMs & 73.44 & 71.47 & 77.77 & 80.89 & 74.58 & 82.25 & 87.41 & 87.35 & 92.27 & 84.21 & 81.43 & 89.47 & 80.29 & 79.71 & 87.33 & 87.27 & 79.81 & 89.23 & 82.25 & 79.06 & 86.39 & 82.57 \\
\rowcolor{gray!20}
Average X-Large LLMs & -- & 81.25 & -- & -- & 84.05 & -- & -- & 96.10 & -- & -- & 89.47 & -- & -- & 86.67 & -- & -- & 89.86 & -- & -- & 87.90 & -- & 87.90 \\
Average Large LLMs & -- & 75.62 & -- & -- & 77.61 & -- & -- & 91.77 & -- & -- & 86.84 & -- & -- & 80.00 & -- & -- & 86.23 & -- & -- & 83.02 & -- & 83.02 \\
\rowcolor{gray!20}
Average Medium LLMs & -- & 71.46 & -- & -- & 75.46 & -- & -- & 91.92 & -- & -- & 82.11 & -- & -- & 78.67 & -- & -- & 82.61 & -- & -- & 80.37 & -- & 80.37 \\
Average Small LLMs & -- & 65.88 & -- & -- & 69.57 & -- & -- & 77.32 & -- & -- & 72.21 & -- & -- & 73.07 & -- & -- & 72.46 & -- & -- & 71.75 & -- & 71.75 \\
\rowcolor{gray!20}
Average LLMs & -- & 70.57 & -- & -- & 73.95 & -- & -- & 85.64 & -- & -- & 79.14 & -- & -- & 77.09 & -- & -- & 79.31 & -- & -- & 77.62 & -- & 77.62 \\
Average X-Large & 78.44 & 80.62 & 84.06 & 85.89 & 83.64 & 88.65 & 92.42 & 95.24 & 96.97 & 87.37 & 90.18 & 92.63 & 87.33 & 88.44 & 90.67 & 92.03 & 88.89 & 92.03 & 87.24 & 87.83 & 90.84 & 88.39 \\
\rowcolor{gray!20}
Average Large & 79.38 & 76.72 & 82.50 & 86.50 & 78.83 & 86.20 & 90.91 & 92.64 & 94.59 & 87.37 & 86.05 & 93.16 & 86.67 & 84.00 & 90.00 & 90.58 & 84.78 & 89.13 & 86.90 & 83.84 & 89.26 & 84.98 \\
Average Medium & 73.62 & 72.03 & 77.12 & 81.96 & 76.76 & 82.21 & 89.35 & 90.96 & 93.07 & 83.37 & 84.21 & 90.11 & 80.80 & 78.67 & 86.67 & 87.25 & 82.25 & 88.99 & 82.73 & 80.81 & 86.36 & 82.26 \\
\rowcolor{gray!20}
Average Small & 68.88 & 65.19 & 74.00 & 75.58 & 67.73 & 78.16 & 82.08 & 78.10 & 88.66 & 82.53 & 72.21 & 86.11 & 74.40 & 73.33 & 85.60 & 84.06 & 72.61 & 88.41 & 77.92 & 71.53 & 83.49 & 74.66 \\
Average Open & 73.44 & 71.08 & 77.77 & 80.89 & 74.31 & 82.25 & 87.41 & 86.60 & 92.27 & 84.21 & 80.42 & 89.47 & 80.29 & 78.56 & 87.33 & 87.27 & 79.59 & 89.23 & 82.25 & 78.43 & 86.39 & 80.39 \\
\rowcolor{gray!20}
Average Proprietary & 78.00 & 77.62 & 81.75 & 80.12 & 78.77 & 84.42 & 89.18 & 87.10 & 94.81 & 85.82 & 84.21 & 89.23 & 85.25 & 82.67 & 88.80 & 85.51 & 85.51 & 92.46 & 83.98 & 82.65 & 88.58 & 85.07 \\
Average & 74.64 & 72.17 & 78.82 & 80.69 & 75.05 & 82.82 & 87.88 & 86.68 & 92.94 & 84.63 & 81.05 & 89.41 & 81.59 & 79.24 & 87.72 & 86.80 & 80.58 & 90.08 & 82.71 & 79.13 & 86.96 & 81.17 \\
\bottomrule
    \end{tabular}
    }%
  \caption{\dsname Cultural Origin Question Answering -- Regions (\coqar) results. The reported score is relaxed accuracy. The columns \textbf{I} and \textbf{T} stand for \textbf{image-only} and \textbf{text-only} inputs to the model.}
  \label{tab:coqa-regions:scores}
\end{table}

\begin{table}[htbp]
  \centering
  \renewcommand{\arraystretch}{.97}
    \resizebox{\textwidth}{!}{%
    \begin{tabular}{lccc ccc ccc ccc ccc ccc | cccc}
    \toprule
     & \multicolumn{3}{c}{\textsc{West EU \& North Am.}} &
     \multicolumn{3}{c}{\textsc{East EU}} &
     \multicolumn{3}{c}{\textsc{Asia \& Pacific}} &
     \multicolumn{3}{c}{\textsc{Lat. Am. \& Carib.}} &
     \multicolumn{3}{c}{\textsc{Arab}} &
     \multicolumn{3}{c}{\textsc{Subs. Africa}} &
     \multicolumn{4}{c}{\textsc{Average}} \\
    \cmidrule(lr){2-4} \cmidrule(lr){5-7} \cmidrule(lr){8-10} \cmidrule(lr){11-13} \cmidrule(lr){14-16} \cmidrule(lr){17-19} \cmidrule(lr){20-23}
    & I & T & I+T & I & T & I+T & I & T & I+T & I & T & I+T & I & T & I+T & I & T & I+T & I & T & I+T & Avg. \\
\midrule
\rowcolor{gray!20}
Claude 3.5 Sonnet & 79.23 & 96.72 & 95.63 & 82.35 & 97.65 & 96.47 & 76.62 & 97.84 & 95.67 & 70.21 & 98.94 & 100.00 & 76.47 & 97.65 & 96.47 & 83.82 & 97.06 & 91.18 & 78.12 & 97.64 & 95.90 & 90.55 \\
GPT-4o & 93.44 & 95.08 & 96.17 & 94.71 & 98.24 & 98.24 & 93.51 & 97.40 & 98.27 & 97.87 & 98.94 & 98.94 & 95.29 & 95.29 & 98.82 & 95.59 & 97.06 & 100.00 & 95.07 & 97.00 & 98.41 & 96.83 \\
\rowcolor{gray!20}
InternVL2.5 78B & 83.06 & 94.54 & 97.81 & 80.59 & 95.88 & 97.65 & 83.12 & 93.51 & 96.54 & 81.91 & 98.94 & 98.94 & 90.59 & 97.65 & 97.65 & 83.82 & 97.06 & 98.53 & 83.85 & 96.26 & 97.85 & 92.65 \\
Qwen2.5 72B & -- & 93.44 & -- & -- & 96.47 & -- & -- & 94.81 & -- & -- & 98.94 & -- & -- & 97.65 & -- & -- & 94.12 & -- & -- & 95.90 & -- & 95.90 \\
\rowcolor{gray!20}
GPT-4o Mini & 89.07 & 93.99 & 95.63 & 90.00 & 95.29 & 97.65 & 90.48 & 93.51 & 96.97 & 90.43 & 95.74 & 100.00 & 94.12 & 88.24 & 97.65 & 91.18 & 95.59 & 98.53 & 90.88 & 93.73 & 97.74 & 94.11 \\
Qwen2.5 32B & -- & 91.26 & -- & -- & 93.53 & -- & -- & 91.77 & -- & -- & 94.68 & -- & -- & 95.29 & -- & -- & 92.65 & -- & -- & 93.20 & -- & 93.20 \\
\rowcolor{gray!20}
InternVL2.5 38B & 78.69 & 91.80 & 92.35 & 77.06 & 91.18 & 92.94 & 77.49 & 93.07 & 93.94 & 79.79 & 95.74 & 96.81 & 84.71 & 94.12 & 95.29 & 88.24 & 92.65 & 98.53 & 80.99 & 93.09 & 94.98 & 89.69 \\
Qwen2 VL 72B & 87.98 & 87.43 & 95.08 & 94.12 & 90.59 & 96.47 & 90.04 & 90.04 & 97.84 & 91.49 & 97.87 & 98.94 & 92.94 & 89.41 & 98.82 & 91.18 & 97.06 & 98.53 & 91.29 & 92.07 & 97.61 & 93.66 \\
\rowcolor{gray!20}
Gemini Flash & 90.56 & 89.01 & 97.27 & 90.59 & 88.82 & 97.06 & 91.77 & 90.48 & 98.70 & 90.43 & 93.62 & 97.87 & 90.59 & 88.24 & 97.65 & 88.24 & 95.59 & 97.06 & 90.36 & 90.96 & 97.60 & 92.97 \\
InternVL2.5 26B & 78.14 & 87.98 & 92.90 & 78.24 & 88.24 & 94.71 & 76.19 & 90.48 & 93.94 & 80.85 & 94.68 & 94.68 & 81.18 & 91.76 & 91.76 & 80.88 & 91.18 & 95.59 & 79.25 & 90.72 & 93.93 & 87.96 \\
\rowcolor{gray!20}
Qwen2.5 7B & -- & 86.34 & -- & -- & 88.24 & -- & -- & 85.28 & -- & -- & 95.74 & -- & -- & 90.59 & -- & -- & 94.12 & -- & -- & 90.05 & -- & 90.05 \\
Aya Expanse 8B & -- & 87.43 & -- & -- & 88.24 & -- & -- & 90.04 & -- & -- & 93.62 & -- & -- & 88.24 & -- & -- & 89.71 & -- & -- & 89.54 & -- & 89.54 \\
\rowcolor{gray!20}
InternLM2.5 20B & -- & 86.89 & -- & -- & 87.06 & -- & -- & 90.91 & -- & -- & 90.43 & -- & -- & 85.88 & -- & -- & 89.71 & -- & -- & 88.48 & -- & 88.48 \\
MiniCPM V 2.6 & 81.97 & 84.70 & 90.16 & 81.18 & 86.47 & 92.94 & 78.79 & 87.45 & 92.21 & 86.17 & 87.23 & 92.55 & 82.35 & 89.41 & 96.47 & 88.24 & 92.65 & 94.12 & 83.12 & 87.98 & 93.08 & 88.06 \\
\rowcolor{gray!20}
Qwen2 VL 7B & 87.43 & 83.61 & 90.71 & 82.35 & 85.29 & 92.94 & 87.01 & 84.85 & 94.37 & 91.49 & 88.30 & 94.68 & 84.71 & 88.24 & 96.47 & 92.65 & 94.12 & 97.06 & 87.61 & 87.40 & 94.37 & 89.79 \\
Qwen2.5 3B & -- & 81.42 & -- & -- & 85.88 & -- & -- & 84.85 & -- & -- & 92.55 & -- & -- & 88.24 & -- & -- & 86.76 & -- & -- & 86.62 & -- & 86.62 \\
\rowcolor{gray!20}
InternLM2.5 7B & -- & 83.61 & -- & -- & 85.88 & -- & -- & 85.71 & -- & -- & 90.43 & -- & -- & 77.65 & -- & -- & 88.24 & -- & -- & 85.25 & -- & 85.25 \\
Centurio Qwen & 78.69 & 82.51 & 89.07 & 78.82 & 82.94 & 89.41 & 78.79 & 84.42 & 92.64 & 76.60 & 85.11 & 91.49 & 80.00 & 83.53 & 88.24 & 79.41 & 91.18 & 92.65 & 78.72 & 84.95 & 90.58 & 84.75 \\
\rowcolor{gray!20}
Centurio Aya & 65.57 & 83.61 & 85.79 & 72.35 & 81.76 & 85.88 & 75.76 & 85.71 & 88.31 & 74.47 & 87.23 & 82.98 & 70.59 & 80.00 & 89.41 & 66.18 & 89.71 & 89.71 & 70.82 & 84.67 & 87.01 & 80.83 \\
InternVL2.5 8B & 68.31 & 82.51 & 88.52 & 70.59 & 84.71 & 90.00 & 75.32 & 86.58 & 91.34 & 75.53 & 87.23 & 94.68 & 76.47 & 83.53 & 90.59 & 82.35 & 82.35 & 89.71 & 74.76 & 84.49 & 90.81 & 83.35 \\
\rowcolor{gray!20}
Phi 3.5 Mini & -- & 80.87 & -- & -- & 82.94 & -- & -- & 83.98 & -- & -- & 84.04 & -- & -- & 82.35 & -- & -- & 88.24 & -- & -- & 83.74 & -- & 83.74 \\
InternVL2.5 4B & 68.85 & 77.05 & 89.62 & 72.35 & 82.94 & 89.41 & 71.43 & 86.15 & 90.48 & 76.60 & 87.23 & 89.36 & 72.94 & 81.18 & 84.71 & 76.47 & 82.35 & 97.06 & 73.11 & 82.82 & 90.11 & 82.01 \\
\rowcolor{gray!20}
Phi 3.5 Vision & 72.13 & 79.78 & 86.89 & 68.82 & 82.94 & 92.35 & 69.70 & 81.82 & 89.61 & 74.47 & 91.49 & 91.49 & 81.18 & 77.65 & 90.59 & 76.47 & 82.35 & 95.59 & 73.79 & 82.67 & 91.09 & 82.52 \\
Qwen2.5 1.5B & -- & 78.69 & -- & -- & 81.18 & -- & -- & 82.68 & -- & -- & 82.98 & -- & -- & 75.29 & -- & -- & 80.88 & -- & -- & 80.28 & -- & 80.28 \\
\rowcolor{gray!20}
Qwen2 VL 2B & 83.06 & 74.32 & 87.43 & 84.71 & 77.06 & 87.65 & 83.55 & 80.95 & 90.48 & 92.55 & 81.91 & 94.68 & 83.53 & 76.47 & 91.76 & 89.71 & 80.88 & 94.12 & 86.18 & 78.60 & 91.02 & 85.27 \\
Qwen2.5 0.5B & -- & 65.03 & -- & -- & 68.82 & -- & -- & 72.29 & -- & -- & 75.53 & -- & -- & 69.41 & -- & -- & 77.94 & -- & -- & 71.51 & -- & 71.51 \\
\rowcolor{gray!20}
InternVL2.5 1B & 61.20 & 66.12 & 73.77 & 59.41 & 65.88 & 73.53 & 62.34 & 75.76 & 77.06 & 67.02 & 74.47 & 76.60 & 56.47 & 63.53 & 75.29 & 55.88 & 72.06 & 70.59 & 60.39 & 69.64 & 74.47 & 68.17 \\
InternLM2.5 1.8B & -- & 63.39 & -- & -- & 66.47 & -- & -- & 71.00 & -- & -- & 67.02 & -- & -- & 58.82 & -- & -- & 64.71 & -- & -- & 65.23 & -- & 65.23 \\
\rowcolor{gray!20}
InternVL2.5 2B & 62.84 & 65.57 & 74.32 & 61.76 & 64.71 & 72.35 & 61.04 & 68.40 & 80.95 & 67.02 & 68.09 & 80.85 & 67.06 & 55.29 & 74.12 & 73.53 & 63.24 & 77.94 & 65.54 & 64.22 & 76.76 & 68.84 \\
Gemini Pro & 76.67 & 43.17 & 92.70 & 75.88 & 39.41 & 92.94 & 78.79 & 34.20 & 92.64 & 78.72 & 39.36 & 93.62 & 78.82 & 35.29 & 91.76 & 82.35 & 19.12 & 94.12 & 78.54 & 35.09 & 92.96 & 68.86 \\
\midrule
Average X-Large LVLMs & 85.52 & 90.98 & 96.45 & 87.35 & 93.24 & 97.06 & 86.58 & 91.77 & 97.19 & 86.70 & 98.40 & 98.94 & 91.76 & 93.53 & 98.24 & 87.50 & 97.06 & 98.53 & 87.57 & 94.16 & 97.73 & 93.16 \\
\rowcolor{gray!20}
Average Large LVLMs & 78.42 & 89.89 & 92.62 & 77.65 & 89.71 & 93.82 & 76.84 & 91.77 & 93.94 & 80.32 & 95.21 & 95.74 & 82.94 & 92.94 & 93.53 & 84.56 & 91.91 & 97.06 & 80.12 & 91.90 & 94.46 & 88.82 \\
Average Medium LVLMs & 76.39 & 83.39 & 88.85 & 77.06 & 84.24 & 90.24 & 79.13 & 85.80 & 91.77 & 80.85 & 87.02 & 91.28 & 78.82 & 84.94 & 92.24 & 81.76 & 90.00 & 92.65 & 79.01 & 85.90 & 91.17 & 85.36 \\
\rowcolor{gray!20}
Average Small LVLMs & 69.62 & 72.57 & 82.40 & 69.41 & 74.71 & 83.06 & 69.61 & 78.61 & 85.71 & 75.53 & 80.64 & 86.60 & 72.24 & 70.82 & 83.29 & 74.41 & 76.18 & 87.06 & 71.80 & 75.59 & 84.69 & 77.36 \\
Average LVLMs & 75.57 & 81.54 & 88.17 & 75.88 & 82.90 & 89.16 & 76.47 & 84.94 & 90.69 & 79.71 & 87.54 & 91.34 & 78.91 & 82.27 & 90.08 & 80.36 & 86.34 & 92.12 & 77.82 & 84.26 & 90.26 & 84.11 \\
\rowcolor{gray!20}
Average X-Large LLMs & -- & 93.44 & -- & -- & 96.47 & -- & -- & 94.81 & -- & -- & 98.94 & -- & -- & 97.65 & -- & -- & 94.12 & -- & -- & 95.90 & -- & 95.90 \\
Average Large LLMs & -- & 89.07 & -- & -- & 90.29 & -- & -- & 91.34 & -- & -- & 92.55 & -- & -- & 90.59 & -- & -- & 91.18 & -- & -- & 90.84 & -- & 90.84 \\
\rowcolor{gray!20}
Average Medium LLMs & -- & 85.79 & -- & -- & 87.45 & -- & -- & 87.01 & -- & -- & 93.26 & -- & -- & 85.49 & -- & -- & 90.69 & -- & -- & 88.28 & -- & 88.28 \\
Average Small LLMs & -- & 73.88 & -- & -- & 77.06 & -- & -- & 78.96 & -- & -- & 80.43 & -- & -- & 74.82 & -- & -- & 79.71 & -- & -- & 77.48 & -- & 77.48 \\
\rowcolor{gray!20}
Average LLMs & -- & 81.67 & -- & -- & 84.06 & -- & -- & 84.85 & -- & -- & 87.81 & -- & -- & 82.67 & -- & -- & 86.10 & -- & -- & 84.53 & -- & 84.53 \\
Average X-Large & 85.52 & 91.80 & 96.45 & 87.35 & 94.31 & 97.06 & 86.58 & 92.78 & 97.19 & 86.70 & 98.58 & 98.94 & 91.76 & 94.90 & 98.24 & 87.50 & 96.08 & 98.53 & 87.57 & 94.74 & 97.73 & 94.07 \\
\rowcolor{gray!20}
Average Large & 78.42 & 89.48 & 92.62 & 77.65 & 90.00 & 93.82 & 76.84 & 91.56 & 93.94 & 80.32 & 93.88 & 95.74 & 82.94 & 91.76 & 93.53 & 84.56 & 91.54 & 97.06 & 80.12 & 91.37 & 94.46 & 89.83 \\
Average Medium & 76.39 & 84.29 & 88.85 & 77.06 & 85.44 & 90.24 & 79.13 & 86.26 & 91.77 & 80.85 & 89.36 & 91.28 & 78.82 & 85.15 & 92.24 & 81.76 & 90.26 & 92.65 & 79.01 & 86.79 & 91.17 & 86.45 \\
\rowcolor{gray!20}
Average Small & 69.62 & 73.22 & 82.40 & 69.41 & 75.88 & 83.06 & 69.61 & 78.79 & 85.71 & 75.53 & 80.53 & 86.60 & 72.24 & 72.82 & 83.29 & 74.41 & 77.94 & 87.06 & 71.80 & 76.53 & 84.69 & 77.42 \\
Average Open & 75.57 & 81.60 & 88.17 & 75.88 & 83.41 & 89.16 & 76.47 & 84.90 & 90.69 & 79.71 & 87.66 & 91.34 & 78.91 & 82.45 & 90.08 & 80.36 & 86.24 & 92.12 & 77.82 & 84.38 & 90.26 & 84.29 \\
\rowcolor{gray!20}
Average Proprietary & 85.79 & 83.59 & 95.48 & 86.71 & 83.88 & 96.47 & 86.23 & 82.68 & 96.45 & 85.53 & 85.32 & 98.09 & 87.06 & 80.94 & 96.47 & 88.24 & 80.88 & 96.18 & 86.59 & 82.88 & 96.52 & 88.66 \\
Average & 78.26 & 81.93 & 90.10 & 78.73 & 83.49 & 91.08 & 79.04 & 84.53 & 92.21 & 81.24 & 87.27 & 93.11 & 81.05 & 82.20 & 91.76 & 82.43 & 85.34 & 93.19 & 80.13 & 84.13 & 91.91 & 85.02 \\
\bottomrule
    \end{tabular}
    }%
  \caption{\dsname Cultural Origin Question Answering -- Country (\coqac) results. The reported score is relaxed accuracy. The columns \textbf{I} and \textbf{T} stand for \textbf{image-only} and \textbf{text-only} inputs to the model.}
  \label{tab:coqa-countries:scores}
\end{table}

%% file: src/992_4_appendix_results_ckqa.tex
\newpage
\subsection{\ckqa}
\label{appendix:sec:analyses:ckqa}
\subsubsection{LLM-as-a-Judge Evaluation}
\label{appendix:sec:analyses:ckqa:judge}
To evaluate the \ckqad and \ckqan tasks, we used GPT-4o (\texttt{gpt-4o-2024-11-20}) as a judge using the prompts shown in the next section.
For each sample, we used the same system prompt and generated user prompts per sample individually.
\input{src/992_4_appendix_results_ckqa_prompt}

\subsubsection{Results}
\label{appendix:sec:analyses:ckqa:results}
\begin{table}[htbp]
  \centering
  \renewcommand{\arraystretch}{.97}
    \resizebox{\textwidth}{!}{
    \begin{tabular}{lccc ccc ccc ccc ccc ccc | cccc}
    \toprule
     & \multicolumn{3}{c}{\textsc{West EU \& North Am.}} &
     \multicolumn{3}{c}{\textsc{East EU}} &
     \multicolumn{3}{c}{\textsc{Asia \& Pacific}} &
     \multicolumn{3}{c}{\textsc{Lat. Am. \& Carib.}} &
     \multicolumn{3}{c}{\textsc{Arab}} &
     \multicolumn{3}{c}{\textsc{Subs. Africa}} &
     \multicolumn{4}{c}{\textsc{Average}} \\
    \cmidrule(lr){2-4} \cmidrule(lr){5-7} \cmidrule(lr){8-10} \cmidrule(lr){11-13} \cmidrule(lr){14-16} \cmidrule(lr){17-19} \cmidrule(lr){20-23}
    & I & T & I+T & I & T & I+T & I & T & I+T & I & T & I+T & I & T & I+T & I & T & I+T & I & T & I+T & Avg. \\
    \midrule
\rowcolor{gray!20}
GPT-4o & 46.98 & 56.78 & 57.21 & 38.20 & 54.30 & 54.87 & 44.71 & 59.20 & 58.56 & 34.08 & 51.53 & 52.45 & 44.41 & 57.76 & 56.91 & 29.04 & 50.68 & 51.37 & 39.57 & 55.04 & 55.23 & 49.95 \\
Claude 3.5 Sonnet & 43.05 & 56.64 & 55.60 & 35.20 & 55.97 & 50.07 & 39.54 & 59.67 & 54.05 & 26.84 & 53.32 & 49.23 & 41.45 & 56.78 & 53.68 & 24.73 & 50.34 & 44.79 & 35.14 & 55.45 & 51.24 & 47.28 \\
\rowcolor{gray!20}
Gemini Pro & 42.28 & 53.29 & 57.21 & 36.80 & 50.07 & 53.67 & 37.47 & 52.94 & 55.18 & 29.18 & 49.44 & 50.00 & 38.68 & 48.68 & 54.08 & 22.05 & 41.23 & 46.23 & 34.41 & 49.28 & 52.73 & 45.47 \\
Qwen2.5 72B & -- & 47.55 & -- & -- & 45.17 & -- & -- & 50.62 & -- & -- & 42.70 & -- & -- & 44.47 & -- & -- & 37.95 & -- & -- & 44.74 & -- & 44.74 \\
\rowcolor{gray!20}
Qwen2.5 32B & -- & 47.89 & -- & -- & 43.73 & -- & -- & 48.45 & -- & -- & 40.71 & -- & -- & 42.17 & -- & -- & 39.25 & -- & -- & 43.70 & -- & 43.70 \\
GPT-4o Mini & 34.36 & 48.89 & 55.70 & 27.70 & 46.63 & 54.00 & 30.73 & 49.05 & 53.61 & 24.95 & 43.72 & 49.44 & 36.84 & 47.50 & 54.21 & 21.03 & 39.25 & 46.78 & 29.27 & 45.84 & 52.29 & 42.47 \\
\rowcolor{gray!20}
Gemini Flash & 36.54 & 52.75 & 54.70 & 29.67 & 46.87 & 51.30 & 31.78 & 50.40 & 51.62 & 23.20 & 46.07 & 49.07 & 32.43 & 46.64 & 51.45 & 16.44 & 36.37 & 42.12 & 28.34 & 46.52 & 50.04 & 41.63 \\
Phi 3.5 Mini & -- & 40.40 & -- & -- & 35.23 & -- & -- & 38.27 & -- & -- & 34.80 & -- & -- & 34.87 & -- & -- & 30.00 & -- & -- & 35.60 & -- & 35.60 \\
\rowcolor{gray!20}
Aya Expanse 8B & -- & 40.17 & -- & -- & 36.13 & -- & -- & 39.42 & -- & -- & 34.18 & -- & -- & 36.32 & -- & -- & 26.71 & -- & -- & 35.49 & -- & 35.49 \\
Qwen2.5 7B & -- & 38.39 & -- & -- & 36.50 & -- & -- & 38.78 & -- & -- & 34.23 & -- & -- & 34.01 & -- & -- & 29.04 & -- & -- & 35.16 & -- & 35.16 \\
\rowcolor{gray!20}
InternLM2.5 20B & -- & 37.01 & -- & -- & 34.13 & -- & -- & 36.59 & -- & -- & 31.17 & -- & -- & 32.83 & -- & -- & 27.53 & -- & -- & 33.21 & -- & 33.21 \\
Llama 3.2 11B Vision & -- & 36.44 & -- & -- & 32.77 & -- & -- & 35.75 & -- & -- & 30.00 & -- & -- & 33.68 & -- & -- & 27.40 & -- & -- & 32.67 & -- & 32.67 \\
\rowcolor{gray!20}
InternVL2.5 38B & 23.72 & 41.21 & 37.62 & 18.63 & 38.80 & 37.03 & 20.51 & 41.55 & 39.96 & 23.72 & 33.32 & 38.72 & 24.08 & 35.46 & 39.67 & 15.96 & 32.47 & 33.49 & 21.10 & 37.14 & 37.75 & 32.00 \\
InternVL2.5 78B & 19.33 & 40.84 & 36.28 & 17.63 & 37.73 & 37.10 & 19.16 & 41.57 & 38.23 & 19.64 & 37.50 & 36.58 & 22.89 & 35.66 & 42.57 & 14.86 & 33.22 & 35.34 & 18.92 & 37.75 & 37.68 & 31.45 \\
\rowcolor{gray!20}
Qwen2 VL 72B & 20.67 & 40.81 & 41.01 & 17.37 & 37.03 & 42.13 & 18.23 & 40.02 & 42.10 & 14.08 & 36.02 & 37.60 & 23.42 & 36.12 & 41.91 & 10.62 & 29.38 & 35.14 & 17.40 & 36.56 & 39.98 & 31.31 \\
InternLM2.5 7B & -- & 34.33 & -- & -- & 32.30 & -- & -- & 34.62 & -- & -- & 31.17 & -- & -- & 29.93 & -- & -- & 23.49 & -- & -- & 30.97 & -- & 30.97 \\
\rowcolor{gray!20}
Qwen2.5 3B & -- & 32.75 & -- & -- & 28.90 & -- & -- & 33.05 & -- & -- & 28.47 & -- & -- & 26.58 & -- & -- & 22.81 & -- & -- & 28.76 & -- & 28.76 \\
InternVL2.5 26B & 11.91 & 39.97 & 34.43 & 12.63 & 36.10 & 34.07 & 13.76 & 38.92 & 34.00 & 13.11 & 34.18 & 28.98 & 15.33 & 34.01 & 36.38 & 9.38 & 29.59 & 27.95 & 12.69 & 35.46 & 32.64 & 26.93 \\
\rowcolor{gray!20}
Qwen2 VL 7B & 14.09 & 32.82 & 38.09 & 14.27 & 29.53 & 37.17 & 12.72 & 33.12 & 37.43 & 17.09 & 28.32 & 35.97 & 17.17 & 29.28 & 35.46 & 9.38 & 20.55 & 30.96 & 14.12 & 28.94 & 35.85 & 26.30 \\
MiniCPM V 2.6 & 18.49 & 34.70 & 36.28 & 13.60 & 30.47 & 34.60 & 15.88 & 33.76 & 34.96 & 18.67 & 29.03 & 34.34 & 17.50 & 30.20 & 36.84 & 9.93 & 18.42 & 24.52 & 15.68 & 29.43 & 33.59 & 26.23 \\
\rowcolor{gray!20}
Centurio Qwen & 14.87 & 31.38 & 32.05 & 14.47 & 29.23 & 29.97 & 15.07 & 31.57 & 34.76 & 16.38 & 27.40 & 35.77 & 18.62 & 27.30 & 36.32 & 13.01 & 20.41 & 32.60 & 15.40 & 27.88 & 33.58 & 25.62 \\
Phi 3.5 Vision & 12.08 & 36.64 & 35.03 & 12.43 & 32.10 & 35.23 & 10.09 & 33.36 & 32.30 & 13.78 & 29.74 & 29.80 & 16.97 & 31.97 & 33.42 & 11.99 & 25.75 & 26.78 & 12.89 & 31.59 & 32.09 & 25.53 \\
\rowcolor{gray!20}
InternVL2.5 8B & 6.81 & 36.28 & 29.97 & 6.80 & 31.33 & 29.13 & 9.07 & 33.72 & 30.49 & 6.58 & 29.74 & 30.36 & 12.11 & 30.53 & 30.26 & 2.53 & 22.67 & 20.96 & 7.32 & 30.71 & 28.53 & 22.19 \\
InternVL2.5 4B & 5.44 & 35.81 & 27.89 & 5.40 & 33.07 & 27.80 & 5.97 & 35.71 & 28.08 & 7.14 & 34.29 & 27.24 & 9.01 & 28.62 & 29.54 & 4.79 & 25.68 & 23.63 & 6.29 & 32.20 & 27.36 & 21.95 \\
\rowcolor{gray!20}
Qwen2.5 1.5B & -- & 24.03 & -- & -- & 20.77 & -- & -- & 26.66 & -- & -- & 20.87 & -- & -- & 21.45 & -- & -- & 16.23 & -- & -- & 21.67 & -- & 21.67 \\
InternLM2.5 1.8B & -- & 23.56 & -- & -- & 22.30 & -- & -- & 22.94 & -- & -- & 18.52 & -- & -- & 21.78 & -- & -- & 14.66 & -- & -- & 20.63 & -- & 20.63 \\
\rowcolor{gray!20}
Qwen2 VL 2B & 11.41 & 23.29 & 31.95 & 11.57 & 20.03 & 28.90 & 12.48 & 20.97 & 29.00 & 14.18 & 20.51 & 28.16 & 16.32 & 18.29 & 29.47 & 11.92 & 13.63 & 25.96 & 12.98 & 19.45 & 28.91 & 20.45 \\
InternVL2.5 1B & 9.87 & 24.09 & 16.51 & 8.50 & 20.43 & 16.83 & 9.78 & 21.28 & 18.50 & 11.22 & 20.41 & 16.07 & 12.83 & 16.51 & 18.75 & 9.93 & 14.93 & 17.60 & 10.35 & 19.61 & 17.38 & 15.78 \\
\rowcolor{gray!20}
InternVL2.5 2B & 5.30 & 23.26 & 18.72 & 4.80 & 19.50 & 21.90 & 4.47 & 22.26 & 20.58 & 7.30 & 21.48 & 19.95 & 9.34 & 21.38 & 20.72 & 5.68 & 15.96 & 18.77 & 6.15 & 20.64 & 20.11 & 15.63 \\
Centurio Aya & 4.80 & 29.33 & 5.94 & 5.30 & 25.50 & 7.53 & 2.94 & 28.85 & 5.02 & 7.50 & 24.23 & 8.47 & 4.28 & 24.21 & 4.21 & 4.45 & 19.38 & 5.89 & 4.88 & 25.25 & 6.18 & 12.10 \\
\rowcolor{gray!20}
Qwen2.5 0.5B & -- & 13.96 & -- & -- & 11.77 & -- & -- & 14.40 & -- & -- & 11.43 & -- & -- & 8.29 & -- & -- & 8.70 & -- & -- & 11.42 & -- & 11.42 \\
\midrule
Average X-Large LVLMs & 20.00 & 40.83 & 38.64 & 17.50 & 37.38 & 39.62 & 18.70 & 40.80 & 40.16 & 16.86 & 36.76 & 37.09 & 23.16 & 35.89 & 42.24 & 12.74 & 31.30 & 35.24 & 18.16 & 37.16 & 38.83 & 31.38 \\
\rowcolor{gray!20}
Average Large LVLMs & 17.81 & 40.59 & 36.02 & 15.63 & 37.45 & 35.55 & 17.14 & 40.24 & 36.98 & 18.42 & 33.75 & 33.85 & 19.70 & 34.74 & 38.03 & 12.67 & 31.03 & 30.72 & 16.90 & 36.30 & 35.20 & 29.46 \\
Average Medium LVLMs & 11.81 & 33.49 & 28.47 & 10.89 & 29.80 & 27.68 & 11.14 & 32.79 & 28.53 & 13.24 & 28.12 & 28.98 & 13.94 & 29.20 & 28.62 & 7.86 & 21.47 & 22.99 & 11.48 & 29.15 & 27.55 & 24.18 \\
\rowcolor{gray!20}
Average Small LVLMs & 8.82 & 28.62 & 26.02 & 8.54 & 25.03 & 26.13 & 8.56 & 26.72 & 25.69 & 10.72 & 25.29 & 24.24 & 12.89 & 23.35 & 26.38 & 8.86 & 19.19 & 22.55 & 9.73 & 24.70 & 25.17 & 19.87 \\
Average LVLMs & 12.77 & 33.79 & 30.13 & 11.67 & 30.24 & 29.96 & 12.15 & 32.83 & 30.39 & 13.60 & 29.08 & 29.14 & 15.70 & 28.88 & 31.11 & 9.60 & 23.30 & 25.68 & 12.58 & 29.69 & 29.40 & 24.41 \\
\rowcolor{gray!20}
Average X-Large LLMs & -- & 47.55 & -- & -- & 45.17 & -- & -- & 50.62 & -- & -- & 42.70 & -- & -- & 44.47 & -- & -- & 37.95 & -- & -- & 44.74 & -- & 44.74 \\
Average Large LLMs & -- & 42.45 & -- & -- & 38.93 & -- & -- & 42.52 & -- & -- & 35.94 & -- & -- & 37.50 & -- & -- & 33.39 & -- & -- & 38.46 & -- & 38.46 \\
\rowcolor{gray!20}
Average Medium LLMs & -- & 37.63 & -- & -- & 34.98 & -- & -- & 37.61 & -- & -- & 33.19 & -- & -- & 33.42 & -- & -- & 26.41 & -- & -- & 33.87 & -- & 33.87 \\
Average Small LLMs & -- & 26.94 & -- & -- & 23.79 & -- & -- & 27.06 & -- & -- & 22.82 & -- & -- & 22.59 & -- & -- & 18.48 & -- & -- & 23.62 & -- & 23.62 \\
\rowcolor{gray!20}
Average LLMs & -- & 34.55 & -- & -- & 31.54 & -- & -- & 34.89 & -- & -- & 29.84 & -- & -- & 30.25 & -- & -- & 25.12 & -- & -- & 31.03 & -- & 31.03 \\
Average X-Large & 20.00 & 43.07 & 38.64 & 17.50 & 39.98 & 39.62 & 18.70 & 44.07 & 40.16 & 16.86 & 38.74 & 37.09 & 23.16 & 38.75 & 42.24 & 12.74 & 33.52 & 35.24 & 18.16 & 39.68 & 38.83 & 35.83 \\
\rowcolor{gray!20}
Average Large & 17.81 & 41.52 & 36.02 & 15.63 & 38.19 & 35.55 & 17.14 & 41.38 & 36.98 & 18.42 & 34.84 & 33.85 & 19.70 & 36.12 & 38.03 & 12.67 & 32.21 & 30.72 & 16.90 & 37.38 & 35.20 & 33.96 \\
Average Medium & 11.81 & 34.87 & 28.47 & 10.89 & 31.53 & 27.68 & 11.14 & 34.40 & 28.53 & 13.24 & 29.81 & 28.98 & 13.94 & 30.61 & 28.62 & 7.86 & 23.12 & 22.99 & 11.48 & 30.72 & 27.55 & 27.41 \\
\rowcolor{gray!20}
Average Small & 8.82 & 27.78 & 26.02 & 8.54 & 24.41 & 26.13 & 8.56 & 26.89 & 25.69 & 10.72 & 24.05 & 24.24 & 12.89 & 22.97 & 26.38 & 8.86 & 18.84 & 22.55 & 9.73 & 24.16 & 25.17 & 21.74 \\
Average Open & 12.77 & 34.11 & 30.13 & 11.67 & 30.79 & 29.96 & 12.15 & 33.70 & 30.39 & 13.60 & 29.40 & 29.14 & 15.70 & 29.46 & 31.11 & 9.60 & 24.07 & 25.68 & 12.58 & 30.26 & 29.40 & 27.21 \\
\rowcolor{gray!20}
Average Proprietary & 40.64 & 53.67 & 56.08 & 33.51 & 50.77 & 52.78 & 36.85 & 54.25 & 54.60 & 27.65 & 48.82 & 50.04 & 38.76 & 51.47 & 54.07 & 22.66 & 43.57 & 46.26 & 33.35 & 50.43 & 52.31 & 45.36 \\
Average & 20.11 & 37.27 & 36.96 & 17.42 & 34.01 & 35.96 & 18.65 & 37.02 & 36.76 & 17.30 & 32.53 & 34.64 & 21.77 & 33.01 & 37.15 & 13.04 & 27.22 & 31.10 & 18.05 & 33.51 & 35.43 & 30.14 \\
    \bottomrule
  \end{tabular}
  }
  \caption{Average Judge Score for the \dsname Cultural Knowledge Question Answering (CKQA) -- Describing. The columns \textbf{I}, \textbf{T}, and \textbf{I+T} stand for \textbf{image-only}, \textbf{text-only}, and \textbf{image+text} input to the model.}
  \label{tab:ckqa-desc:scores}
\end{table}
\begin{table}[htbp]
  \centering
  \renewcommand{\arraystretch}{.97}
    \resizebox{\textwidth}{!}{%
    \begin{tabular}{l c c c c c c |c}
    \toprule
     & \multicolumn{1}{l}{\textsc{West EU \& North Am.}} 
     & \multicolumn{1}{l}{\textsc{East EU}}
     & \multicolumn{1}{l}{\textsc{Asian \& Pacific}}
     & \multicolumn{1}{l}{\textsc{Latin-America \& Caribbean}}
     & \multicolumn{1}{l}{\textsc{Arab}}
     & \multicolumn{1}{l|}{\textsc{Subsaharian Africa}}
     & \multicolumn{1}{l}{\textsc{Average}} \\
\midrule
\rowcolor{gray!20}
GPT-4o & 37.79 & 32.57 & 37.68 & 30.15 & 38.03 & 28.42 & 34.11 \\
Claude 3.5 Sonnet & 40.27 & 33.63 & 39.29 & 25.71 & 38.16 & 24.25 & 33.55 \\
\rowcolor{gray!20}
GPT-4o Mini & 34.46 & 28.73 & 33.08 & 23.67 & 34.87 & 25.89 & 30.12 \\
Centurio Qwen & 18.69 & 19.10 & 21.97 & 18.67 & 25.46 & 15.96 & 19.98 \\
\rowcolor{gray!20}
Gemini Pro & 16.91 & 15.60 & 16.71 & 11.13 & 17.30 & 10.55 & 14.70 \\
Gemini Flash & 15.77 & 16.27 & 14.87 & 11.60 & 14.61 & 11.30 & 14.07 \\
\rowcolor{gray!20}
InternVL2.5 38B & 14.06 & 12.60 & 16.24 & 10.71 & 21.12 & 8.36 & 13.85 \\
Phi 3.5 Vision & 15.17 & 13.67 & 13.54 & 12.45 & 14.28 & 10.75 & 13.31 \\
\rowcolor{gray!20}
InternVL2.5 78B & 12.08 & 14.73 & 14.89 & 7.35 & 15.72 & 7.53 & 12.05 \\
InternVL2.5 26B & 11.51 & 10.50 & 13.16 & 7.65 & 14.34 & 7.74 & 10.82 \\
\rowcolor{gray!20}
InternVL2.5 1B & 10.20 & 9.43 & 10.42 & 10.71 & 14.80 & 8.22 & 10.63 \\
Qwen2 VL 72B & 11.04 & 10.07 & 9.96 & 7.40 & 11.45 & 8.56 & 9.75 \\
\rowcolor{gray!20}
MiniCPM V 2.6 & 8.89 & 8.60 & 11.42 & 4.74 & 10.99 & 9.79 & 9.07 \\
Centurio Aya & 6.95 & 6.57 & 6.06 & 8.78 & 5.20 & 7.40 & 6.83 \\
\rowcolor{gray!20}
InternVL2.5 2B & 6.31 & 6.80 & 6.17 & 7.14 & 8.49 & 3.08 & 6.33 \\
InternVL2.5 4B & 6.28 & 5.47 & 5.07 & 6.02 & 9.28 & 5.00 & 6.19 \\
\rowcolor{gray!20}
InternVL2.5 8B & 6.51 & 5.30 & 4.54 & 6.48 & 9.28 & 3.77 & 5.98 \\
Qwen2 VL 2B & 5.40 & 4.27 & 7.35 & 3.62 & 5.53 & 3.63 & 4.97 \\
\rowcolor{gray!20}
Qwen2 VL 7B & 5.27 & 5.63 & 4.78 & 4.03 & 6.32 & 3.70 & 4.96 \\
\midrule
Average X-Large LVLMs & 11.56 & 12.40 & 12.42 & 7.38 & 13.58 & 8.04 & 10.90 \\
\rowcolor{gray!20}
Average Large LVLMs & 12.78 & 11.55 & 14.70 & 9.18 & 17.73 & 8.05 & 12.34 \\
Average Medium LVLMs & 9.26 & 9.04 & 9.75 & 8.54 & 11.45 & 8.12 & 9.36 \\
\rowcolor{gray!20}
Average Small LVLMs & 8.67 & 7.93 & 8.51 & 7.99 & 10.48 & 6.14 & 8.29 \\
Average LVLMs & 9.88 & 9.48 & 10.40 & 8.27 & 12.30 & 7.39 & 9.62 \\
\rowcolor{gray!20}
Average X-Large & 11.56 & 12.40 & 12.42 & 7.38 & 13.58 & 8.04 & 10.90 \\
Average Large & 12.78 & 11.55 & 14.70 & 9.18 & 17.73 & 8.05 & 12.34 \\
\rowcolor{gray!20}
Average Medium & 9.26 & 9.04 & 9.75 & 8.54 & 11.45 & 8.12 & 9.36 \\
Average Small & 8.67 & 7.93 & 8.51 & 7.99 & 10.48 & 6.14 & 8.29 \\
\rowcolor{gray!20}
Average Open & 9.88 & 9.48 & 10.40 & 8.27 & 12.30 & 7.39 & 9.62 \\
Average Proprietary & 29.04 & 25.36 & 28.33 & 20.45 & 28.59 & 20.08 & 25.31 \\
\rowcolor{gray!20}
Average & 14.92 & 13.66 & 15.12 & 11.47 & 16.59 & 10.73 & 13.75 \\
\bottomrule
    \end{tabular}
    }%
  \caption{Average Judge Score for the \dsname Cultural Knowledge Question Answering (CKQA) -- Naming.}
  \label{tab:ckqa-naming:scores}
\end{table}

%% file: src/992_4_appendix_results_ckqa_prompt.tex
\onecolumn
\newpage
\subsubsection*{System Prompt}
\label{appendix:sec:analyses:ckqa:judge:sys_prompt}
\begin{tcolorbox}[
    enhanced, 
    breakable,
    skin first=enhanced,
    skin middle=enhanced,
    skin last=enhanced,
]
\begin{minted}[fontsize=\footnotesize,breaklines]{markdown}
# Your Role

You are an impartial judge who excels at critical and analytical thinking.

# Your Task

Your task is it to thoroughly analyze and evaluate the correctness of a generated answer to a Cultural Knowledge Test.
1. Carefully analyze the ground truth and the generated answer.
2. Provide a brief summary (1 - 3 sentences) of your analysis, covering the accuracy, relevance, and completeness of the generated answer.
3. Provide a one or two-sentence explanation justifying your final score. Ensure that your explanation and score are consistent with each other and accurately reflect the quality of the generated answer in relation to the ground truth.
4. Provide a single number from 0 to 100 representing the correctness of the generated answer, where:
    0 = Completely incorrect or irrelevant.
    25 = Mostly incorrect or irrelevant.
    50 = Partially correct or relevant.
    75 = Mostly correct and relevant.
    100 = Perfectly correct and complete.

    You may use any whole number within this range to reflect nuanced judgments.

# Output Format

Provide your evaluation in the following format:

```xml
<evaluation>
<analysis>
<!-- Put your analysis summary here -->
</analysis>
<explanation>
<!-- Put your explanation here -->
</explanation>
<score>
<!-- Put your score here -->
</score>
</evaluation>
```
\end{minted}
\end{tcolorbox}
\subsubsection*{User Prompt Template}
\label{appendix:sec:analyses:ckqa:synth:usr_prompt}

\begin{tcolorbox}[
    enhanced, 
    breakable,
    skin first=enhanced,
    skin middle=enhanced,
    skin last=enhanced,
]
\begin{minted}[fontsize=\footnotesize,breaklines]{markdown}
Evaluate the correctness of the generated answer with respect to Ground Truth.

# Ground Truth

```
{GROUND_TRUTH}
```

# Generated Answer

```
{GENERATED_ANSWER}
```

# Evaluation

\end{minted}
\end{tcolorbox}